%% file: main.tex
\documentclass{article} % For LaTeX2e
\input{imports}

\title{\Large{EquiformerV2: Improved Equivariant Transformer \\
for Scaling to Higher-Degree Representations}}

\author{%
  Yi-Lun Liao$^1$ \quad
  Brandon Wood$^2$ \quad
  Abhishek Das$^{2\star}$ \quad %\thanks{denotes equal contribution.} 
  Tess Smidt$^{1\star}$ \\[0.05in]
  {$^1$Massachusetts Institute of Technology \quad
  $^2$FAIR, Meta \quad
  $^\star$Equal contribution
  } \\[0.05in]
  {
  \small{\texttt{\{ylliao,tsmidt\}@mit.edu}} \quad
  \small{\texttt{\{bmwood,abhshkdz\}@meta.com}}
  } 
  \\
    \texttt{\href{https://github.com/atomicarchitects/equiformer_v2}{\color{crimson}{https://github.com/atomicarchitects/equiformer\_v2}}}
}

\iclrfinalcopy % Uncomment for camera-ready version, but NOT for submission.

\begin{document}
\begin{adjustwidth}{-3cm}{-3cm}
\maketitle
\end{adjustwidth}

\input{content/0_abstract}
\input{content/1_introduction}
%\input{content/4_related_works}
\input{content/2_background}

\input{content/3_equiformer_v2}
\input{content/5_experiments}
\input{content/6_conclusion}

\bibliographystyle{iclr2024_conference}
\bibliography{iclr2024_conference}

\newpage
\input{content/8_appendix}

\end{document}

%% file: imports.tex
\input{math_commands.tex}

\usepackage{times}
\usepackage{relsize}
\usepackage[T1]{fontenc}
\usepackage[scaled=0.90]{inconsolata}
\usepackage[utf8]{inputenc}
\usepackage[english]{babel}

\usepackage{epsfig}
\usepackage{graphicx}
\usepackage{wrapfig}
\usepackage[belowskip=0pt,aboveskip=0pt,font=small]{caption}
\usepackage[belowskip=0pt,aboveskip=0pt,font=small]{subcaption}
\usepackage{changepage}

\usepackage{amsfonts, amsmath, amsthm, amssymb, mathabx}
\usepackage{algorithm,algorithmicx,algpseudocode}
\usepackage{siunitx}
\usepackage{pifont}
\usepackage{nicefrac}
\usepackage{microtype}
\usepackage{textcomp}
\usepackage{stmaryrd}
\SetSymbolFont{stmry}{bold}{U}{stmry}{m}{n}
\usepackage{upgreek}
\usepackage{bm}
\usepackage{cases}
\usepackage{mathtools}
\usepackage{multirow}

\usepackage{rotating}
\usepackage{booktabs}
\usepackage{tabularx}

\usepackage{enumitem}
\usepackage[olditem,oldenum]{paralist}

\usepackage{alltt}
\usepackage{listings}

\usepackage{url}
\usepackage{xspace}
\usepackage{comment}
\usepackage{color}
\usepackage[table]{xcolor}
\usepackage{afterpage}
\usepackage{pdfpages}
\usepackage{framed}
\usepackage{fancybox}
\usepackage{cuted}

\usepackage{quoting}
\quotingsetup{vskip=0pt}
\usepackage{epigraph}

\definecolor{CustomRed}{RGB}{163, 31, 52}
\definecolor{crimson}{RGB}{163, 31, 52}
\definecolor{Belize}{RGB}{41,128,185}
\usepackage{hyperref}
\usepackage{url}

\hypersetup{
    colorlinks,
    linkcolor={crimson},
    citecolor={blue!50!black},
    urlcolor={blue!80!black}
}

\usepackage{lipsum}
\usepackage{xparse}
\usepackage{tikz}
\usepackage[normalem]{ulem}
\usepackage[toc,page,header]{appendix}
\usepackage{minitoc}

\usepackage[export]{adjustbox}

\usepackage[capitalize,nameinlink]{cleveref} % better than autoref; import last

\newcommand{\cmark}{\ding{51}}%
\newcommand{\xmark}{\ding{55}}%

\newcommand\mcc[1]{\multicolumn{1}{c}{#1}}  % centering a single table cell
\newlength\savewidth\newcommand\shline{\noalign{\global\savewidth\arrayrulewidth
  \global\arrayrulewidth 1pt}\hline\noalign{\global\arrayrulewidth\savewidth}}
\newcommand{\tablestyle}[2]{\setlength{\tabcolsep}{#1}\renewcommand{\arraystretch}{#2}\centering\footnotesize}
\definecolor{baselinecolor}{gray}{.9}
\newcommand{\baseline}[1]{\cellcolor{baselinecolor}{#1}}
\newcolumntype{x}[1]{>{\centering\arraybackslash}p{#1pt}}
\newcolumntype{y}[1]{>{\raggedright\arraybackslash}p{#1pt}}
\newcolumntype{z}[1]{>{\raggedleft\arraybackslash}p{#1pt}}

\newcommand*\circled[2][1.6]{\tikz[baseline=(char.base)]{
    \node[shape=circle, draw, inner sep=1pt,
        minimum height={\f@size*#1},] (char) {\vphantom{WAH1g}#2};}}

\input{mysymbols}

\usepackage{iclr2024_conference, times}

%% file: math_commands.tex
\usepackage{amsmath,amsfonts,bm}

\def\Figref#1{Figure~\ref{#1}}

\def\tableref#1{Table~\ref{#1}}

\def\threetablerefs#1#2#3{Tables~\ref{#1}, \ref{#2}, \ref{#3}}
\def\eqref#1{equation~\ref{#1}}
\def\1{\bm{1}}

\DeclareMathAlphabet{\mathsfit}{\encodingdefault}{\sfdefault}{m}{sl}
\SetMathAlphabet{\mathsfit}{bold}{\encodingdefault}{\sfdefault}{bx}{n}

%% file: mysymbols.tex
\usepackage{eso-pic}
\usepackage{xspace}
\makeatletter
\DeclareRobustCommand\onedot{\futurelet\@let@token\@onedot}
\def\@onedot{\ifx\@let@token.\else.\null\fi\xspace}

\def\eg{e.g\onedot}

\makeatother

\newcommand{\todo}[1]{\textcolor{red}{#1}}
\definecolor{crimson}{RGB}{163, 31, 52}

\newcommand{\revision}[1]{{#1}}

\newcommand{\ad}[1]{{\textcolor{magenta}{(AD: #1)}}}
\newcommand{\ade}[1]{{\textcolor{magenta}{#1}}}

\newcommand{\angstrom}{\textup{\AA} }

\newcommand{\linewidthbox}[1]{%
  \begingroup
    \sbox0{\ignorespaces#1\unskip}%
    \leavevmode
    \ifdim\wd0>\linewidth
      \hbox to\linewidth{%
        \hss\resizebox{\linewidth}{!}{\copy0 }\hss
      }%
    \else
      \copy0 %
    \fi
  \endgroup
}

\crefname{section}{Sec.}{Sec.}
\crefname{appendix}{App.}{App.}
\crefname{definition}{Def.}{Defs.}
\crefname{proposition}{Prop.}{Props.}

%% file: content/0_abstract.tex
%\vspace{-3mm}
\begin{abstract}
%\vspace{-3mm}
Equivariant Transformers such as Equiformer have demonstrated the efficacy of applying Transformers to the domain of 3D atomistic systems. 
%However, they are still limited to small degrees of equivariant representations due to their computational complexity.
However, they are limited to small degrees of equivariant representations due to their computational complexity.
%, and it is unclear whether the architectures can generalize well to higher degrees.
%In this paper, we investigate whether these architectures can scale well to higher degrees.
In this paper, we investigate whether these architectures can scale well to higher degrees.
% and propose an improved equivariant Transformer -- EquiformerV2.
% 
%Starting from Equiformer, we first replace tensor products with eSCN convolutions to efficiently incorporate higher-order tensors. 
Starting from Equiformer, we first replace $SO(3)$ convolutions with eSCN convolutions to efficiently incorporate higher-degree tensors. 
% 
%Then, to better leverage the power of higher degrees, we propose three architectural improvements -- attention re-normalization, separable $S^2$ activation and separable layer normalization.
Then, to better leverage the power of higher degrees, we propose three architectural improvements -- attention re-normalization, separable $S^2$ activation and separable layer normalization.
% 
%Putting it all together, we propose EquiformerV2,
%Putting this all together, we propose EquiformerV2,
% Experiments on the large-scale OC20 dataset verify the proposed architectural improvements, and EquiformerV2
%which outperforms previous state-of-the-art methods on the large-scale OC20 dataset by up to $15\%$ on forces, $5\%$ on energies, offers better speed-accuracy trade-offs, and $2\times$ reduction in DFT calculations needed for computing adsorption energies.
Putting this all together, we propose EquiformerV2, which outperforms previous state-of-the-art methods on large-scale OC20 dataset by up to $9\%$ on forces, $4\%$ on energies, offers better speed-accuracy trade-offs, and $2\times$ reduction in DFT calculations needed for computing adsorption energies.
\revision{Additionally, EquiformerV2 trained on only OC22 dataset outperforms GemNet-OC trained on both OC20 and OC22 datasets, achieving much better data efficiency.}
\revision{Finally, we compare EquiformerV2 with Equiformer on QM9 and OC20 S2EF-2M datasets to better understand the performance gain brought by higher degrees.}

%Besides, EquiformerV2 can transfer well to smaller QM9 and MD17 datasets, demonstrating the possibility of utilizing a unified architecture across a wide range of datasets.
%and achieve competitve results to previous methods.

%\todo{different from eSCN: normalization and regularization lead to training longer and generalizing better}

\end{abstract}

%% file: content/1_introduction.tex
%\vspace{-3mm}
\section{Introduction}
\label{sec:intro}

In recent years, machine learning (ML) models have shown promising results in accelerating and scaling high-accuracy but compute-intensive quantum mechanical calculations by effectively accounting for key features of atomic systems, such as the discrete nature of atoms, and Euclidean and permutation symmetries~\citep{neural_message_passing_quantum_chemistry,deep_potential_molecular_dynamics,push_limit_of_md_100m,dimenet_pp,nequip,deep_potential_molecular_dynamics_simulation,se3_wavefunction,gemnet_xl,quantum_scaling,adsorbml}.
% Machine learning (ML) models for estimating energies and forces of
% 3D atomistic systems have the potential to accelerate and scale predictions from computationally-expensive \textit{ab initio} quantum mechanical calculations~\cite{neural_message_passing_quantum_chemistry,deep_potential_molecular_dynamics,push_limit_of_md_100m,dimenet_pp,nequip,deep_potential_molecular_dynamics_simulation,se3_wavefunction,gemnet_xl,quantum_scaling,adsorbml}
% such as density functional theory (DFT)~\cite{dft_1, dft_2} \revision{, which is to predict atomic energies and forces}.
%By bringing down computational costs from hours or days to a few seconds, these methods enable new insights in many applications such as molecular simulations, material design and drug discovery.
By bringing down computational costs from hours or days to fractions of seconds, these methods enable new insights in many applications such as molecular simulations, material design and drug discovery.
A promising class of ML models that have enabled this progress is equivariant graph neural networks (GNNs)~\citep{tfn, 3dsteerable, kondor2018clebsch, se3_transformer, nequip, segnn, allergo, equiformer, escn}.

Equivariant GNNs treat 3D atomistic systems as graphs, and incorporate inductive biases such that their internal representations and predictions are equivariant to 3D translations, rotations and optionally inversions.
%Specifically, they build up equivariant embeddings of each node as
%geometric tensors of irreducible representations (or irreps), and have
%interactions or message passing between nodes based on equivariant operations
%such as tensor products.
Specifically, they build up equivariant features of each node as vector spaces of irreducible representations (or irreps) and have interactions or message passing between nodes based on equivariant operations such as tensor products.
% One of the most important examples is accelerating the calculations of density functional theory (DFT)~\cite{dft_1, dft_2}, which is to estimate atomic energies and forces.
% Although DFT is critical to many applications like drug discovery and material design, it is compute intensive and requires hours or days for a single DFT calculation.
% This gives opportunities to using machine learning, which can significantly reduce computational costs to seconds.
% Among many machine learning models, equivariant graph neural networks (GNNs) have demonstrated promising results~\cite{tfn, 3dsteerable, kondor2018clebsch, se3_transformer, nequip, segnn, allergo, equiformer, escn}.
% They treat 3D atomistic systems as graphs and incorporate 3D-related inductive biases, including equivariance to 3D translation, 3D rotation and optionally 3D inversion, into their networks for modeling the behaviors of atoms in 3D spaces.
% More concretely, they use vector spaces of irreducible representations (irreps) as equivariant feature representations and act on the equivariant features with equivariant operations such as tensor products.
% Since they consider 3D information without complicating graph structures, they can directly inherit the design of generic GNNs and even Transformers~\cite{transformer, transformer_in_vision}.
Recent works on equivariant Transformers, specifically Equiformer~\citep{equiformer}, have shown the efficacy of applying Transformers~\citep{transformer}, which have previously enjoyed widespread success in computer vision~\citep{detr, vit, deit}, language~\citep{bert,gpt3}, and graphs~\citep{generalization_transformer_graphs, spectral_attention, graphormer, graphormer_3d}, to this domain of 3D atomistic systems.
%\revision{and that equivariant Transformers can be more expressive than equivariant message passing networks like SEGNN~\cite{segnn}}.
% Since they consider 3D information without complicating graph structures,
% they can directly inherit the design of generic GNNs and even
% Transformers~\cite{transformer, transformer_in_vision}.
% Similar to the widespread success of generalizing Transformers to computer vision~\cite{detr, vit, deit} and graphs~\cite{generalization_transformer_graphs, spectral_attention, graphormer, graphormer_3d}, recent works on equivariant Transformers~\cite{se3_transformer, torchmd_net, eqgat, equiformer}, particularly Equiformer~\cite{equiformer}, have demonstrated the efficacy of applying Transformers to equivariant GNNs and 3D atomistic systems.
%Particularly, graph neural networks (GNNs), which naturally treat the set-like nature of collections of atoms and mimic the topological and geometric nature of atomic interactions, have gained increasing popularity due to their performance.
%, which is similar to convolutional neural networks (CNNs).
%Neural networks are top performers in existing atomistic benchmarks due to the ease with which inductive biases can be incorporated to exploit the symmetry of training data.
% One main factor leading to the success of neural networks is incorporating inductive biases to exploit the symmetry of training data.

A bottleneck in scaling Equiformer as well as other equivariant GNNs is the computational complexity of tensor products, especially when we increase the
maximum degree of irreps $L_{max}$.
% \ade{Equiformer} as well as other equivariant GNNs, does not scale well with the
% maximum degree of \ade{irreps} $L_{max}$.
%
This limits these models to use small values of $L_{max}$ (e.g., $L_{max} \leq$ 3),
which consequently limits their performance.
%Higher degrees can better capture angular resolution and directional
%information, critical for accurate prediction of atomic energies
%and forces.
Higher degrees can better capture angular resolution and directional information, which is critical to accurate prediction of atomic energies and forces~\citep{nequip, scn, escn}.
% small values of $L_{max}$ limit the performance of equivariant GNNs.
%
% tasks, which are sensitive to subtle changes in atomic positions, like predicting DFT force fields.
To this end,
eSCN~\citep{escn} recently proposes efficient convolutions to reduce $SO(3)$
tensor products to $SO(2)$ linear operations, bringing down the computational
cost from $\mathcal{O}(L_{max}^6)$ to $\mathcal{O}(L_{max}^3)$ and enabling scaling to larger values of $L_{max}$ (\eg, $L_{max}$ up to $8$).
%\todo{eSCN uses these efficient tensor products for higher $L_{max}$ within a simple SEGNN~\cite{segnn}-like message passing network design.
%Equiformer's attention-based message passing has been shown to be more performant than SEGNN.}
However, except using efficient convolutions for higher $L_{max}$, eSCN still follows SEGNN-like message passing network~\citep{segnn} design, and Equiformer has been shown to improve upon SEGNN.
More importantly, this ability to use higher $L_{max}$ challenges whether the previous design of equivariant Transformers can scale well to higher-degree representations.

%\todo{In this paper, we are interested in adapting this ability to use higher-degree representations in tensor products to equivariant Transformers.}
%In this paper, we are interested in marrying the benefits of eSCN convolutions, which enable scaling to higher $L_{max}$, with equivariant Transformers.
% In this paper, we are interested in marrying the benefits of eSCN convolutions with equivariant Transformers.
%\revision{The ability to use higher $L_{max}$ introduced recently by eSCN convolutions challenges whether the previous design of equivariant Transformers can scale well to higher-degree representations.}
%\revision{In this paper, we are interested in marrying the benefits of eSCN convolutions with equivariant Transformers.}
In this paper, we are interested in adapting eSCN convolutions for higher-degree representations to equivariant Transformers.
% Since eSCN convolutions allows higher degrees, in this paper, we investigate whether the design of previous equivariant Transformers can scale well to higher degrees.
%We start with Equiformer~\cite{equiformer} and replace tensor products with eSCN convolutions.
We start with Equiformer~\citep{equiformer} and replace $SO(3)$ convolutions with eSCN convolutions.
%
% \ade{But we find that this naive combination is prone to overfitting and generalizes poorly when scaled to higher $L_{max}$.}
% % ??? Equiformer has room for improvement.
% \ad{Provide more specific intuition. Did we see overfitting? Was training unstable?}
We find that naively incorporating eSCN convolutions does not result in better performance than the original eSCN model.
% (Index 1 versus ``eSCN baseline'' in~\tableref{tab:ablations}).
%
Therefore, to better leverage the power of higher degrees, we propose three architectural improvements -- attention re-normalization, separable $S^2$ activation and separable layer normalization.
%
%Putting it all together, we propose EquiformerV2, which is developed on the large and diverse OC20 dataset~\cite{oc20}.
Putting these all together, we propose EquiformerV2, which is developed on large and diverse OC20 dataset~\citep{oc20}.
%We start with ablations of EquiformerV2 on the OC20
%S2EF-2M dataset (Sec.~\ref{subsec:ablation_studies}), where
\ade{
    %we show that each change contributes to better performance when scaling to
    %higher degrees, and we study how EquiformerV2 scales with higher degrees,
    %orders, depths, and widths of the network.
    %
    %Next, in Sec.~\ref{subsec:main_results}, we show that EquiformerV2
}
Experiments on OC20 show that EquiformerV2
outperforms previous state-of-the-art methods with improvements of up to $9\%$ on forces and $4\%$ on energies, and offers better speed-accuracy trade-offs
compared to existing invariant and equivariant GNNs.
Additionally, when used in the AdsorbML algorithm~\citep{adsorbml} for performing
adsorption energy calculations, EquiformerV2
achieves the highest success rate and
$2\times$ reduction in DFT calculations to achieve comparable
adsorption energy accuracies as previous methods.
Furthermore, EquiformerV2 trained on only OC22~\citep{oc22} dataset outperforms GemNet-OC~\citep{gasteiger_gemnet_oc_2022} trained on both OC20 and OC22 datasets, achieving much better data efficiency.
Finally, we compare EquiformerV2 with Equiformer on QM9 dataset~\citep{qm9_1, qm9_2} and OC20 S2EF-2M dataset to better understand the performance gain of higher degrees and the improved architecture.

\definecolor{mygreen}{RGB}{64, 179, 79}
\begin{figure*}[t]
%\begin{adjustwidth}{-4.5cm}{-4.5cm}
\includegraphics[width=0.9\linewidth]{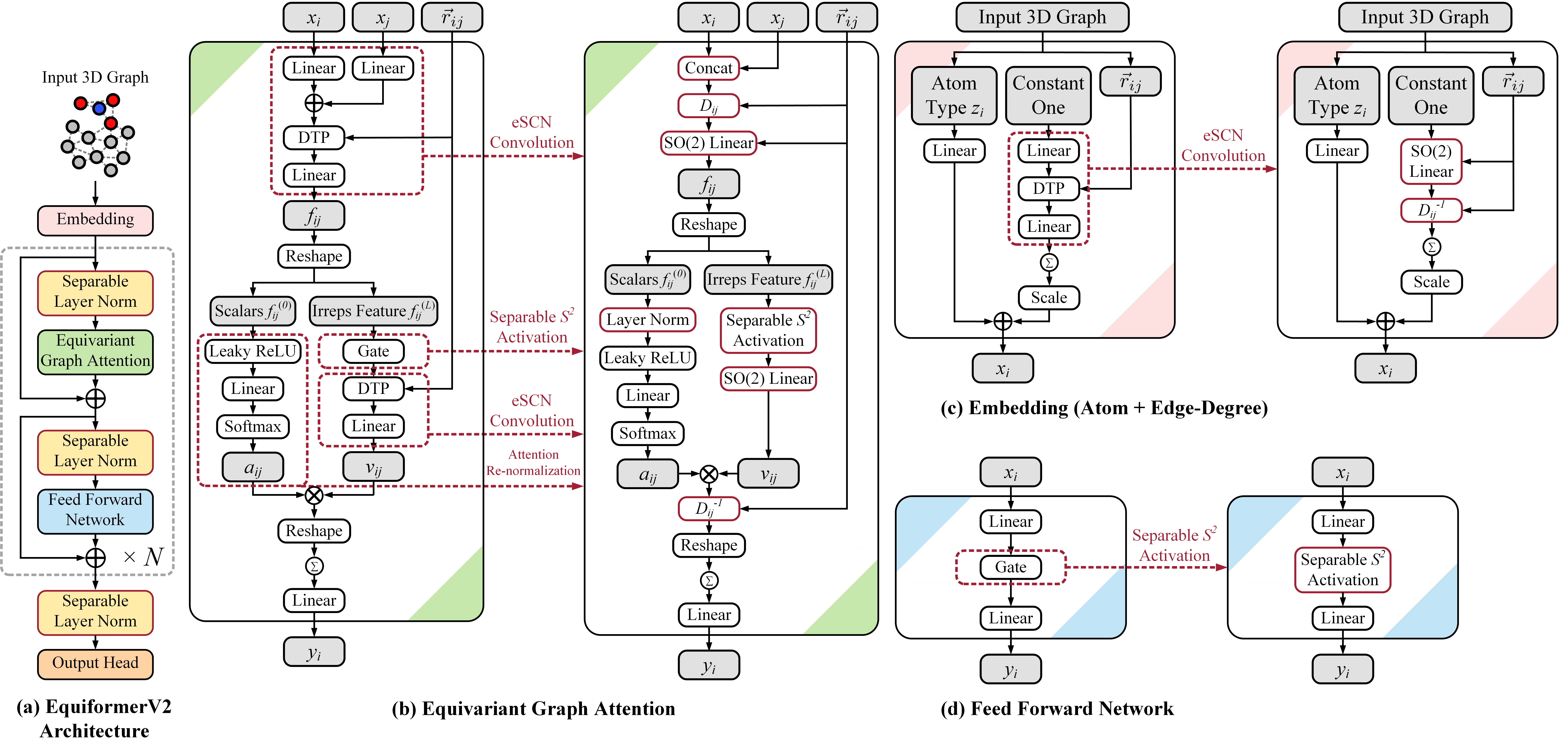}
\centering
%\end{adjustwidth}
%\vspace{0.5pt}
%\vspace{-1mm}
\caption{
%\textbf{Overview of EquiformerV2.}
Overview of EquiformerV2.
We highlight the differences from Equiformer~\citep{equiformer} in {\color{crimson}red}.
%For \textbf{(b)}, \textbf{(c)}, and \textbf{(d)}, the left figure is the original module in Equiformer, and the right figure is the revised module in EquiformerV2.
For (b), (c), and (d), the left figure is the original module in Equiformer, and the right figure is the revised module in EquiformerV2.
Input 3D graphs are embedded with atom and edge-degree embeddings and processed with Transformer blocks, which consist of equivariant graph attention and feed forward networks.
``$\otimes$'' denotes multiplication, ``$\oplus$'' denotes addition, and $\sum$ within a circle denotes summation over all neighbors.
``DTP'' denotes depth-wise tensor products used in Equiformer.
\colorbox{baselinecolor}{Gray} cells indicate intermediate irreps features.
}
%\vspace{-6mm}
\label{fig:equiformer_v2}
\end{figure*}

%% file: content/2_background.tex
%\vspace{-3mm}
\section{Background}
\label{sec:background}
%\vspace{-3mm}

We present background relevant to this work here and discuss related works in Sec.~\ref{sec:related_work}. 

%The proposed EquiformerV2 is built on top of Equiformer~\cite{equiformer} (Sec.~\ref{subsec:equiformer}) and eSCN convolutions~\cite{escn} for efficient tensor products (Sec.~\ref{subsec:so2_convolution}).
\input{content/2_1_e3_equivariant_neural_networks}
\input{content/2_2_equiformer}

\input{content/2_3_escn_convolution}

%% file: content/2_1_e3_equivariant_neural_networks.tex
%\vspace{-3mm}
\subsection{$SE(3)/E(3)$-Equivariant Neural Networks}
%\vspace{-2mm}
We discuss the relevant background of $SE(3)/E(3)$-equivariant neural networks here.
Please refer to Sec.~\ref{appendix:sec:additional_background} in appendix for more details of equivariance and group theory.
%Please refer to Sec. {\color{crimson}A} in appendix for
%more details of equivariance and group theory.
%\todo{define irreps before referring to them, specifically in terms of sh
%coefficients, no. of channels, and dimensionality of the tensor}

%\todo{why equivariance?}
Including equivariance in neural networks can serve as a strong prior knowledge, which can therefore improve data efficiency and generalization.
Equivariant neural networks use equivariant irreps features built from vector spaces of irreducible representations (irreps) to achieve equivariance to 3D rotation.
Specifically, the vector spaces are $(2L + 1)$-dimensional, where degree $L$ is a non-negative integer.
%$L$ can determine how fast vectors rotate, and higher $L$ are critical to tasks sensitive to angular information like predicting forces~\cite{nequip, scn, escn}.
%$L$ can be intuitively interpreted as the ``angular frequency'' of the vectors, i.e.,  how fast the vectors rotate with respect to a rotation of the coordinate system, and higher $L$ is critical to tasks sensitive to angular information like predicting forces~\cite{nequip, scn, escn}.
$L$ can be intuitively interpreted as the angular frequency of the vectors, i.e.,  how fast the vectors rotate with respect to a rotation of the coordinate system.
Higher $L$ is critical to tasks sensitive to angular information like predicting forces~\citep{nequip, scn, escn}.
Vectors of degree $L$ are referred to as type-$L$ vectors, and they are rotated with Wigner-D matrices $D^{(L)}$ when rotating coordinate systems.
Euclidean vectors $\vec{r}$ in $\mathbb{R}^3$ can be projected into type-$L$ vectors by using spherical harmonics $Y^{(L)}(\frac{\vec{r}}{|| \vec{r} ||})$.
We use order $m$ to index the elements of type-$L$ vectors, where $-L \leq m \leq L$.
We concatenate multiple type-$L$ vectors to form an  equivariant irreps feature $f$.
Concretely, $f$ has $C_{L}$ type-$L$ vectors, where $0 \leq L \leq L_{max}$ and $C_{L}$ is the number of channels for type-$L$ vectors.
In this work, we mainly consider $C_L = C$, and the size of $f$ is $(L_{max} + 1)^2 \times C$.
We index $f$ by channel $i$, degree $L$, and order $m$ and denote as $f^{(L)}_{m, i}$.

%\todo{Mention message passing}

Equivariant GNNs update irreps features by passing messages of transformed irreps features between nodes.
%Tensor products are necessary to interact different type-$L$ vectors during message passing.
To interact different type-$L$ vectors during message passing, we use tensor products, which generalize multiplication to equivariant irreps features.
Denoted as $\otimes_{L_1, L_2}^{L_3}$, the tensor product uses Clebsch-Gordan coefficients to combine type-$L_1$ vector $f^{(L_1)}$ and type-$L_2$ vector $g^{(L_2)}$ and produces type-$L_3$ vector $h^{(L_3)}$:
\begin{equation}
h^{(L_3)}_{m_3} = (f^{(L_1)} \otimes_{L_1, L_2}^{L_3} g^{(L_2)})_{m_3} = \sum_{m_1 = -L_1}^{L_1} \sum_{m_2 = -L_2}^{L_2} C^{(L_3, m_3)}_{(L_1, m_1)(L_2, m_2)} f^{(L_1)}_{m_1} g^{(L_2)}_{m_2}
\label{eq:tensor_product}
\end{equation}
where $m_1$ denotes order and refers to the $m_1$-th element of $f^{(L_1)}$.
Clebsch-Gordan coefficients $C^{(L_3, m_3)}_{(L_1, m_1)(L_2, m_2)}$ are non-zero only when $| L_1 - L_2 | \leq L_3 \leq | L_1 + L_2 |$ and thus restrict output vectors to be of certain degrees.
We typically discard vectors with $L > L_{max}$, where $L_{max}$ is a hyper-parameter, to prevent vectors of increasingly higher dimensions.
In many works, message passing is implemented as equivariant convolutions, which perform tensor products between input irreps features $x^{(L_1)}$ and spherical harmonics of relative position vectors $Y^{(L_2)}(\frac{\vec{r}}{|| \vec{r} ||})$.
%Specifically, the output irreps feature $y^{(L_o)}$ of degree $L_o$ is calculated as follows:
%\begin{equation}
%y^{(L_o)} = \sum_{L_i = 0}^{L_{max}} \sum_{L_f = 0}^{L_{max}} w_{L_i, L_f, L_o} \left( x^{(L_i)} \otimes^{L_o}_{L_i, L_f} Y^{(L_f)}\left( \frac{\vec{r}}{ \| \vec{r} \|} \right) \right)
%\label{eq:equivariant_convolution}
%\end{equation}

%% file: content/2_2_equiformer.tex
%\vspace{-3mm}
\subsection{Equiformer}
\label{subsec:equiformer}
%\vspace{-2mm}

Equiformer~\citep{equiformer} is an $SE(3)$/$E(3)$-equivariant GNN that combines
%the inductive biases of equivariance with the strength of Transformers~\cite{transformer, vit}.
the inductive biases of equivariance with the strength of Transformers.
%
%First, to incorporate equivariance, they replace scalars with vector spaces of irreducible representations, or \textit{irreps}, as internal feature representations.
First, Equiformer replaces scalar node features with equivariant irreps features to incorporate equivariance.
Next, it performs equivariant operations on these irreps features and
equivariant graph attention for message passing.
These operations include tensor products and
%The resulting equivariant irreps features contain vectors of different degrees $L$, with higher $L$ corresponding to higher angular frequencies.
% operations, which include tensor products
% and the equivariant version of operations in typical Transformers.
% The latter consists of
equivariant linear operations, equivariant layer normalization~\citep{layer_norm}
and gate activation~\citep{3dsteerable}.
For stronger expressivity in the attention compared to typical Transformers,
Equiformer uses non-linear functions for both attention weights and message
passing.
% to improve the expressivity of attention in typical Transformers.
% Second, they propose equivariant graph attention, which leverages
% non-linear functions for both attention weights and message passing to improve the expressivity of attention in typical Transformers.
%
Additionally, Equiformer
% since Equiformer follows the design of Transformers,
% it can directly
incorporates regularization techniques commonly used by Transformers, e.g., dropout~\citep{dropout} to attention weights~\citep{gat} and stochastic depth~\citep{drop_path} to
the outputs of equivariant graph attention and feed forward networks.
%
%Please refer to the Equiformer paper~\cite{equiformer} for more details.

%% file: content/2_3_escn_convolution.tex
%\vspace{-3mm}
\subsection{eSCN Convolution}
\label{subsec:so2_convolution}
eSCN convolutions~\citep{escn} are proposed to use $SO(2)$ linear operations for efficient tensor products.
We provide an outline and intuition for their method here, please refer to Sec.~\ref{appendix:subsec:escn_convolution} and their work~\citep{escn} for mathematical details.
%We provide an outline and intuition for their method here, and please refer to Sec. {\color{crimson}A} and their work~\cite{escn} for mathematical details.

%A traditional $SO(3)$ tensor product involves spherical harmonic projections of relative positions $Y^{(L_f)}_{m_f}(\vec{r_{ij}})$.
%A traditional $SO(3)$ tensor product interacts input irreps features $x^{(L_i)}_{m_i}$ and spherical harmonic projections of relative positions $Y^{(L_f)}_{m_f}(\vec{r_{ij}})$ with Clebsch-Gordan coefficients $C^{(L_o, m_o)}_{(L_i, m_i), (L_f, m_f)}$.
A traditional $SO(3)$ convolution interacts input irreps features $x^{(L_i)}_{m_i}$ and spherical harmonic projections of relative positions $Y^{(L_f)}_{m_f}(\vec{r_{ij}})$ with an $SO(3)$ tensor product with Clebsch-Gordan coefficients $C^{(L_o, m_o)}_{(L_i, m_i), (L_f, m_f)}$.
The projection $Y^{(L_f)}_{m_f}(\vec{r_{ij}})$ becomes sparse if we rotate the relative position vector $\vec{r}_{ij}$ with a rotation matrix $D_{ij}$ to align with the direction of $L = 0$ and $m = 0$, which corresponds to the z axis traditionally but the y axis in the conventions of $\texttt{e3nn}$~\citep{e3nn}.
Concretely, given $D_{ij} \vec{r}_{ij}$ aligned with the y axis, $Y^{(L_f)}_{m_f}( D_{ij} \vec{r}_{ij}) \neq 0$ only for $m_f = 0$.
%The SO(3) tensor product operation in the SO(3) convolution interacts the input $x_{l_i, m_i}$ and filter proportional to $Y_{l_f m_f}(\vec{r_{st}})$ using Clebsch-Gordan coefficients C.
If we consider only $m_f = 0$, $C^{(L_o, m_o)}_{(L_i, m_i), (L_f, m_f)}$ can be simplified, and $C^{(L_o, m_o)}_{(L_i, m_i), (L_f, 0)} \neq 0$ only when $m_{i} = \pm m_{o}$.
Therefore, the original expression depending on $m_i$, $m_f$, and $m_o$ is now reduced to only depend on $m_o$.
This means we are no longer mixing all integer values of $m_i$ and $m_f$, and  outputs of order $m_o$ are linear combinations of inputs of order $\pm m_o$. eSCN convolutions go one step further and replace the remaining non-trivial paths of the $SO(3)$ tensor product with an $SO(2)$ linear operation to allow for additional parameters of interaction between $\pm m_o$ without breaking equivariance.
%To summarize, eSCN convolutions achieve efficient tensor products by first rotating irreps features based on relative position vectors and then performing $SO(2)$ linear operations on the rotated features.
%%%To summarize, eSCN convolutions achieve efficient equivariant convolutions by first rotating irreps features based on relative position vectors and then performing $SO(2)$ linear operations on the rotated features.
%%%The key idea is that the rotation sparsifies tensor products and simplifies the computation. 

%During message passing, they first align the primary axis (y axis) of node embeddings $x_s$ with edge directions $\hat{r}_{ts}$. 
%The alignment sparsifies tensor products and decreases the number of degrees of rotational freedom from two to one, which enables reducing $SO(3)$ tensor products to $SO(2)$ linear operations.

%% file: content/3_equiformer_v2.tex
%\vspace{-3mm}
\section{EquiformerV2}
\label{sec:equiformer_v2}
%\vspace{-3mm}

%Starting from Equiformer~\cite{equiformer}, we first use eSCN convolutions to scale up degrees of representations (Sec.~\ref{subsec:incorporating_escn_convolutions_for_efficient_tensor_products_and_higher_degrees}).
Starting from Equiformer~\citep{equiformer}, we first use eSCN convolutions to scale to higher-degree representations (Sec.~\ref{subsec:incorporating_escn_convolutions_for_efficient_tensor_products_and_higher_degrees}).
%Then, to better leverage the power of high degrees, we propose three architectural improvements -- attention re-normalization (Sec.~\ref{subsec:attention_renormalization}), separable $S^2$ activation (Sec.~\ref{subsec:separable_s2_activation}) and separable layer normalization (Sec.~\ref{subsec:separable_layer_normalization}).
Then, we propose three architectural improvements, which yield further performance gain when using higher degrees: attention re-normalization (Sec.~\ref{subsec:attention_renormalization}), separable $S^2$ activation (Sec.~\ref{subsec:separable_s2_activation}) and separable layer normalization (Sec.~\ref{subsec:separable_layer_normalization}).
Figure~\ref{fig:equiformer_v2} illustrates the overall architecture of EquiformerV2 and the differences from Equiformer.

\input{content/3_1_incorporating_escn_convolutions_for_efficient_tensor_products_and_higher_degrees}

\input{content/3_2_attention_renormalization}
\input{content/3_3_separable_s2_activation}
\input{content/3_4_separable_layer_normalization}
\input{content/3_5_overall_architecture}

%% file: content/3_1_incorporating_escn_convolutions_for_efficient_tensor_products_and_higher_degrees.tex
%\vspace{-3mm}
%\subsection{Incorporating eSCN Convolutions for Efficient Tensor Products and Higher Degrees}
\subsection{Incorporating eSCN Convolutions for Higher Degrees}
\label{subsec:incorporating_escn_convolutions_for_efficient_tensor_products_and_higher_degrees}
%\vspace{-2mm}

%Tensor products are used to compute features sent from one node to another during message passing.
%and have the computational complexity $O(L_{max}^6)$, where $L_{max}$ is the maximum degrees of irreps feature representations. 
%The high complexity prevents Equiformer from using high $L_{max}$.
%The computational complexity of $SO(3)$ tensor products used in traditional $SO(3)$ convolutions during equivariant message passing scale unfavorably with $L_{max}$. 
The computational complexity of $SO(3)$ tensor products used in traditional $SO(3)$ convolutions scale unfavorably with $L_{max}$. 
%Because of this, it is impractical for Equiformer to use beyond $L_{max} = 1$ for large-scale datasets like OC20~\citep{oc20} and beyond $L_{max}=3$ for small-scale datasets like MD17~\citep{md17_1, md17_2, md17_3}.
Because of this, it is impractical for Equiformer to use $L_{max} > 2$ for large-scale datasets like OC20.
%Therefore, it is impractical for Equiformer to use beyond $L_{max} = 1$ for large-scale datasets like OC20~\citep{oc20} and beyond $L_{max}=3$ for small-scale datasets like MD17~\citep{md17_1, md17_2, md17_3}.
%However, the high complexity prevents Equiformer from using high values for maximum degrees of irreps features $L_{max}$.
%Specifically, Equiformer can only use $L_{max} = 1$ for large-scale OC20 dataset~\cite{oc20} and $L_{max} = 3$ for small-scale MD17 dataset~\cite{md17_1, md17_2, md17_3}.
Since higher $L_{max}$ can better capture angular information and are correlated with model expressivity~\citep{nequip}, low values of $L_{max}$ can lead to limited performance on certain tasks such as predicting forces.
Therefore, we replace original tensor products with eSCN convolutions for efficient tensor products, enabling Equiformer to scale up $L_{max}$ to  $6 \text{ or } 8$ on OC20 dataset.
Equiformer uses equivariant graph attention for message passing.
The attention consists of depth-wise tensor products, which mix information across different degrees, and linear layers, which mix information between channels of the same degree.
Since eSCN convolutions mix information across both degrees and channels, we replace the $SO(3)$ convolution, which involves one depth-wise tensor product layer and one linear layer, with a single eSCN convolutional layer, which consists of a rotation matrix $D_{ij}$ and an $SO(2)$ linear layer as shown in Figure~\ref{fig:equiformer_v2}b. 
%This is analogous to replacing MBConv with Fused-MBConv as in EfficientNetV2~\cite{efficientnetv2}.

%% file: content/3_2_attention_renormalization.tex
%\vspace{-3mm}
\subsection{Attention Re-normalization}
\label{subsec:attention_renormalization}
%\vspace{-2mm}
Equivariant graph attention in Equiformer uses tensor products to project node embeddings $x_i$ and $x_j$, which contain vectors of degrees from $0$ to $L_{max}$, to scalar features $f_{ij}^{(0)}$ and applies non-linear functions to $f_{ij}^{(0)}$ for attention weights $a_{ij}$. 
We propose attention re-normalization and introduce one additional layer normalization (LN)~\citep{layer_norm} before non-linear functions.
Specifically, given $f_{ij}^{(0)}$, we first apply LN and then use one leaky ReLU layer and one linear layer to calculate $z_{ij} = w_{a} ^\top \text{LeakyReLU}(\text{LN}(f_{ij}^{(0)}))$ and $a_{ij} = \text{softmax}_{j}(z_{ij}) = \frac{\text{exp}(z_{ij})}{\sum_{k \in \mathcal{N}(i)} \text{exp}(z_{ik})}$, where $w_{a}$ is a learnable vector of the same dimension as $f_{ij}^{(0)}$.
\revision{The motivation is similar to ViT-22B~\citep{vit_22b}, where they find that they need an additional layer normalization to stabilize training when increasing widths.}
\revision{When we scale up $L_{max}$, we effectively increase the number of input channels for calculating $f_{ij}^{(0)}$.}
\revision{By normalizing $f_{ij}^{(0)}$, the  additional LN can make sure the inputs to subsequent non-linear functions and softmax operations still lie within the same range as lower $L_{max}$ is used.}
\revision{This empirically improves the performance as shown in Table~\ref{tab:ablations}.}

%The node embeddings $x_i$ and $x_j$ are obtained by applying equivariant layer normalization to previous outputs.
%We note that vectors of different degrees in $x_i$ and $x_j$ are normalized independently, and therefore when they are projected to the same degree, the resulting $f_{ij}^{(0)}$ can be less well-normalized.
%This issue can be more significant as we scale up maximum degree $L_{max}$ and vectors of more degrees are normalized separately.
%To address the issue, we propose attention re-normalization and introduce one additional layer normalization (LN)~\cite{layer_norm} before non-linear functions.
%Specifically, given $f_{ij}^{(0)}$, we first apply LN and then use one leaky ReLU layer and one linear layer to calculate $z_{ij} = w_{a} ^\top \text{LeakyReLU}(\text{LN}(f_{ij}^{(0)}))$ and $a_{ij} = \text{softmax}_{j}(z_{ij}) = \frac{\text{exp}(z_{ij})}{\sum_{k \in \mathcal{N}(i)} \text{exp}(z_{ik})}$, where $w_{a}$ is a learnable vector of the same dimension as $f_{ij}^{(0)}$.
%\begin{equation}
%z_{ij} = a^\top \text{LeakyReLU}(\text{LN}(f_{ij}^{(0)}))
%\quad\text{and}\quad
%a_{ij} = \text{softmax}_{j}(z_{ij}) = \frac{\text{exp}(z_{ij})}{\sum_{k \in \mathcal{N}(i)} \text{exp}(z_{ik})}
%\label{eq:attention_renormalization}
%\end{equation}
%where $a$ is learnable and of the same dimension as $f_{ij}^{(0)}$ and $z_{ij}$ is a single scalar.

%% file: content/3_3_separable_s2_activation.tex
%\vspace{-3mm}
\subsection{Separable $S^2$ Activation}
\label{subsec:separable_s2_activation}
%\vspace{-2mm}

\begin{figure*}[t]
    \centering
    \begin{minipage}[b]{0.50\textwidth}
        %\begin{subfigure}{\textwidth}
        \centering
        \includegraphics[width=\textwidth]{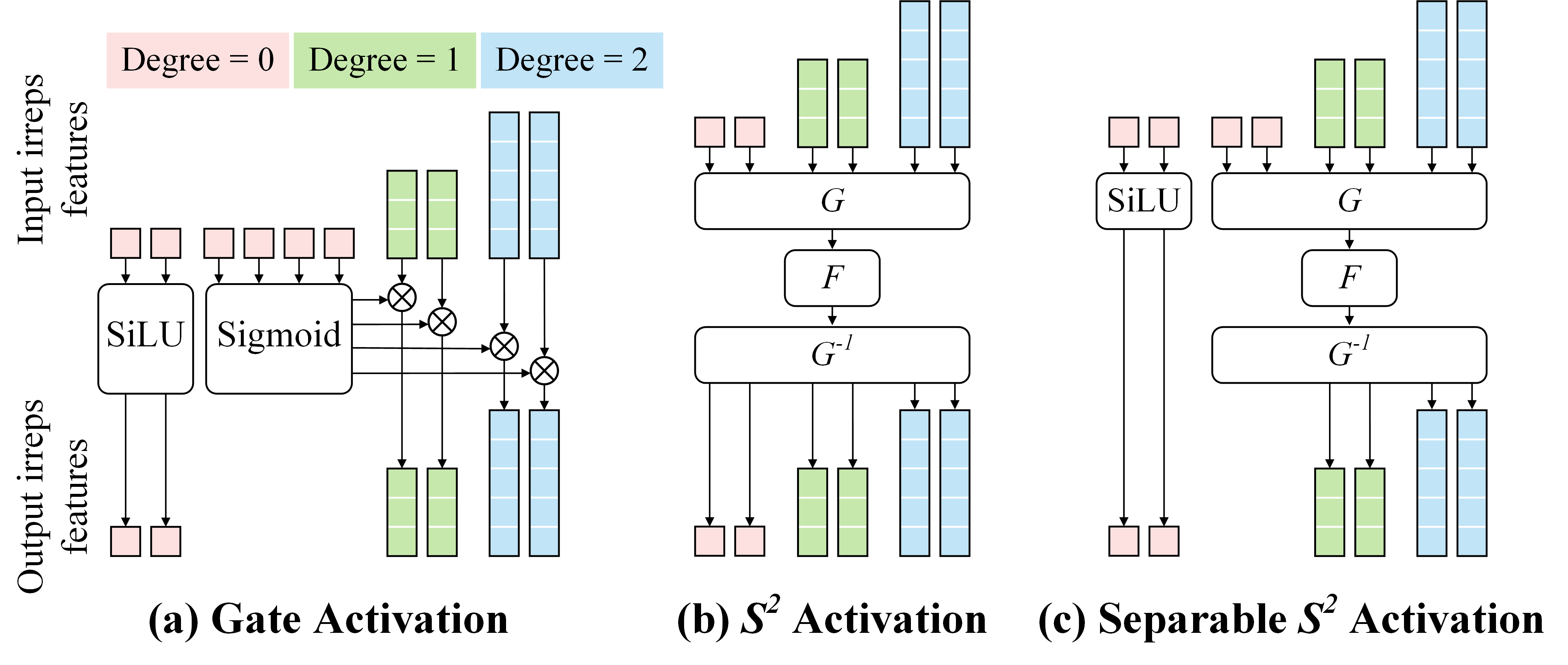}
        %{\scriptsize \textbf{(a)}}
        %\end{subfigure}
        %\vspace{0.5pt}
        \vspace{-3mm}
        \captionof{figure}{
        %\textbf{Different activation functions.} 
        Illustration of different activation functions.
        $G$ denotes conversion from vectors to point samples on a sphere, $F$ can typically be a SiLU activation or MLPs, and $G^{-1}$ is the inverse of $G$.}
        \label{fig:activation}
    \end{minipage}
    \hfill
    \begin{minipage}[b]{0.465\textwidth}
        %\begin{subfigure}{\textwidth}
        \centering
        \includegraphics[width=\textwidth]{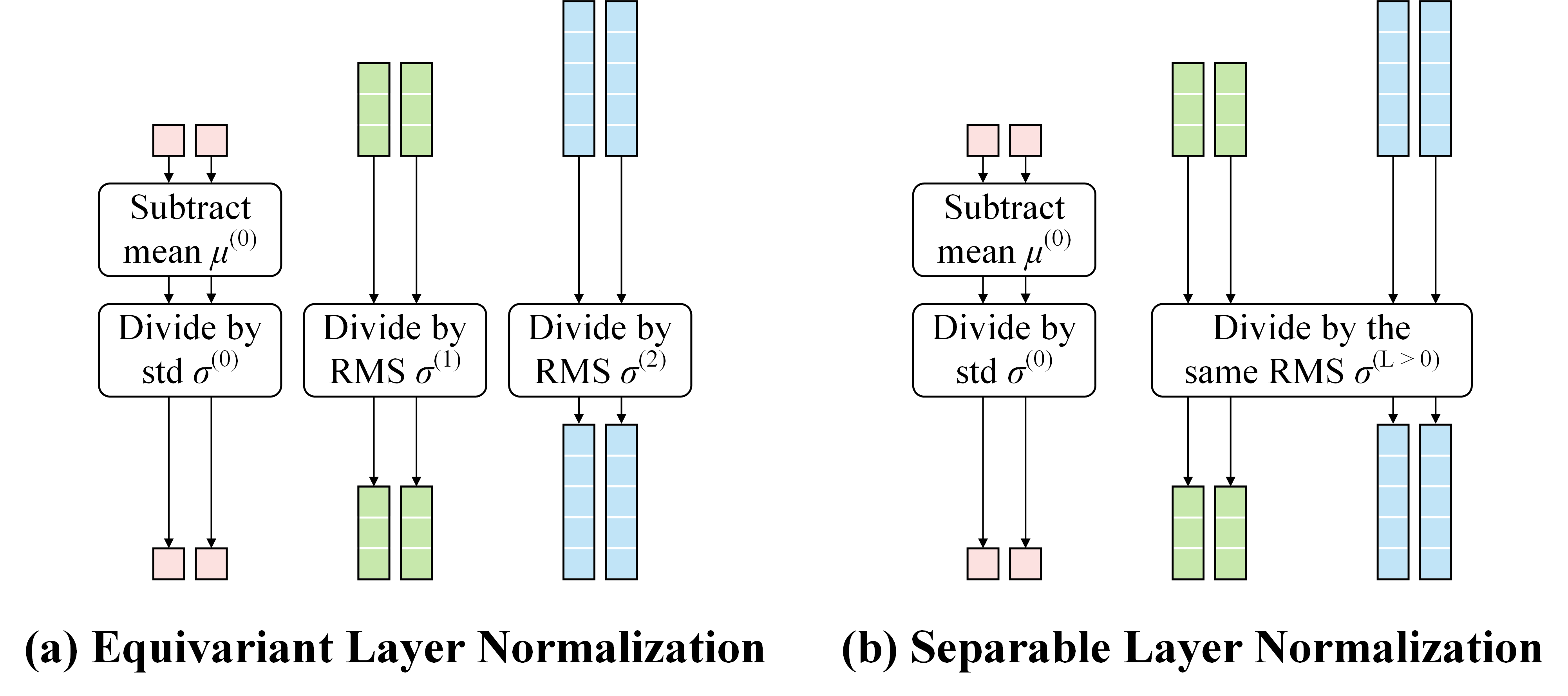}
        %\vspace{0.5pt}
        \vspace{-3mm}
        \captionof{figure}{
        %\textbf{Different layer normalization schemes.} 
        Illustration of how statistics are calculated in different normalizations.
        ``std'' denotes standard deviation, and ``RMS'' denotes root mean square.\\}
        \label{fig:normalization}
    \end{minipage}
    %\vspace{5pt}
%\vspace{-5mm}
\end{figure*}

The gate activation~\citep{3dsteerable} used by Equiformer applies sigmoid activation to scalar features to obtain non-linear weights and then multiply irreps features of degree $> 0$ with non-linear weights to add non-linearity to equivariant features.
%The activation, however, only accounts for the interaction from vectors of degree $0$ to those of degree $> 0$ and could be sub-optimal when we scale up $L_{max}$.
The activation only accounts for the interaction from vectors of degree $0$ to those of degree $> 0$ and can be sub-optimal when we scale up $L_{max}$.
%
%To better mix the information across degrees, SCN~\citep{scn} and eSCN propose to use $S^2$ activation~\cite{spherical_cnn}.
To better mix the information across degrees, SCN~\citep{scn} and eSCN adopt $S^2$ activation~\citep{spherical_cnn}.
% Given irreps feature $x \in \mathbb{R} ^{N \times (L_{max} + 1)^{2} \times C}$ of maximum degree $L_{max}$ and $C$ channels, t
The activation first converts vectors of all degrees to point samples on a sphere for each channel, applies unconstrained functions $F$ to those samples, and finally convert them back to vectors.
Specifically, given an input irreps feature $x \in \mathbb{R} ^{(L_{max} + 1)^{2} \times C}$, the output is $y = G^{-1}(F (G (x)))$,
%\begin{equation}
%y = G^{-1}(F (G (x)))
%\label{eq:separable_s2_activation}
%\end{equation}
%where $x \in \mathbb{R} ^{(L_{max} + 1)^{2} \times C}$ is an input irreps feature of maximum degree $L_{max}$ and $C$ channels, $G$ denotes the conversion from vectors to point samples on a sphere, $F$ can be typical SiLU activation~\cite{silu, swish} or typical MLPs, and $G^{-1}$ is the inverse of $G$.
where $G$ denotes the conversion from vectors to point samples on a sphere, $F$ can be typical SiLU activation~\citep{silu, swish} or typical MLPs, and $G^{-1}$ is the inverse of $G$.
We provide more details of $S^2$ activation in Sec.~\ref{appendix:subsec:s2_activation}.

%While $S^2$ activation can better mix vectors of different degrees, we find that directly replacing the gate activation with $S^2$ activation results in training instability (Index 3 in~\tableref{tab:ablations}).
However, we find that directly replacing the gate activation with $S^2$ activation in Equiformer results in large gradients and training instability (Index 3 in~\tableref{tab:ablations}).
To address the issue, we propose separable $S^2$ activation, which separates activation for vectors of degree $0$ and those of degree $> 0$.
Similar to gate activation, we have more channels for vectors of degree $0$.
As shown in Figure~\ref{fig:activation}c, we apply a SiLU activation to the first part of vectors of degree $0$, and the second part of vectors of degree $0$ are used for $S^2$ activation along with vectors of higher degrees.
After $S^2$ activation, we concatenate the first part of vectors of degree $0$ with vectors of degrees $> 0$ as the final output and ignore the second part of vectors of degree $0$.
\revision{Separating the activation for vectors of degree $0$ and those of degree $> 0$ prevents large gradients, enabling using more expressive $S^2$ activation for better performance.}
Additionally, we use separable $S^2$ activation in feed forward networks (FFNs).
Figure~\ref{fig:activation} illustrates the differences between gate activation,
$S^2$ activation and separable $S^2$ activation.
%, and Figure \todo{[TODO]} compares FFNs with different activation functions.

%% file: content/3_4_separable_layer_normalization.tex
%\vspace{-5mm}
\subsection{Separable Layer Normalization}
\label{subsec:separable_layer_normalization}
%\vspace{-2mm}
%As mentioned in Sec.~\ref{subsec:attention_renormalization}, equivariant layer normalization used by Equiformer normalizes vectors of different degrees independently, and when those vectors are projected to the same degree, the projected vectors can be less well-normalized.
\revision{Equivariant layer normalization used by Equiformer normalizes vectors of different degrees independently.}
\revision{However, it potentially ignores the relative importance of different degrees since the relative magnitudes between different degrees become the same after the normalization.} 
Therefore, instead of performing normalization to each degree independently, motivated by the separable $S^2$ activation mentioned above, we propose separable layer normalization (SLN), which separates normalization for vectors of degree $0$ and those of degrees $> 0$.
Mathematically, let $x \in \mathbb{R} ^{(L_{max} + 1)^{2} \times C}$ denote an input irreps feature of maximum degree $L_{max}$ and $C$ channels, and $x^{(L)}_{m, i}$ denote the $L$-th degree, $m$-th order and $i$-th channel of $x$.
SLN calculates the output $y$ as follows.
\begin{comment}
\begin{equation}
\begin{aligned}
\text{for}\quad L = 0, \quad &  y^{(0)} = \gamma^{(0)} \circ \left( \frac{x^{(0)} - \mu^{(0)}}{\sigma^{(0)}} \right) + \beta^{(0)}
\\
& \text{where}\quad
\mu^{(0)} = \frac{1}{C} \sum_{i = 1}^{C}x^{(0)}_{0, i}
\quad\text{and}\quad
\sigma^{(0)} = \sqrt{\frac{1}{C}\sum_{i = 1}^{C}(x^{(0)}_{0, i} - \mu^{(0)})^2}
\label{eq:separable_layer_norm_l0}
\end{aligned}
\end{equation}
\vspace{-3mm}
\begin{equation}
\begin{aligned}
\text{for}\quad L > 0, \quad &  y^{(L)} = \gamma^{(L)} \circ \left( \frac{x^{(L)}}{\sigma^{(L > 0)}} \right)
\quad\text{where}\quad
\sigma^{(L > 0)} = \sqrt{ \frac{1}{L_{max}} \sum_{L = 1}^{L_{max}}\left( \sigma^{(L)}\right)^2 } \\
& \quad\text{and}\quad
\sigma^{(L)} = \sqrt{ \frac{1}{C}\sum_{i = 1}^{C}\frac{1}{2L + 1} \sum_{m = -L}^{L} \left( x_{m, i}^{(L)} \right)^2 }
\label{eq:separable_layer_norm_l}
\end{aligned}
\end{equation}
\end{comment}
%
For $L = 0$, $y^{(0)} = \gamma^{(0)} \circ \left( \frac{x^{(0)} - \mu^{(0)}}{\sigma^{(0)}} \right) + \beta^{(0)}$, where $\mu^{(0)} = \frac{1}{C} \sum_{i = 1}^{C}x^{(0)}_{0, i}$ and $\sigma^{(0)} = \sqrt{\frac{1}{C}\sum_{i = 1}^{C}(x^{(0)}_{0, i} - \mu^{(0)})^2}$.
For $L > 0$, $y^{(L)} = \gamma^{(L)} \circ \left( \frac{x^{(L)}}{\sigma^{(L > 0)}} \right)$, where $\sigma^{(L > 0)} = \sqrt{ \frac{1}{L_{max}} \sum_{L = 1}^{L_{max}}\left( \sigma^{(L)}\right)^2 }$ and $\sigma^{(L)} = \sqrt{ \frac{1}{C}\sum_{i = 1}^{C}\frac{1}{2L + 1} \sum_{m = -L}^{L} \left( x_{m, i}^{(L)} \right)^2 }$.
$\gamma^{(0)}, \gamma^{(L)}, \beta^{(0)} \in \mathbb{R}^{C}$ are learnable parameters, $\mu^{(0)}$ and $\sigma^{(0)}$ are mean and standard deviation of vectors of degree $0$, $\sigma^{(L)}$ and $\sigma^{(L>0)}$ are root mean square values (RMS), and $\circ$ denotes element-wise product.
%The computation of $y^{(0)}$corresponds to typical layer normalization.
%We note that the difference between equivariant layer normalization and SLN lies only in $y^{(L)}$ with $L > 0$ and that equivariant layer normalization divides $x^{(L)}$ by $\sigma^{(L)}$, which is calculated independently for each degree $L$, instead of $\sigma^{(L > 0)}$, which considers all degrees $L > 0$.
Figure~\ref{fig:normalization} compares how $\mu^{(0)}$, $\sigma^{(0)}$, $\sigma^{(L)}$ and $\sigma^{(L>0)}$ are calculated in equivariant layer normalization and SLN.
\revision{Preserving the relative magnitudes between degrees $> 0$ improves performance as shown in Table~\ref{tab:ablations}.}

% \begin{figure*}[t]
% \includegraphics[width=0.65\linewidth]{figures/normalization.png}
% \centering
% \caption{\textbf{Comparison of how statistics are calculated in different normalizations.}
% ``std'' denotes standard deviation, and ``RMS'' denotes root mean square.
% }
% \vspace{-4mm}
% \label{fig:normalization}
% \end{figure*}

%% file: content/3_5_overall_architecture.tex
%\vspace{-3mm}
\subsection{Overall Architecture}
\label{subsec:overall_architecture}
%\vspace{-2mm}

%Here, we discuss all the other modules in EquiformerV2 and focus on the differences from Equiformer.

%\vspace{-3mm}
\paragraph{Equivariant Graph Attention.}
Figure~\ref{fig:equiformer_v2}b illustrates equivariant graph attention after the above modifications.
%As described in Sec.~\ref{subsec:incorporating_escn_convolutions_for_efficient_tensor_products_and_higher_degrees}, given node embeddings $x_i$ and $x_j$, we first concatenate them along the channel dimension and then rotate them with rotation matrices $D_{ij}$ based on their relative positions or edge directions $\vec{r}_{ij}$.
Given node embeddings $x_i$ and $x_j$, we first concatenate them along the channel dimension and then rotate them with rotation matrices $D_{ij}$ based on their relative positions or edge directions $\vec{r}_{ij}$.
%The rotation enables reducing $SO(3)$ tensor products to $SO(2)$ linear operations, and we replace depth-wise tensor products and linear layers between $x_i$, $x_j$ and $f_{ij}$ with a single $SO(2)$ linear layer.
We replace depth-wise tensor products and linear layers between $x_i$, $x_j$ and $f_{ij}$ with a single $SO(2)$ linear layer.
To consider the information of relative distances $|| \vec{r}_{ij} ||$, we transform $|| \vec{r}_{ij} ||$ with a radial function to obtain edge distance embeddings and then multiply the edge distance embeddings with concatenated node embeddings before the first $SO(2)$ linear layer.
We split the outputs $f_{ij}$ of the first $SO(2)$ linear layer into two parts.
The first part is scalar features $f_{ij} ^{(0)}$, which only contains vectors of degree $0$, and the second part is irreps features $f_{ij}^{(L)}$ and includes vectors of all degrees up to $L_{max}$.
%Same as Equiformer, we use one leaky ReLU layer, one linear layer and a final softmax layer to transform $f_{ij}^{(0)}$ into attention weights $a_{ij}$.
As mentioned in Sec.~\ref{subsec:attention_renormalization}, we first apply an additional LN to $f_{ij}^{(0)}$ and then follow the design of Equiformer by applying one leaky ReLU layer, one linear layer and a final softmax layer to obtain attention weights $a_{ij}$.
As for value $v_{ij}$, we replace the gate activation with separable $S^2$ activation with $F$ being a single SiLU activation and then apply the second $SO(2)$ linear layer.
While in Equiformer, the message $m_{ij}$ sent from node $j$ to node $i$ is $m_{ij} = a_{ij} \times v_{ij}$, here we need to rotate $a_{ij} \times v_{ij}$ back to original coordinate frames and the message $m_{ij}$ becomes $D_{ij}^{-1} (a_{ij} \times v_{ij})$.
Finally, we can perform $h$ parallel equivariant graph attention functions given $f_{ij}$.
The $h$ different outputs are concatenated and projected with a linear layer to become the final output $y_i$.
Parallelizing attention functions and concatenating can be implemented with ``Reshape''.

%\vspace{-3mm}
\paragraph{Feed Forward Network.}
%As illustrated in Figure~\ref{fig:equiformer_v2}d, we replace the gate activation with separable $S^2$ activation.
As shown in Figure~\ref{fig:equiformer_v2}d, we replace the gate activation with separable $S^2$ activation.
The function $F$ consists of a two-layer MLP, with each linear layer followed by SiLU, and a final linear layer.

%\vspace{-3mm}
\paragraph{Embedding.}
This module consists of atom embedding and edge-degree embedding.
The former is the same as that in Equiformer.
For the latter, as depicted in the right branch in  Figure~\ref{fig:equiformer_v2}c, we replace original linear layers and depth-wise tensor products with a single $SO(2)$ linear layer followed by a rotation matrix $D_{ij} ^{-1}$.
Similar to equivariant graph attention, we consider the information of relative distances by multiplying the outputs of the $SO(2)$ linear layer with edge distance embeddings.

%\vspace{-3mm}
\paragraph{Radial Basis and Radial Function.}
We represent relative distances $ || \vec{r}_{ij} || $ with a finite radial basis like Gaussian radial basis functions~\citep{schnet} to capture their subtle changes.
We transform radial basis with a learnable radial function to generate edge distance embeddings.
The function consists of a two-layer MLP, with each linear layer followed by LN and SiLU, and a final linear layer.

%\vspace{-3mm}
\paragraph{Output Head.}
%To predict scalar quantities like energy, we use one feed forward network to transform irreps features on each node into a scalar and then perform sum aggregation over all nodes.
To predict scalar quantities like energy, we use a feed forward network to transform irreps features on each node into a scalar and then sum over all nodes.
%As for predicting forces acting on each node, we use a block of equivariant graph attention and treat the output of degree $1$ as our predictions.
For predicting atom-wise forces, we use a block of equivariant graph attention and treat the output of degree $1$ as our predictions.

%% file: content/5_experiments.tex
%\vspace{-8mm}
\section{Experiments}
\label{sec:experiments}

\input{content/5_1_oc20}

\input{content/5_2_oc22}
\input{content/5_3_comparison_with_equiformer}

%% file: content/5_1_oc20.tex
%\vspace{-3mm}
\subsection{OC20 Dataset}
%\vspace{-3mm}

Our main experiments focus on the large and diverse OC20 dataset~\citep{oc20}.
Please refer to Sec.~\ref{appendix:subsec:detailed_description_of_oc20_dataset} for details on the dataset.
We first conduct ablation studies on EquiformerV2 trained on the 2M split of OC20 S2EF dataset (Sec.~\ref{subsec:ablation_studies}).
Then, we report the results of training All and All+MD splits (Sec.~\ref{subsec:main_results}).
Additionally, we investigate the performance of EquiformerV2 when used in the AdsorbML algorithm~\citep{adsorbml} (Sec.~\ref{subsec:adsorbml_results}).
Please refer to Sec.~\ref{appendix:sec:details_of_architecture} and~\ref{appendix:sec:details_of_experiments_on_oc20} for details of models and training.

%##############################################################################
% Ablations / scaling results on OC20-2M.
\begin{table*}[t]
\centering
%#################################################
% Ablations
%#################################################
\subfloat[
%\textbf{Architectural improvements.} Attention re-normalization improves energies,
Architectural improvements.
Attention re-normalization improves energies,
and separable $S^2$ activation (``Sep. $S^2$'') and separable layer normalization (``SLN'') improve forces.
\label{tab:ablations}
]{
\centering
\begin{minipage}{0.53\linewidth}{\begin{center}

\tablestyle{4pt}{1.05}
\scalebox{0.7}{
\begin{tabular}{x{16}x{64}x{36}x{48}x{24}x{24}x{24}}
%{x{24}x{36}x{36}x{56}x{24}x{24}x{24}}
\toprule[1.2pt]
& Attention & & & & & \\
Index & re-normalization & Activation & Normalization & Epochs & forces & energy \\
%\shline
\midrule[1.2pt]
1 & \xmark & Gate & LN & 12 & 21.85 & 286 \\
2 & \cmark & Gate & LN & 12 & 21.86 & 279 \\
3 & \cmark & $S^2$ & LN & 12 & \multicolumn{2}{c}{didn't converge} \\
4 & \cmark & Sep. $S^2$ & LN & 12 & 20.77 & 285 \\
5 & \cmark & Sep. $S^2$ & SLN & 12 & \baseline{20.46} & \baseline{285} \\
6 & \cmark & Sep. $S^2$ & LN & 20 & 20.02 & 276 \\
7 & \cmark & Sep. $S^2$ & SLN & 20 & 19.72 & 278 \\
% \cmark & $S^2$ & SLN & 12 & 20.34 & 290 \\
% \cmark & Sep. $S^2$ & RMS & 12 & 19.84 & 279
%\hline
\midrule
8 & \multicolumn{3}{l}{eSCN baseline} & 12 & 21.3 & 294 \\
\bottomrule[1.2pt]
\end{tabular}
}
\vspace{1.5mm}
\end{center}}
\end{minipage}
}
\hspace{1.0em}
%#################################################
% No. of training epochs, higher LR
%#################################################
\subfloat[
%\textbf{Training epochs}. 
Training epochs.
Training for more epochs consistently leads to better results.
\label{tab:training_epochs}
]{
\centering
\begin{minipage}{0.4\linewidth}{\begin{center}
\tablestyle{4pt}{1.05}
\scalebox{0.7}{
\begin{tabular}{x{24}x{24}x{24}x{24}x{24}x{24}}
\toprule[1.2pt]
& & \multicolumn{2}{c}{eSCN} & \multicolumn{2}{c}{EquiformerV2} \\
\cmidrule(lr){3-4} \cmidrule(lr){5-6}
$L_{max}$ & Epochs & forces & energy & forces & energy \\
%\shline
\midrule[1.2pt]
6 & 12 & 21.3 & 294 & \baseline{20.46} & \baseline{285} \\
6 & 20 & 20.6 & 290 & 19.78 & 280 \\
6 & 30 & 20.1 & 285 & 19.42 & 278 \\
%\hline
\midrule
8 & 12 & 21.3 & 296 & 20.46 & 279 \\
8 & 20 & - & - & 19.95 & 273 \\
\bottomrule[1.2pt]
\multicolumn{6}{c}{~} \\
\multicolumn{6}{c}{~} \\
\multicolumn{6}{c}{~} \\
%\bottomrule[1.2pt]
\end{tabular}
}
\vspace{1.5mm}
\end{center}}
\end{minipage}
}
\hspace{2em}
\\[0.1in]
%#################################################
% Scaling up Lmax
%#################################################
\subfloat[
%\textbf{Degrees $L_{max}$.} 
Degrees $L_{max}$.
Higher degrees are consistently helpful.
\label{tab:l_degrees}
]{
\centering
\begin{minipage}{0.3\linewidth}{\begin{center}
\tablestyle{4pt}{1.05}
\scalebox{0.7}{
\begin{tabular}{x{24}x{24}x{24}x{24}x{24}}
\toprule[1.2pt]
& \multicolumn{2}{c}{eSCN} & \multicolumn{2}{c}{EquiformerV2} \\
\cmidrule(lr){2-3} \cmidrule(lr){4-5}
$L_{max}$ & forces & energy & forces & energy \\
%\shline
\midrule[1.2pt]
4 & 22.2 & 291 & 21.37 & 284 \\
6 & 21.3 & 294 & \baseline{20.46} & \baseline{285} \\
8 & 21.3 & 296 & 20.46 & 279 \\
\bottomrule[1.2pt]
\multicolumn{5}{c}{~} \\
\end{tabular}
}
\vspace{1.5mm}
\end{center}}
\end{minipage}
}
\hspace{0.6em}
%#################################################
% Scaling up Mmax
%#################################################
\subfloat[
%\textbf{Orders $M_{max}$.} 
Orders $M_{max}$.
Higher orders mainly improve energy predictions.
\label{tab:m_orders}
]{
\centering
\begin{minipage}{0.3\linewidth}{\begin{center}
\tablestyle{4pt}{1.05}
\scalebox{0.7}{
\begin{tabular}{x{24}x{24}x{24}x{24}x{24}}
\toprule[1.2pt]
& \multicolumn{2}{c}{eSCN} & \multicolumn{2}{c}{EquiformerV2} \\
\cmidrule(lr){2-3} \cmidrule(lr){4-5}
$M_{max}$ & forces & energy & forces & energy \\
%\shline
\midrule[1.2pt]
2 & 21.3 & 294 & \baseline{20.46} & \baseline{285} \\
3 & 21.2 & 295 & 20.24 & 284 \\
4 & 21.2 & 298 & 20.24 & 282 \\
6 & - & - & 20.26 & 278 \\
\bottomrule[1.2pt]
\end{tabular}
}
\vspace{1.5mm}
\end{center}}
\end{minipage}
}
\hspace{0.6em}
%#################################################
% No. of layers
%#################################################
\subfloat[
%\textbf{No. of layers}. Adding more Transformer blocks can help
%both force and energy accuracies.
%\textbf{Number of blocks.}
Number of Transformer blocks.
Adding more blocks can help both force and energy predictions.
\label{tab:num_layers}
]{
\centering
\begin{minipage}{0.3\linewidth}{\begin{center}
\tablestyle{4pt}{1.05}
\scalebox{0.7}{
\begin{tabular}{x{24}x{24}x{24}x{24}x{24}}
\toprule[1.2pt]
& \multicolumn{2}{c}{eSCN} & \multicolumn{2}{c}{EquiformerV2} \\
\cmidrule(lr){2-3} \cmidrule(lr){4-5}
Layers & forces & energy & forces & energy \\
%\shline
\midrule[1.2pt]
8 & 22.4 & 306 & 21.18 & 293 \\
12 & 21.3 & 294 & \baseline{20.46} & \baseline{285} \\
16 & 20.5 & 283 & 20.11 & 282 \\
\bottomrule[1.2pt]
\multicolumn{5}{c}{~} \\
\end{tabular}
}
\vspace{1.5mm}
\end{center}}
\end{minipage}
}
%\\[0.1in]
\hspace{0.6em}

%\vspace{2mm}
\vspace{1mm}

%#################################################
% Speed-accuracy trade-off
%#################################################
% %\vspace{3mm}
% \subfloat[
% \textbf{Speed-accuracy trade-offs}.
% \label{tab:speed_accuracy}
% ]{
% \centering
% \begin{minipage}{0.45\linewidth}{\begin{center}
% \tablestyle{4pt}{1.05}
% \scalebox{0.75}{
% \begin{tabular}{x{24}x{24}x{24}x{24}x{24}x{24}}
% & & \multicolumn{2}{c}{eSCN} & \multicolumn{2}{c}{EquiformerV2} \\
% \cmidrule(lr){3-4} \cmidrule(lr){5-6}
% $L_{max}$ & Width & forces & speed & forces & speed \\
% \shline
% 4 & 0.50 & 22.7 & 29.83 & 21.46 & 41.37 \\
% 4 & 0.75 & 21.7 & 26.36 & 19.95 & 26.53 \\
% 4 & 1.00 & 21.7 & 24.26 & 19.83 & 20.83 \\
% 6 & 1.00 & 20.5 & 18.15 & 19.42 & 13.27 \\
% %\multicolumn{5}{c}{~} \\
% \vspace{1.5mm}
% \end{tabular}
% }
% \end{center}}
% \end{minipage}
% }
% \hspace{2em}
%#################################################
%\vspace{-.1em}
% \vspace{-2mm}
\caption{
%EquiformerV2 ablations and scaling experiments. We report mean absolute
%errors for forces (in meV/\angstrom) and energy (in meV), so lower is better.
%\textbf{Ablation results with EquiformerV2}.
%Ablation results with EquiformerV2.
Ablation studies of EquiformerV2.
We report mean absolute errors (MAE) for forces in meV/\angstrom
 and energy in meV, and lower is better.
%
%Speed is calculated as no. of samples forwarded per GPU-second, so higher is
%better.
%Speed is reported as the number of structures processed per GPU-second measured on V100 GPUs during inference, and higher is better.
%
All models are trained on the 2M split of the OC20 S2EF dataset, and
errors are averaged over the four validation sub-splits.
%
%Base model settings are marked in \colorbox{baselinecolor}{gray}.
The base model setting is marked in \colorbox{baselinecolor}{gray}.
}
%\vspace{-20pt}
\label{tab:ablations_all}
\end{table*}
%##############################################################################

%#################################################
% Speed-accuracy trade-off
%#################################################

\begin{comment}
\begin{table}[t]
\begin{tabular}{c}
    \tablestyle{4pt}{1.05}
    \scalebox{0.75}{
    \begin{tabular}{x{24}x{24}x{24}x{24}x{24}x{24}}
        & & \multicolumn{2}{c}{eSCN} & \multicolumn{2}{c}{EquiformerV2} \\
        \cmidrule(lr){3-4} \cmidrule(lr){5-6}
        $L_{max}$ & Width & forces & speed & forces & speed \\
        \shline
        4 & 0.50 & 22.7 & 29.83 & 21.46 & 41.37 \\
        4 & 0.75 & 21.7 & 26.36 & 19.95 & 26.53 \\
        4 & 1.00 & 21.7 & 24.26 & 19.83 & 20.83 \\
        6 & 1.00 & 20.5 & 18.15 & 19.42 & 13.27 \\
    \end{tabular}
    }
    \includegraphics[scale=0.35, valign=m]{neurips_2023/figures/20230515_speed_accuracy.png}
\end{tabular}
\caption{Tree representation of localities and their location types given by the Open Street Map.}
\label{fig:speed_accuracy_trade_offs}
\end{table}
\end{comment}

%\vspace{-2.5mm}
\subsubsection{Ablation Studies}
\label{subsec:ablation_studies}
%\vspace{-2mm}

%\textbf{Ablation and scaling results}.
%Across tables, our `base' EquiformerV2 variant is highlighted in \colorbox{baselinecolor}{gray}.
%
%We perform ablation studies on the proposed architectural improvements discussed in Sec.~\ref{sec:equiformer_v2}.
%Additionally, we study how scaling maximum degree $L_{max}$, maximum order $M_{max}$, the number of Transformer blocks $N$ and the number of training epochs affects performance on energies and forces.
%The results are summarized in Table~\ref{tab:ablations_all}.

%Our `base' EquiformerV2 variant is highlighted in \colorbox{baselinecolor}{gray} in Table~\ref{tab:ablations_all}.

%\vspace{-3mm}
%\textbf{Architectural Improvements}.
\paragraph{Architectural Improvements.}
%In~\tableref{tab:ablations}, we ablate the three proposed architectural
%changes -- attention re-normalization, separable $S^2$ activation and separable layer
%normalization.
In~\tableref{tab:ablations}, we start with incorporating eSCN convolutions into Equiformer for higher-degree representations (Index 1) and then ablate the three proposed architectural improvements.
%
%First, with attention renormalization (row 1~\vs. 2), energy errors improve by
%$2.4\%$, while force errors are about the same.
%
First, with attention re-normalization, energy mean absolute errors (MAE) improve by $2.4\%$, while force MAE are about the same (Index 1 and 2).
%Next, we try replacing the gated equivariant non-linearity with $S^2$ activation
%(as in eSCN~\cite{escn}), but that does not converge (row 3).
%Next, we replace the gate activation with $S^{2}$ activation used in SCN~\cite{scn} and eSCN~\cite{escn}, but that does not converge (row 3).
Second, we replace the gate activation with $S^{2}$ activation, but that does not converge (Index 3).
%
%Instead, replacing it with separable $S^2$ activation (row 4), where we have
%separate paths for invariant and equivariant features, converges to $5\%$ better
%forces (albeit hurting energies).
%Instead, using the proposed separable $S^2$ activation (Index 4), where we have separate paths for invariant and equivariant features, converges to $5\%$ better forces albeit hurting energies.
%Instead, using the proposed separable $S^2$ activation (Index 4) converges to $5\%$ better forces albeit hurting energies.
%
%Similarly, separating layer normalization for invariant and equivariant features
%(row 5) further improves forces by ${\sim}1.5\%$.
%Similarly, separating layer normalization for invariant and equivariant features (Index 5) further improves forces by $1.5\%$.
With the proposed $S^2$ activation, we stabilize training and successfully leverage more expressive activation to improve force MAE (Index 4).
%Similarly, replacing equivariant layer normalization with separable layer normalization (row 5) further improves forces by $1.5\%$.
Third, replacing equivariant layer normalization with separable layer normalization further improves force MAE (Index 5).
%
%Finally, these modifications enable training for longer without overfitting
%(row 7), further improving forces by $3.6\%$ and recovering energies to similar
%accuracies as row 2.
%Finally, these modifications enable training for longer without overfitting (row 7), further improving forces by $3.6\%$ and recovering energies to similar accuracies as row 2.
%
%Overall, our modifications put together improve forces by ${\sim}10\%$ and
%energies by ${\sim}3\%$.
%%%Overall, our modifications improve forces by $10\%$ and energies by $3\%$.
We note that simply incorporating eSCN convolutions into Equiformer
%(row 1) and using higher degrees does not result in improving over the `eSCN baseline', and that the proposed architectural changes are necessary.
and using higher degrees (Index 1) do not result in better performance than the original eSCN baseline (Index 8), and that the proposed architectural changes are necessary.
We additionally compare the performance gain of architectural improvements with that of training for longer.
Attention re-normalization (Index 1 and 2) improves energy MAE by the same amount as increasing training epochs from $12$ to $20$ (Index 5 and 7).
The improvement of SLN (Index 4 and 5) in force MAE is about $40\%$ of that of increasing training epochs from $12$ to $20$ (Index 5 and 7), and SLN is about $6\%$ faster in training.
We also conduct the ablation studies on the QM9 dataset and summarize the results in Sec.~\ref{appendix:subsec:qm9_ablation_study}.

%\textbf{Scaling of Parameters}.
%\vspace{-3mm}
\paragraph{Scaling of Parameters.}
%In~\threetablerefs{tab:l_degrees}{tab:m_orders}{tab:num_layers}, we systematically vary the maximum degree $L_{max}$, the maximum order $M_{max}$, and the number of Transformer blocks and compare with equivalent eSCN variants.
In~\threetablerefs{tab:l_degrees}{tab:m_orders}{tab:num_layers}, we vary the maximum degree $L_{max}$, the maximum order $M_{max}$, and the number of Transformer blocks and compare with equivalent eSCN variants.
%
%There are several key takeaways. 
%First, across all experiments, EquiformerV2
%performs better than its eSCN counterparts.
Across all experiments, EquiformerV2 performs better than its eSCN counterparts.
%
%Second, while one might intuitively expect higher resolution features and larger
%models to perform better, this is only true for EquiformerV2, not eSCN.
Besides, while one might intuitively expect higher resolutions and larger models to perform better, this is only true for EquiformerV2, not eSCN.
%For example, varying $L_{max}$ from $6 \rightarrow 8$ or $M_{max}$ from
%$3 \rightarrow 4$ with eSCN hurts energy predictions, while helping them
%with EquiformerV2.
For example, increasing $L_{max}$ from $6$ to $8$ or $M_{max}$ from $3$ to $4$ degrades the performance of eSCN on energy predictions but helps that of EquiformerV2.
%
%Finally, we find that varying $L_{max}$ from $4 \rightarrow 6$ or no. of layers from
%$8 \rightarrow 12 \rightarrow 16$ with EquiformerV2 significantly improves force
%predictions.
%
%In~\tableref{tab:training_epochs}, we show that longer training regimes are
%crucial.
%
%Increasing the training epochs from $12 \rightarrow 30$ improves force
%and energy predictions by $5\%$ and $2.5\%$ respectively with $L_{max}=6$.
%In~\tableref{tab:training_epochs}, we show that longer training regimes are
%crucial.
%Increasing the number of training epochs from $12$ to $30$ improves force and energy predictions by $5\%$ and $2.5\%$ respectively with $L_{max}=6$.
%Increasing the training epochs from $12$ to $30$ with $L_{max}=6$ improves force and energy predictions by $5\%$ and $2.5\%$, respectively.
In~\tableref{tab:training_epochs}, we show that increasing the training epochs from $12$ to $30$ epochs can consistently improve performance.

%\paragraph{Comparison of Speed-Accuracy Trade-offs.}

%\vspace{-3mm}
\paragraph{Speed-Accuracy Trade-offs.}

\begin{figure*}[t]
  \centering
  \begin{minipage}[b]{0.35\textwidth}
      \begin{subfigure}{\textwidth}
        \includegraphics[width=0.9\linewidth]{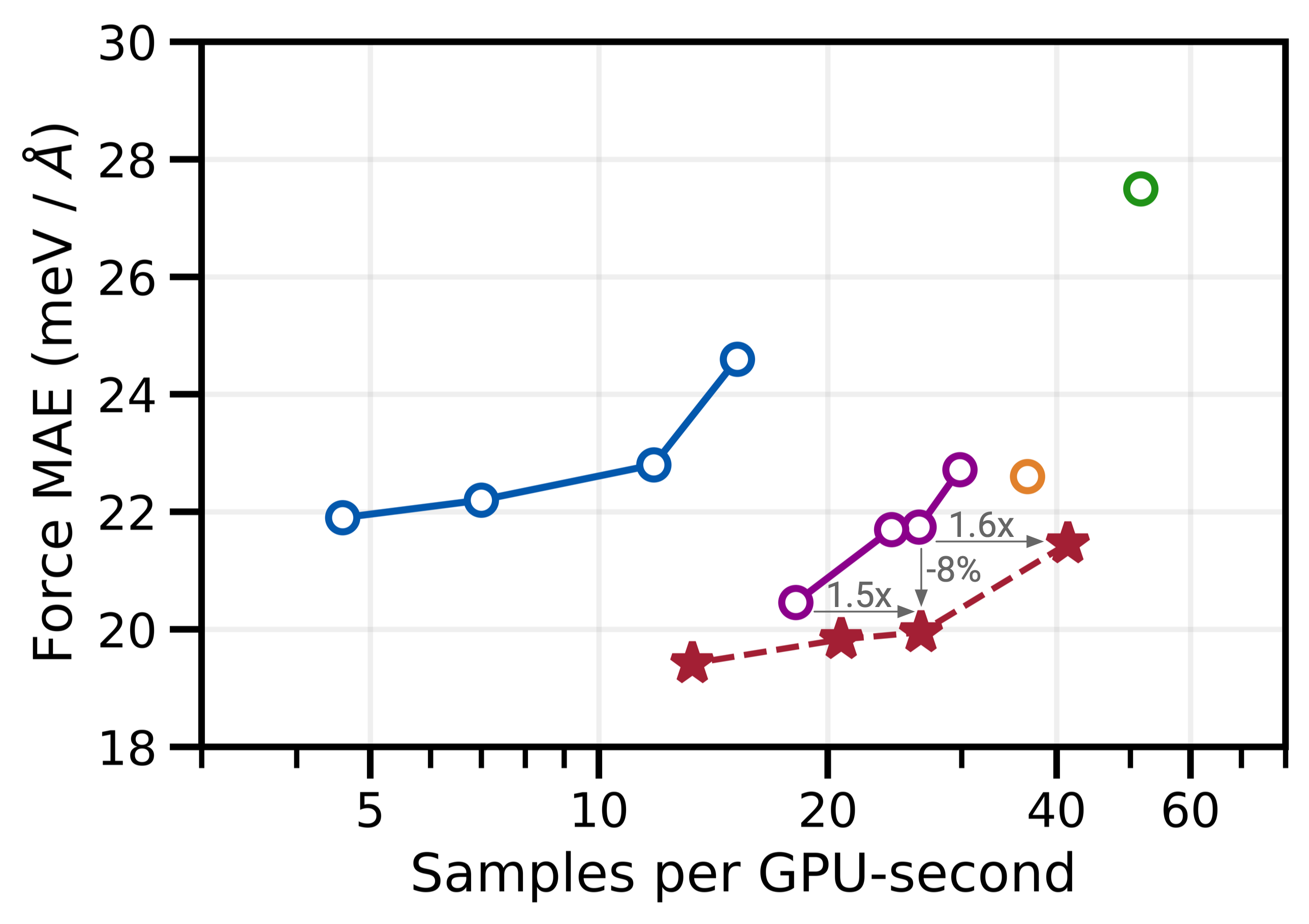}
          \caption{Trade-offs between inference speed and validation force MAE.}
          \label{fig:speed_accuracy}
      \end{subfigure}
  \end{minipage}
  \hspace{5pt}
  \begin{minipage}[b]{0.35\textwidth}
      \begin{subfigure}{\linewidth}
        \includegraphics[width=0.9\linewidth]{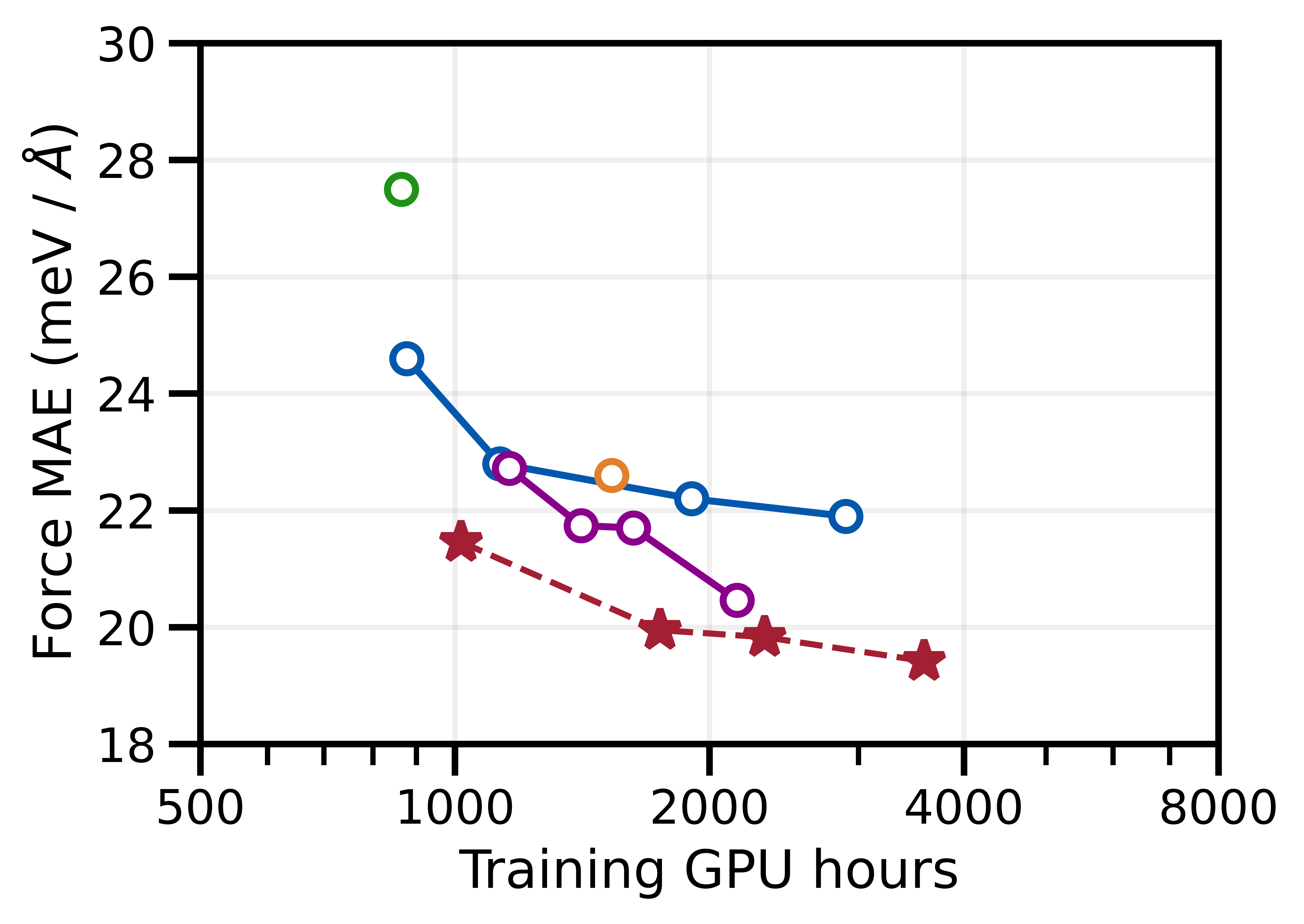}
          \caption{Trade-offs between training cost and validation force MAE.}
          \label{fig:compute_accuracy}
      \end{subfigure}
  \end{minipage}
  % \hspace{5pt}
  \begin{minipage}[b]{0.2\textwidth}
    \begin{subfigure}{\linewidth}
      \includegraphics[width=0.9\linewidth]{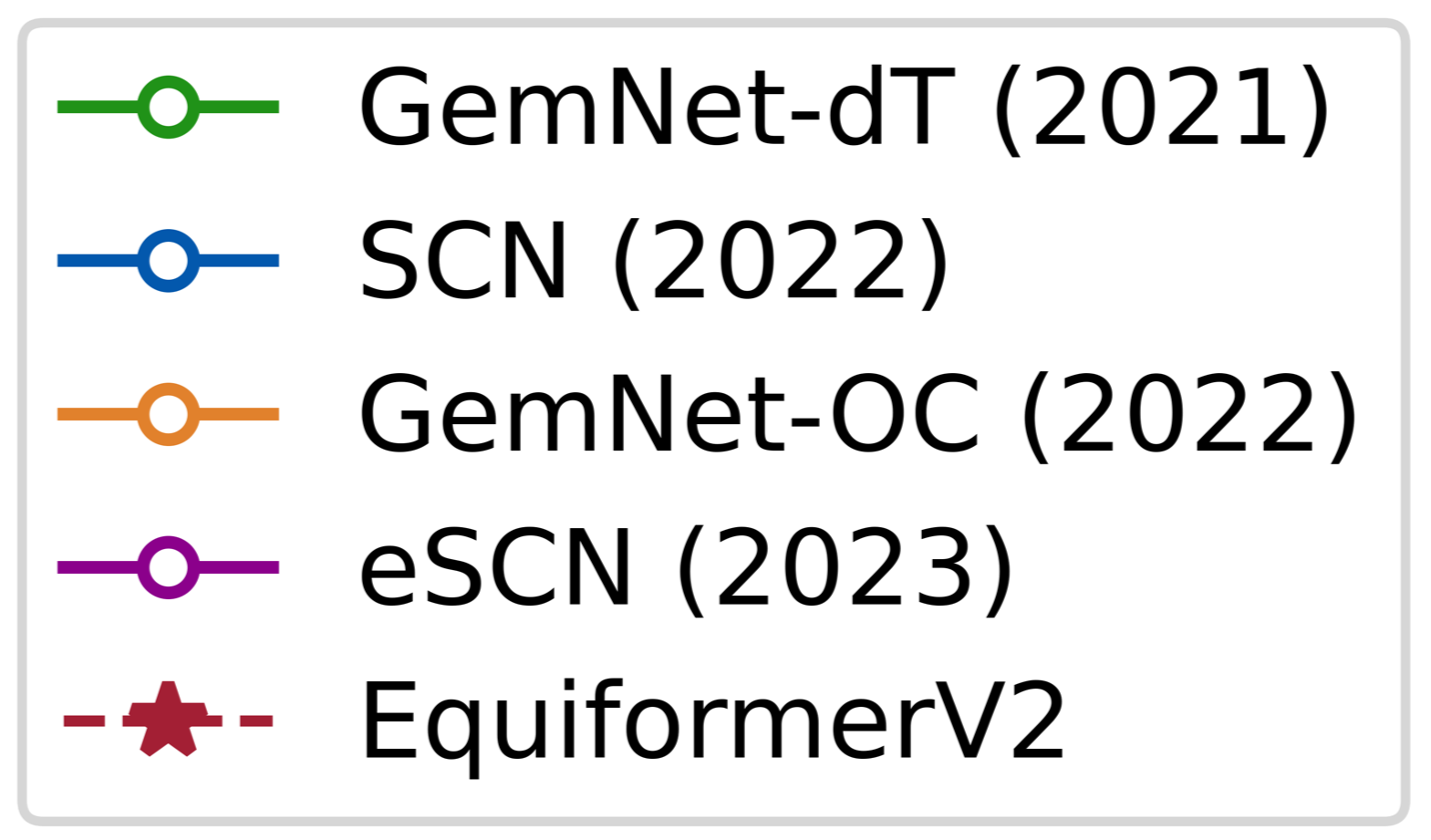}
      \\[0.74in]
    \end{subfigure}
\end{minipage}
  \vspace{4pt}
  \caption{
  %EquiformerV2 offers better accuracy trade-offs both in terms of inference speed as well as training cost compared to prior works.
  EquiformerV2 achieves better accuracy trade-offs both in terms of inference speed as well as training cost.
  All models are trained on the S2EF-2M split and measured on V100 GPUs.
  }
\label{fig:combined_train_test_throughput}
  %\vspace{-18pt}
  
\end{figure*}

We compare trade-offs between inference speed or training time and force MAE among prior works and EquiformerV2 and summarize the results in~\Figref{fig:speed_accuracy} and~\Figref{fig:compute_accuracy}. 
EquifomerrV2 achieves the lowest force MAE across a wide range of inference speed and training cost.

%##############################################################################
\begin{table}[t]
    \centering
    \sisetup{round-mode=figures, round-precision=3}
    \resizebox{1\linewidth}{!}{
    \begin{tabular}{l   l rr  S[table-format=2.1,round-mode=places,round-precision=1] S[table-format=3]   S[table-format=2.1] S[table-format=3]   S[table-format=2.1]   S[table-format=2.1]   S[table-format=2.1] S[table-format=3]}
    \toprule[1.2pt]
    & & & & \mcc{Throughput} & \multicolumn{2}{c}{S2EF validation} & \multicolumn{2}{c}{S2EF test}                                                                                                                                               & \multicolumn{2}{c}{IS2RS test}                                       & \mcc{IS2RE test}                                 \\
    \cmidrule(lr){5-5} \cmidrule(lr){6-7} \cmidrule(lr){8-9} \cmidrule(lr){10-11} \cmidrule(lr){12-12}
    Training & & \mcc{Number of}& \mcc{Training time} & \mcc{Samples /} & \mcc{Energy MAE}          & \mcc{Force MAE}                     & \mcc{Energy MAE}          & \mcc{Force MAE}                     & \mcc{AFbT}                     & \mcc{ADwT}                     & \mcc{Energy MAE}                            \\
     set & Model & \mcc{parameters $\downarrow$} & \mcc{(GPU-days) $\downarrow$} & \mcc{GPU sec. $\uparrow$} & \mcc{\si{(\milli\electronvolt)} $\downarrow$} & \mcc{\si[per-mode=symbol]{(\milli\electronvolt\per\angstrom)} $\downarrow$} & \mcc{\si{(\milli\electronvolt)} $\downarrow$} & \mcc{\si[per-mode=symbol]{(\milli\electronvolt\per\angstrom)} $\downarrow$} & \mcc{\si{(\percent)} $\uparrow$} & \mcc{\si{(\percent)} $\uparrow$} & \mcc{\si{(\milli\electronvolt)} $\downarrow$} \\ 
     \midrule[1.2pt]
    % \multirow{5}{*}{\rotatebox[origin=c]{90}{OC-2M}}
    %     & SchNet &     \mcc{-} &                     1400.7911 &                     78.346525 &                  1371.16045 &                     77.136025 &                      \mcc{-} &                       \mcc{-} &                       \mcc{-} \\
    %     & DimeNet$^{++}$ &     \mcc{-} &                    805.174225 &                     65.658725 &            761.1773 &                       62.9794 &                        \mcc{-} &                       \mcc{-} &                       \mcc{-} \\
    %     & SpinConv &     \mcc{-} &                    406.438925 &                      36.19075 &                  401.117325 &                      35.49425 &                       \mcc{-} &                       \mcc{-} &                       \mcc{-} \\
    %     & GemNet-dT &     \mcc{-} &                    358.426625 &                      29.53875 &                  323.315625 &                      28.08195 &                      16.7375 &                         54.79 &                    437.737025 \\
    %     & GemNet-OC &     \mcc{-} &                  285.657475 &                25.651625 &         273.71645 &                 24.29935 &           19.57 &                   56.43 &        407.17400000000004 \\ \midrule
    \multirow{8}{*}{\rotatebox[origin=c]{90}{OC20 S2EF-All}}
        %& CGCNN~\cite{cgcnn} &     \mcc{-} &                       589.675 &                        74.025 &                    607.725 &                          73.3 &                       \mcc{-} &                       \mcc{-} &                       \mcc{-} \\
        & SchNet~\citep{schnet} & 9.1M & 194 &    \mcc{-} &                       548.525 &                          56.8 &                   540.375 &                         54.65 &                      \mcc{-} &                       14.3575 &                       749 \\
        %& ForceNet-large~\cite{forcenet} &   15.276060 &                       \mcc{-} &                     33.500425 &                  \mcc{-} &                         31.98 &                      12.6575 &                       49.6225 &                       \mcc{-} \\
        & DimeNet++-L-F+E~\citep{dimenet_pp} & 10.7M & 1600 &    4.565029 &                       515.475 &                         32.75 &                    479.925 &                          31.3 &                      21.745 &                        51.665 &                       558.725 \\
        & SpinConv~\citep{spinconv} & 8.5M & 275 &   6.020185 &                     371.01165 &                     41.197875 &                  336.25 &                       29.6575 &                      16.6675 &                        53.625 &                       436.675 \\
        & GemNet-dT~\citep{gemnet} & 32M & 492 &  25.807363 &                    315.118775 &                     27.172475 &                  292.4128 &                      24.21645 &                      27.5975 &                       58.6775 &                     399.72965 \\
        %& GemNet-XL~\cite{gemnet_xl} &    1.498224 &                       \mcc{-} &                       \mcc{-} &                   270.075 &          20.450000000000003 &                      30.82 &                    62.655 &                        371.15 \\
        & GemNet-OC~\citep{gasteiger_gemnet_oc_2022} & 39M & 336 &   18.334700 &          244.28979999999999 &        21.719050000000003 &        232.95987499999998 &                       20.6671 &       35.275 &                         60.33 &                 354.576675 \\
        & SCN L=8 K=20~\citep{scn} & 271M & 645 &           \mcc{-} &            \mcc{-} &            \mcc{-} &            244 &            17.7 &          40.3 &          67.1 &          330 \\
        & eSCN L=6 K=20~\citep{escn} & 200M & 600 &          2.9 &           \mcc{-} &           \mcc{-} &           242 &           17.1 &          48.524 &          65.708 &          341 \\
        & EquiformerV2 ($\lambda_E=2$, $153$M) &  153M & 1368 &       1.8 & \textbf{236} &         \textbf{15.7} &         \textbf{229} &         \textbf{14.8} &         \textbf{53.0} &           \textbf{69.0} &           \textbf{316} \\
    \midrule
    %\multirow{9}{*}{\rotatebox[origin=c]{90}{\parbox[c]{3.6em}{OC20 S2EF-All+MD}}}
    \multirow{9}{*}{\rotatebox[origin=c]{90}{OC20 S2EF-All+MD}}
        & GemNet-OC-L-E~\citep{gasteiger_gemnet_oc_2022} &  56M & 640 &  7.534085 &          238.93085000000002 &                     22.068625 &                  230.174375 &                       20.9918 &                      \mcc{-} &                       \mcc{-} &                       \mcc{-} \\
        & GemNet-OC-L-F~\citep{gasteiger_gemnet_oc_2022} &  216M & 765 &  3.178525 &                    252.255025 &                  19.9955 &           241.134825 &             19.012375 &          40.555 &      60.442499999999995 &                       \mcc{-} \\
        & GemNet-OC-L-F+E~\citep{gasteiger_gemnet_oc_2022} & - & - &      \mcc{-} &                       \mcc{-} &                       \mcc{-} &                      \mcc{-} &                       \mcc{-} &           \mcc{-} &                       \mcc{-} &                347.634875 \\
        & SCN L=6 K=16 {\scriptsize (4-tap 2-band)}~\citep{scn} & 168M & 414 & \mcc{-} &           \mcc{-} &           \mcc{-} &           228 &           17.8 &          43.3 &          64.9 &           328 \\
        & SCN L=8 K=20~\citep{scn} &  271M & 1280 &    \mcc{-} &           \mcc{-} &           \mcc{-} &          237 &           17.2 &          43.6 &          67.5 &          321 \\
        & eSCN L=6 K=20~\citep{escn} & 200M & 568 &    2.9 &           242.9559936 &           17.09092648 &          228 &           15.6 &          50.296 &          66.707 &          323 \\
        & EquiformerV2 ($\lambda_E=4$, $31$M) & 31M & 705 &        7.1 &           232 &         16.26 &          228 &         15.51 &           47.6 &           68.3 &           315 \\
        & EquiformerV2 ($\lambda_E=2$, $153$M) & 153M & 1368 &         1.8 &           \textbf{230} &         \textbf{14.6} &          227.0 &         \textbf{13.8} &         \textbf{55.4} &           \textbf{69.8} &           \textbf{311} \\
        & EquiformerV2 ($\lambda_E=4$, $153$M) & 153M & 1571 &          1.8 &           \textbf{227} &         15.04 &          \textbf{219} &         14.20 &           54.4 &           69.4 &           \textbf{309} \\
    \bottomrule[1.2pt]
    \end{tabular}}
    \vspace{1mm}
    \caption{OC20 results on S2EF validation and test splits, and IS2RS and IS2RE test splits when trained on OC20 S2EF-All or S2EF-All+MD splits.
      %Models are trained on OC20 S2EF-All or S2EF-All+MD splits.
      Throughput is reported as the number of structures processed per GPU-second during training and measured on V100 GPUs.
      $\lambda_{E}$ is the coefficient of the energy loss.}
    %\vspace{-12pt}
    \label{tab:oc20_all_results}
\end{table}
%##############################################################################

%##############################################################################
%##############################################################################

%\vspace{-2mm}
\subsubsection{Main Results}
\label{subsec:main_results}
%\vspace{-2mm}
%Finally, \tableref{tab:oc20_all_results} reports results on the \textsc{test}
%splits for all the three tasks of OC20, averaged across the in-distribution,
%out-of-distribution adsorbates, out-of-distribution catalysts, and out-of-distribution
%both subsplits.
%\tableref{tab:oc20_all_results} reports results on the test splits for all the
%three tasks of OC20, averaged across the in-distribution, out-of-distribution
%adsorbates, out-of-distribution catalysts, and out-of-distribution both subsplits.
\tableref{tab:oc20_all_results} reports results on the test splits for all the three tasks of OC20 averaged over all four sub-splits.
%three tasks of OC20, averaged across the
%
Models are trained on either OC20 S2EF-All or S2EF-All+MD splits.
All test results are computed via the EvalAI evaluation server\footnote{\href{https://eval.ai/web/challenges/challenge-page/712/overview}{{\tt eval.ai/web/challenges/challenge-page/712}}}.
We train EquiformerV2 of two sizes, one with $153$M parameters and the other with $31$M parameters.
%\revision{As shown in Table~\ref{appendix:tab:oc20_s2ef_hyperparameters}, the smaller one is obtained by reducing the maximum degree $L_{max}$ from $6$ to $4$, the maximum order $M_{max}$ from $3$ to $2$ and the number of Transformer blocks from $20$ to $8$.}
%EquiformerV2 outperforms all previous models across all tasks, improving by $4\%$
%on S2EF energy MAE, by $15\%$ on S2EF force MAE, by $5\%$ absolute on IS2RS
%Average Forces below Threshold (AFbT), and by $4\%$ on IS2RE energy MAE.
%\revision{EquiformerV2 outperforms all previous models across all tasks.}
When trained on the S2EF-All+MD split, EquiformerV2 ($\lambda_E = 4$, $153$M) improves previous state-of-the-art S2EF energy MAE by $4\%$, S2EF force MAE by $9\%$, IS2RS Average Forces below Threshold (AFbT) by absolute $4\%$ and IS2RE energy MAE by $4\%$.
%When trained on the S2EF-All+MD split, EquiformerV2 ($\lambda_E = 4$, $153$M) sets new state-of-the-art for all tasks.
%
%In particular, the AFbT metric is computed via Density Functional Theory (DFT)
%single-point calculations to check if the DFT forces on the predicted relaxed
%structures are close to zero.
In particular, the improvement in force predictions is significant.
%
%Going from SCN~\cite{scn} to eSCN~\cite{escn}, S2EF test force MAE improves from
%$17.2$ meV/\angstrom to $16.2$ meV/\angstrom, largely due to replacing
%approximate equivariance in SCN with strict equivariance in eSCN during
%message passing and scaling to higher degrees.
Going from SCN to eSCN, S2EF test force MAE improves from $17.2$ meV/\angstrom to $15.6$ meV/\angstrom, largely due to replacing approximate equivariance in SCN with strict equivariance in eSCN during message passing.
% , eSCN~\cite{escn} improves upon SCN~\cite{scn} by reducing force MAE from $17.2$meV/\angstrom to $16.2$meV/\angstrom when trained on S2EF-All+MD split.}
%%Similarly, by scaling up the degrees of representations in Equiformer~\cite{equiformer},
%EquiformerV2 further improves force MAE to $13.8$ meV/\textup{\AA}, more than
%%\revision{EquiformerV2 ($\lambda_E = 4$, $153$M) further improves force MAE to $14.2$ meV/\textup{\AA}, }
%doubling the performance gain of SCN$\rightarrow$eSCN.
%%doubling the gain of going from SCN to eSCN.
Similarly, by scaling up the degrees of representations in Equiformer, EquiformerV2 ($\lambda_E = 4$, $153$M) further improves force MAE to $14.2$ meV/\textup{\AA}, which is similar to the gain of going from SCN to eSCN.
%
%These better force predictions also translate to higher IS2RS test AFbT, which is
%computed via DFT single-point calculations to check if the DFT forces on the
%predicted relaxed structures are close to zero.
%A $4\%$ improvement on AFbT is a strong step towards replacing DFT with ML.
Additionally, the smaller EquiformerV2 ($\lambda_E = 4$, $31$M) improves upon previously best results for all metrics except IS2RS AFbT and achieves comparable training throughput to the fastest GemNet-OC-L-E.
Although the training time of EquiformerV2 is higher here, we note that this is because training EquiformerV2 for longer keeps improving performance and that we already demonstrate EquiformerV2 achieves better trade-offs between force MAE and speed.

%All \textsc{test} results are computed via the
%EvalAI evaluation server\footnote{\href{https://eval.ai/web/challenges/challenge-page/712/overview}{{\tt eval.ai/web/challenges/challenge-page/712}}}.

%
% S2EF-\textsc{val}, S2EF-\textsc{test},
% IS2RS-\textsc{test}, IS2RE-\textsc{test}

\begin{comment}
\textbf{Speed-accuracy comparison}.
To be practically useful for atomistic simulations and material screening, a
desideratum for models is to offer flexibility in speed-accuracy tradeoffs.
%
We were interested in investigating this for EquiformerV2, comparing it with
prior works (see \tableref{tab:speed_accuracy},~\Figref{fig:speed_accuracy}).
%
\ad{We omit energy MAEs for brevity, but the trends there are consistent and
reported in the appendix. Verify this claim.}
%
Here, speed is reported as no. of structures forwarded per GPU-second (computed
on V100s) during inference.
%
For the same force MAE as eSCN, EquiformerV2 is up to ${\sim}1.6\times$ faster, and
for the same speed as eSCN, EquiformerV2 is up to ${\sim}8\%$ more accurate.
%
Compared to GemNet-OC~\cite{gasteiger_gemnet_oc_2022} at the same speed,
EquiformerV2 is ${\sim}5\%$ more accurate.
%
Comparing to the closest available EquifomerV2 point, GemNet-dT~\cite{gemnet} is
${\sim}1.25\times$ faster, but ${\sim}30\%$ worse!
%
\ad{Maybe also emphasize that among the variants we tried, you can get 3x speedup if you're willing to lose $10\%$ accuracy.}
%
Overall, EquiformerV2 seems to offer a better speed-accuracy Pareto curve.
\end{comment}

%\textbf{AdsorbML results}.
%\vspace{-3mm}
\subsubsection{AdsorbML Results}
\label{subsec:adsorbml_results}
\cite{adsorbml} recently proposes the AdsorbML algorithm, wherein they show that recent state-of-the-art GNNs can achieve more than $1000\times$ speedup over DFT relaxations at computing adsorption energies within a $0.1$eV margin of DFT results with an $87\%$ success rate.
This is done by using OC20-trained models to perform structure relaxations for an average $90$ configurations of an adsorbate placed on a catalyst surface, followed by DFT single-point calculations for the top-$k$ structures with lowest predicted relaxed energies, as a proxy for calculating the global energy minimum or adsorption energy.
%
%They show that AdsorbML with recent state-of-the-art models (e.g., SCN~\cite{scn})
%can achieve more than $1000\times$ speedup over DFT relaxations at computing
%adsorption energies within a $0.1$eV margin or lower of DFT with an $87\%$
%success rate.
%
We refer readers to Sec.~\ref{appendix:subsec:details_of_adsorbml} and the work~\citep{adsorbml} for more details.
%
%We benchmark AdsorbML with EquiformerV2 and find that it improves over SCN by a
%significant margin (see \tableref{tab:adsorbml}) -- ${\sim}8\%$ and ${\sim}5\%$
%absolute improvements at $k=1$ and $k=2$ respectively. In other words,
%EquiformerV2 at $k=1$ is more accurate at adsorption energy calculations than
%SCN at $k=2$, requiring $2\times$ fewer DFT calculations as a result!
We benchmark AdsorbML with EquiformerV2, and \tableref{tab:adsorbml} shows that \revision{EquiformerV2 ($\lambda_E = 4$, $153$M)} improves over SCN by a significant margin, with $8\%$ and $5\%$ absolute improvements at $k=1$ and $k=2$, respectively.
%In other words, EquiformerV2 at $k=1$ is more accurate at adsorption energy calculations than SCN at $k=2$, requiring $2\times$ fewer DFT calculations as a result!
Moreover, EquiformerV2 ($\lambda_E = 4$, $153$M) at $k=2$ is more accurate at adsorption energy calculations than all the other models even at $k=5$, thus requiring at least $2\times$ fewer DFT calculations.
\revision{Since the speedup is with respect to using DFT for structure relaxations and that ML models are much faster than DFT, the speedup is dominated by the final DFT single-point calculations and ML models with the same value of $k$ have roughly the same speedup.}
%\revision{To better understand the speed-accuracy trade-offs of different models, we compare the AdsorbML success rate averaged over $k$ from $1$ to $5$ and average GPU-seconds of running one structure relaxation in Figure~\ref{fig:adsorbml_speed_sucess}.}
To better understand the speed-accuracy trade-offs of different models, we also report average GPU-seconds of running one structure relaxation in \tableref{tab:adsorbml}.
Particularly, EquiformerV2 ($\lambda_E = 4$, $31$M) improves upon previous methods while being $3.7\times$ to $9.8\times$ faster than GemNet-OC-MD-Large and SCN, respectively.
%\revision{Besides, the larger EquiformerV2 is $2.1\%$ more accurate but $3.7\times$ slower than the smaller one.}

%##############################################################################
\begin{table*}[t]
    \resizebox{\textwidth}{!}{%
    \begin{tabular}{lrcccccccccc}
    \toprule[1.2pt]
     &
     \multicolumn{1}{c}{GPU-seconds} & 
      \multicolumn{2}{c}{\textit{$k=1$}} &
      \multicolumn{2}{c}{\textit{$k=2$}} &
      \multicolumn{2}{c}{\textit{$k=3$}} &
      \multicolumn{2}{c}{\textit{$k=4$}} &
      \multicolumn{2}{c}{\textit{$k=5$}} \\
    \cmidrule(lr){3-4} \cmidrule(lr){5-6} \cmidrule(lr){7-8} \cmidrule(lr){9-10} \cmidrule(lr){11-12}
    Model &
      \multicolumn{1}{c}{per relaxation} $\downarrow$ &
      Success &
      Speedup &
      Success &
      Speedup &
      Success &
      Speedup &
      Success &
      Speedup &
      Success &
      Speedup  
       \\
    \midrule[1.2pt]
    SchNet~\citep{schnet} &
      - &
      2.77\% &
      4266.13 &
      3.91\% &
      2155.36 &
      4.32\% &
      1458.77 &
      4.73\% &
      1104.88 &
      5.04\% &
      892.79 \\
    DimeNet++~\citep{dimenet_pp} &
      - &
      5.34\% &
      4271.23 &
      7.61\% &
      2149.78 &
      8.84\% &
      1435.21 &
      10.07\% &
      1081.96 &
      10.79\% &
      865.20 \\
    PaiNN~\citep{painn} &
      - & 
      27.44\% &
      4089.77 &
      33.61\% &
      2077.65 &
      36.69\% &
      1395.55 &
      38.64\% &
      1048.63 &
      39.57\% &
      840.44 \\
    GemNet-OC~\citep{gasteiger_gemnet_oc_2022} &
      9.4 &
      68.76\% &
      4185.18 &
      77.29\% &
      2087.11 &
      80.78\% &
      1392.51 &
      81.50\% &
      1046.85 &
      82.94\% &
      840.25 \\
    GemNet-OC-MD~\citep{gasteiger_gemnet_oc_2022} &
      9.4 & 
      68.76\% &
      4182.04 &
      78.21\% &
      2092.27 &
      81.81\% &
      1404.11 &
      83.25\% &
      1053.36 &
      84.38\% &
      841.64 \\
    GemNet-OC-MD-Large~\citep{gasteiger_gemnet_oc_2022} &
      45.0 &
      73.18\% &
      4078.76 &
      79.65\% &
      2065.15 &
      83.25\% &
      1381.39 &
      85.41\% &
      1041.50 &
      86.02\% &
      834.46 \\
    SCN-MD-Large~\citep{scn} &
      120.0 & 
      77.80\% &
      3974.21 &
      84.28\% &
      1989.32 &
      86.33\% &
      1331.43 &
      87.36\% &
      1004.40 &
      87.77\% &
      807.00 \\
    EquiformerV2 ($\lambda_E = 4$, $31$M) &
      12.2 &
      84.17\% &
      3983.41 & 
      87.15\% &
      1992.64 &
      87.87\% &
      1331.35 &
      88.69\% &
      1000.62 &
      89.31\% &
      802.95 \\
    EquiformerV2 ($\lambda_E = 4$, $153$M) &
      45.0 & 
      \textbf{85.41\%} &
      4001.71 &
      \textbf{88.90\%} &
      2012.47 &
      \textbf{90.54\%} &
      1352.08 &
      \textbf{91.06\%} &
      1016.31 &
      \textbf{91.57\%} &
      815.87 \\

    \bottomrule[1.2pt]
      
    \end{tabular}%
    }
    \vspace{1mm}
    \caption{AdsorbML results with EquiformerV2 ($\lambda_{E} = 4$, $31$M) and ($\lambda_{E} = 4$, $153$M) trained on S2EF-All+MD from \tableref{tab:oc20_all_results}.
    We visualize the speed-accuracy trade-offs of different models in Figure~\ref{appendix:fig:adsorbml_speed_sucess}.
    }
    %\vspace{-10pt}
    \label{tab:adsorbml}
\end{table*}
%##############################################################################

%% file: content/5_2_oc22.tex
\begin{table}[t!]
    \centering
    \resizebox{1.0\textwidth}{!}{%
    \begin{tabular}{llcrcccccccccc}
    \toprule[1.2pt]
    & & & & \multicolumn{4}{c}{S2EF-Total validation} & \multicolumn{4}{c}{S2EF-Total test} & \multicolumn{2}{c}{IS2RE-Total test} \\ 
    \cmidrule(lr){5-8} \cmidrule(lr){9-12} \cmidrule(lr){13-14}
    & & & \multicolumn{1}{c}{Number of} & \multicolumn{2}{l}{Energy MAE (meV)}$\downarrow$& \multicolumn{2}{l}{Force MAE (meV/Å)}$\downarrow$ & \multicolumn{2}{l}{Energy MAE (meV)}$\downarrow$& \multicolumn{2}{l}{Force MAE (meV/Å)}$\downarrow$ & \multicolumn{2}{l}{Energy MAE (meV)}$\downarrow$\\
    %\cmidrule(lr){4-5} \cmidrule(lr){6-7} \cmidrule(lr){8-9} \cmidrule(lr){10-11}
    Model                                           & Training Set & Linear reference & \multicolumn{1}{c}{parameters} & ID                & OOD              & ID                & OOD & ID & OOD & ID & OOD & ID & OOD              \\
    \midrule[1.2pt]
    GemNet-OC~\citep{gasteiger_gemnet_oc_2022}                                       & OC22         &              & 39M & 545 & 1011 & 30 & 40 & 374               & 829              & 29.4              & 39.6 & 1329 & 1584 \\
    GemNet-OC~\citep{gasteiger_gemnet_oc_2022}                                       & OC22         & \checkmark               & 39M & - & - & - & - & 357               & 1057             & 30.0              & 40.0 & - & -              \\
    GemNet-OC~\citep{gasteiger_gemnet_oc_2022}                                       & OC22 + OC20  & & 39M & 464 & 859 & 27 & 34 & 311               & 689              & 26.9              & 34.2 & 1200 & 1534 \\
    eSCN~\citep{escn}                                            & OC22         & \checkmark & 200M               & - & - & - & -  & 350               & 789              & 23.8              & 35.7 & - & - \\
    \midrule
    EquiformerV2 ($\lambda_E = 1, \lambda_F = 1$)   & OC22         & \checkmark                & 122M & \textbf{343} & \textbf{580} & 24.42 & 34.31 & \textbf{182.8}             & 677.4            & 22.98             & 35.57 & \textbf{1084} & 1444 \\
    %EquiformerV2 ($\lambda_E = 1, \lambda_F = 10$)  & OC22         & \checkmark                & 243.4             & 669.2            & 21.75             & 33.23             \\
    EquiformerV2 ($\lambda_E = 4, \lambda_F = 100$) & OC22         & \checkmark               & 122M & 433 & 629 & \textbf{22.88} & \textbf{30.70} & 263.7             & \textbf{659.8}            & \textbf{21.58}             & \textbf{32.65} & 1119 & \textbf{1440} \\
    \bottomrule[1.2pt]
    \end{tabular}}
    \vspace{1mm}
    \caption{OC22 results on S2EF-Total validation and test splits and IS2RE-Total test split.}
    \label{tab:oc22_results}
    %\vspace{-20pt}
\end{table}

%\vspace{-4mm}
\subsection{OC22 Dataset}
\label{subsec:oc22}

%\vspace{-3mm}
\paragraph{Dataset.}
%\todo{The Open Catalyst 2022 (OC22) dataset~\citep{oc22} focuses on oxide electrocatalysis and consists of about $62$k DFT relaxations obtained with Perdew-Burke-Ernzerhof (PBE) functional calculations.}
The OC22 dataset~\citep{oc22} focuses on oxide electrocatalysis.
%
%One crucial difference in OC22 compared to OC20 is that the energies in OC22
%are all DFT total energies, instead of adsorption energies.
%
One crucial difference between OC22 and OC20 is that the energies in OC22 are DFT total energies. %, which are the most general and closest to a DFT surrogate instead of adsorption energies.
DFT total energies are harder to predict but are the most general and closest to a DFT surrogate, offering the flexibility to study property prediction beyond adsorption energies.
%Analogous to the task definitions in OC20, the primary tasks in OC22 are
%S2EF-Total and IS2RE-Total.
Similar to OC20, the tasks in OC22 are S2EF-Total and IS2RE-Total.
%
%We train models on the OC22 S2EF-Total dataset, which has $8.2$M structures,
%and evaluate them on energy and force MAE on the S2EF-total validation and test
%splits.
We train models on the OC22 S2EF-Total dataset and evaluate them on energy and force MAE on the S2EF-Total validation and test splits. 
We use the trained models to perform structural relaxations and predict relaxed energy.
Relaxed energy predictions are evaluated on the IS2RE-Total test split.

%\vspace{-3mm}
\paragraph{Training Details.}
Please refer to Section~\ref{appendix:subsec:oc22_training_details} for details on architectures, hyper-parameters and training time.

%\vspace{-3mm}
\paragraph{Results.}
We train two EquiformerV2 models with different energy coefficients $\lambda_E$ and force coefficients $\lambda_F$.
We follow the practice of OC22 models and use linear reference~\citep{oc22}.
The results are summarized in Table~\ref{tab:oc22_results}.
EquiformerV2 improves upon previous models on all the tasks.
EquiformerV2 ($\lambda_E = 4$, $\lambda_F = 100$) trained on only OC22 achieves better results on all the tasks than GemNet-OC trained on both OC20 and OC22.
We note that OC22 contains about $8.4$M structures and OC20 contains about $130$M structures, and therefore EquiformerV2 demonstrates significatly better data efficiency.
\revision{Additionally, the performance gap between eSCN and EquiformerV2 is larger than that on OC20, suggesting that more complicated structures can benefit more from the proposed architecture.}
%\todo{Add explicit numbers}
When trained on OC20 S2EF-All+MD, EquiformerV2 ($\lambda_E=4$, $153$M) improves upon eSCN by $4\%$ on energy MAE and $9\%$ on force MAE.
For OC22, EquiformerV2 ($\lambda_E = 4$, $\lambda_F = 100$) improves upon eSCN by $18.9\%$ on average energy MAE and $8.9\%$ on average force MAE.
%, and EquiformerV2 ($\lambda_E = 4$, $\lambda_F = 100$) improves by $18.9\%$ on energy MAE and $8.9\%$ on force MAE.

%% file: content/5_3_comparison_with_equiformer.tex
%\vspace{-3mm}
\subsection{Comparison with Equiformer}
\label{subsec:comparison_with_equiformer}

\begin{table}[t]
%\begin{adjustwidth}{-2.5 cm}{-2.5 cm}
\centering
%\scalebox{0.6}{
\resizebox{1.0\textwidth}{!}{
\begin{tabular}{llcccccccccccc}
%\cline{7-11}
\toprule[1.2pt]
& Task & $\alpha$ & $\Delta \varepsilon$ & $\varepsilon_{\text{HOMO}}$ & $\varepsilon_{\text{LUMO}}$ & $\mu$ & $C_{\nu}$ & $G$ & $H$ & $R^2$ & $U$ & $U_0$ & ZPVE \\ 
%Methods & Units & bohr$^3$ & meV & meV & meV & D & cal/mol K & meV & meV & bohr$^3$ & meV & meV & meV\\
Model & Units & $a_0^3$ & meV & meV & meV & D & cal/mol K & meV & meV & $a_0^2$ & meV & meV & meV\\
\midrule[1.2pt]

%NMP~\citep{neural_message_passing_quantum_chemistry}$^{\dagger}$ & & .092 & 69 & 43 & 38 & .030 & .040 & 19 & 17 & .180 & 20 & 20 & 1.50 \\
%SchNet~\citep{schnet} & & .235 & 63 & 41 & 34 & .033 & .033 & 14 & 14 & .073 & 19 & 14 & 1.70 \\
%Cormorant~\citep{cormorant}$^{\dagger}$ & & .085 & 61 & 34 & 38 & .038 & .026 & 20 & 21 & .961 & 21 & 22 & 2.03 \\
%LieConv~\citep{lieconv}$^{\dagger}$ & & .084 & 49 & 30 & 25 & .032 & .038 & 22 & 24 & .800 & 19 & 19 & 2.28 \\
DimeNet++~\citep{dimenet_pp} & & \textbf{.044} & 33 & 25 & 20 & .030 & .023 & 8 & 7 & .331 & 6 & 6 & 1.21 \\
%TFN~\citep{tfn}$^{\dagger}$ & & .223 & 58 & 40 & 38 & .064 & .101 & - & - & - & - & - & - \\
%\multicolumn{2}{l}{SE(3)-Transformer~\citep{se3_transformer}$^{\dagger}$} & .142 & 53 & 35 & 33 & .051 & .054 & - & - & - & - & - & - \\
EGNN~\citep{egnn}$^{\dagger}$ & & .071 & 48 & 29 & 25 & .029 & .031 & 12 & 12 & .106 & 12 & 11 & 1.55 \\

PaiNN~\citep{painn} & & .045 & 46 & 28 & 20 & .012 & .024 & \textbf{7.35} & \textbf{5.98} & .066 & \textbf{5.83} & \textbf{5.85} & 1.28 \\

TorchMD-NET~\citep{torchmd_net} & & .059 & 36 & 20 & 18 & .011 & .026 & 7.62 & 6.16 & \textbf{.033} & 6.38 & 6.15 & 1.84 \\

SphereNet~\citep{spherenet} & & .046 & 32 & 23 & 18 & .026 & \textbf{.021} & 8 & 6 & .292 & 7 & 6 & \textbf{1.12} \\

SEGNN~\citep{segnn}$^{\dagger}$ & & .060 & 42 & 24 & 21 & .023 & .031 & 15 & 16 & .660 & 13 & 15 & 1.62 \\

%Equiformer& 0.5093& 0.6285 & 0.5053 & 0.5556 & 0.5497 & 5.04 & 2.85 & 4.90 & 3.04 & \\
EQGAT~\citep{eqgat} & & .053 & 32 & 20 & 16 & .011 & .024 & 23 & 24 & .382 & 25 & 25 & 2.00 \\

\midrule

Equiformer~\citep{equiformer} & & .046 & 30 & 15 & 14 & .011 & .023 & 7.63 & 6.63 & .251 & 6.74 & 6.59 & 1.26 \\

EquiformerV2 & & .050          & \textbf{29}       & \textbf{14}        & \textbf{13}        & \textbf{.010} & .023 & 7.57 & 6.22 & .186 &  6.49 & 6.17 & 1.47 \\
%\midrule[0.6pt]
%Equiformer + NAT & 0.4737 & 0.7245 & 0.4862 & 0.6493 & 0.5834 & 6.01 & 2.45 & 5.41 & 2.71 & \\
\bottomrule[1.2pt]

\end{tabular}
}
\vspace{1mm}
%\end{adjustwidth}
\caption{
Mean absolute error results on QM9 test set.  
%\note{Dropout rate = 0.1.}
%%%$\dagger$ denotes using different training, validation, testing data partitions. % as mentioned in SEGNN~\cite{segnn}.
$\dagger$ denotes using different data partitions.
%$\ddagger$ denotes results from SE(3)-Transformer~\cite{se3_transformer}.
}
%\vspace{-10pt}
\label{tab:qm9_results}
\end{table}

\begin{table}[t!]
    \centering
    \scalebox{0.6}{%
    \begin{tabular}{lcccc}
    \toprule[1.2pt]
    Model & $L_{max}$ & Energy MAE (meV)$\downarrow$ & Force MAE (meV/Å)$\downarrow$ & Training time (GPU-hours)$\downarrow$ \\
    \midrule[1.2pt]    
    Equiformer~\citep{equiformer} & $2$ & 297 & 27.57 & 1365 \\
    Equiformer~\citep{equiformer} & $3$ & OOM & OOM & OOM \\
    EquiformerV2 & $2$ & 298 & 26.24 & 600 \\
    EquiformerV2 & $4$ & 284 & 21.37 & 966 \\
    EquiformerV2 & $6$ & 285 & 20.46 & 1412 \\
    \bottomrule[1.2pt]
    \end{tabular}}
    \vspace{1mm}
    \caption{Comparison on OC20 S2EF-2M dataset.
    Errors are averaged over the four validation sub-splits.
    ``OOM'' denotes out-of-memory error, and we cannot use $L_{max} = 3$ for Equiformer.}
    \label{tab:oc20_s2ef_equiformer}
    %\vspace{-20pt}
\end{table}

%\vspace{-3mm}
\paragraph{QM9 Dataset.}
We follow the setting of Equiformer and train EquiformerV2 on the QM9 dataset~\citep{qm9_1, qm9_2} and summarize the results in Table~\ref{tab:qm9_results}.
%We note that the performance gain of using higher degrees and the improved architecture is not as significant as that on OC20 and OC22 datasets.
The performance gain of using higher degrees and the improved architecture is not as significant as that on OC20 and OC22 datasets.
This, however, is not surprising.
The training set of QM9 contains only $110$k examples, which is much less than OC20 S2EF-2M with $2$M examples and OC22 with $8.4$M examples.
\revision{Moreover, QM9 has much less numbers of atoms in each example and much less diverse atom types, and each example has less angular variations.}
%These make better expressivity of higher degrees and improved architectures less helpful.
Nevertheless, EquiformerV2 achieves better results than Equiformer on 9 out of the 12 tasks and is therefore the overall best performing model.
We additionally train EquiformerV2 with Noisy Nodes~\citep{noisy_nodes} to better understand the gain of higher degrees in Sec.~\ref{appendix:subsec:qm9_additional_results}.

%\vspace{-3mm}
\paragraph{OC20 S2EF-2M Dataset.}
We use similar configurations (e.g., numbers of blocks and numbers of channels) and train Equiformer on OC20 S2EF-2M dataset for the same number of epochs as training EquiformerV2.
We vary $L_{max}$ for both Equiformer and EquiformerV2 and compare the results in Table~\ref{tab:oc20_s2ef_equiformer}.
For $L_{max} = 2$, EquiformerV2 is $2.3 \times$ faster than Equiformer since EquiformerV2 uses eSCN convolutions for efficient $SO(3)$ convolutions.
Additionally, EquiformerV2 achieves better force MAE and similar energy MAE, demonstrating the effectiveness of the proposed improved architecture.
For $L_{max} > 2$, we encounter out-of-memory errors when training Equiformer even after we reduce the number of blocks and use the batch size $=1$.
In contrast, We can easily train EquiformerV2 with $L_{max}$ up to $6$.
When increasing $L_{max}$ from $2$ to $4$, EquiformerV2 achieves lower energy MAE and significantly lower force MAE than Equiformer and requires $1.4 \times$ less training time.
%The comparison suggests that more complicated datasets benefit more from more expressive models, enabling better performance and lower computational cost. \todo{[simplify this]}
The comparison suggests that complicated datasets have more performance to gain from using more expressive models, enabling better performance and lower computational cost. 

%\vspace{-3mm}
\paragraph{Discussion.}
One limitation of EquiformerV2 is that the performance gains brought by scaling to higher degrees and the proposed architectural improvements can depend on datasets and tasks.
For small datasets like QM9, the performance gain is not significant.
We additionally compare Equiformer and EquiformerV2 on OC20 IS2RE dataset in Sec.~\ref{appendix:subsec:additional_comparison_on_oc20_is2re}.
For different tasks, the improvements are also different, and force predictions benefit more from better expressivity than energy predictions.
We note that the first issue can be mitigated by first pre-training on large datasets like OC20 and PCQM4Mv2~\citep{pcqm4mv2} optionally via denoising~\citep{noisy_nodes, pretraining_via_denoising} and then fine-tuning on smaller datasets. 
The second issue might be mitigated by combining DFT with ML models. 
For example, AdsorbML uses ML forces for structural relaxations and a single-point DFT for calculating the final relaxed energies.

%% file: content/6_conclusion.tex
%\vspace{-5mm}
\section{Conclusion}
%\vspace{-4mm}

%In this work, we investigate how equivariant Transformers can scale up to higher degrees of equivariant representations.
In this work, we investigate how equivariant Transformers can be scaled up to higher degrees of equivariant representations.
%We replace $SO(3)$ convolutions in Equiformer with eSCN convolutions to efficiently include higher-order tensors.
%However, simply incorporating eSCN convolutions into Equiformer does not improve upon the original eSCN model.
%Therefore, to better leverage the power of higher degrees, we propose three architectural improvements -- attention re-normalization, separable $S^2$ activation and separable layer normalization.
We start by replacing $SO(3)$ convolutions in Equiformer with eSCN convolutions, and then we propose three architectural improvements to better leverage the power of higher degrees -- attention re-normalization, separable $S^2$ activation and separable layer normalization.
With these modifications, we propose EquiformerV2, which outperforms state-of-the-art methods on all the tasks on the OC20 and OC22 datasets, improves speed-accuracy trade-offs, and achieves the best success rate when used in AdsorbML.
\revision{We also compare EquiformerV2 with Equiformer to better understand the performance gain brought by higher degrees and the improved architecture.}

%We discuss broader impacts in Sec.~\ref{appendix:sec:broader_impacts} and limitations in Sec.~\ref{appendix:sec:limitations}.

\section{Ethics Statement}
EquiformerV2 achieves more accurate approximation of quantum mechanical calculations and demonstrates one further step toward being able to replace DFT compute force fields with machine learned ones for practical applications in chemistry and material science.
%
%By demonstrating promising results to encourage positive usage of the method and creating datasets for beneficial applications, we believe that there is much more to be gained by harnessing the ability for applications like material design and drug discovery than to be lost by using the method for adversarial purposes.
We hope these promising results will encourage the community to make further progress in applications like material design and drug discovery, rather than use these methods for adversarial purposes.
We note that these methods only facilitate the identification of molecules or materials with specific properties; there remain substantial hurdles to synthesize and deploy such molecules or materials at scale.
Finally, we note that the proposed method is general and can be applied to different problems like protein structure prediction~\citep{equifold} as long as inputs can be modeled as 3D graphs. 
%Additionally, EquiformerV2 may be adapted to make predictions of other quantum mechanical properties that require higher degree representations such as molecular Hamiltonians and charge density predictions using spherical harmonic basis sets.

\section{Reproducibility Statement }

We include details on architectures, hyper-parameters and training time in Sec.~\ref{appendix:subsec:training_details} (OC20), Sec.~\ref{appendix:subsec:oc22_training_details} (OC22) and Sec.~\ref{appendix:subsec:qm9_training_details} (QM9).

%We submit our code reproducing the results of EquiformerV2 trained on OC20 S2EF-2M dataset.
The code for reproducing the results of EquiformerV2 trained on OC20 S2EF-2M and QM9  datasets is available at \href{https://github.com/atomicarchitects/equiformer_v2}{{\tt \color{crimson} https://github.com/atomicarchitects/equiformer\_v2}}.
%\href{https://github.com/atomicarchitects/equiformer_v2}{\color{crimson}{https://github.com/atomicarchitects/equiformer_v2}}.

%Following the author guide, after the discussion forums are opened for all submitted papers, we will make a comment directed to the reviewers and area chairs and put a link to an anonymous repository. 

%\vspace{-2mm}
%\paragraph{Limitations.}
%Although EquiformerV2 improves upon state-of-the-art methods on the large and diverse OC20 dataset, we acknowledge that the performance gains brought by scaling to higher degrees and the proposed architectural improvements can depend on tasks and datasets.
%For example, the increased expressivity may lead to overfitting on smaller datasets like QM9~\cite{qm9_1, qm9_2} and MD17~\cite{md17_1, md17_2, md17_3}.
%However, the issue can be mitigated by pre-training on large datasets like OC20~\cite{oc20} and PCQM4Mv2~\cite{pcqm4mv2} optionally via denoising~\cite{pretraining_via_denoising} and then finetuning on smaller datasets. 

\section*{Acknowledgement}
We thank Larry Zitnick and Saro Passaro for helpful discussions.
We also thank Muhammed Shuaibi for helping with the DFT evaluations for AdsorbML~\citep{adsorbml}. 
We acknowledge the MIT SuperCloud and Lincoln Laboratory Supercomputing Center~\citep{supercloud} for providing high performance computing and consultation resources that have contributed to the research results reported within this paper.

Yi-Lun Liao and Tess Smidt were supported by DOE ICDI grant DE-SC0022215.

%% file: content/8_appendix.tex
\appendix
\section*{Appendix}

\begin{itemize}
    \item[] \ref{appendix:sec:additional_background}\quad Additional background
    \begin{enumerate}
        \item[] \ref{appendix:subsec:group_theory}\quad Group theory
        \item[] \ref{appendix:subsec:equivariance}\quad Equivariance
        \item[] \ref{appendix:subsec:escn_convolution}\quad eSCN convolution
        \item[]\ref{appendix:subsec:s2_activation}\quad $S^2$ activation
    \end{enumerate}

    \item[]\ref{sec:related_work}\quad Related works
    \begin{enumerate}
        \item[] \ref{subsec:equivariant_networks}\quad \textit{SE(3)/E(3)}-equivariant GNNs
        \item[] \ref{subsec:invariant_networks}\quad Invariant GNNs
    \end{enumerate}
    
    \item[] \ref{appendix:sec:details_of_architecture}\quad Details of architecture
    
    \item[] \ref{appendix:sec:details_of_experiments_on_oc20}\quad Details of experiments on OC20
    \begin{enumerate}
        \item[] \ref{appendix:subsec:detailed_description_of_oc20_dataset}\quad Detailed description of OC20 dataset
        \item[] \ref{appendix:subsec:training_details}\quad Training details
        \item[] \ref{appendix:subsec:details_of_running_relaxations}\quad Details of running relaxations
        \item[] \ref{appendix:subsec:details_of_adsorbml}\quad Details of AdsorbML
        \item[]
        \ref{appendix:subsec:additional_comparison_on_oc20_is2re}\quad Additional comparison with Equiformer on OC20 IS2RE
    \end{enumerate}

    \item[] \ref{appendix:sec:details_of_experiments_on_oc22}\quad Details of experiments on OC22
    \begin{enumerate}
        \item[] \ref{appendix:subsec:oc22_training_details}\quad Training details
    \end{enumerate}

    \item[] \ref{appendix:sec:details_of_experiments_on_qm9}\quad Details of experiments on QM9
    \begin{enumerate}
        \item[] \ref{appendix:subsec:qm9_additional_results}\quad Additional results of training with Noisy Nodes
        \item[] \ref{appendix:subsec:qm9_training_details}\quad Training details
        \item[] \ref{appendix:subsec:qm9_ablation_study}\quad Ablation study on architectural improvements
    \end{enumerate}
    
\end{itemize}

\input{content/8_1_additional_background}

\input{content/4_related_works}
\input{content/8_2_details_of_architecture}

\input{content/8_3_details_of_experiments_on_oc20}
\input{content/8_4_details_of_experiments_on_oc22}
\input{content/8_5_details_of_experiments_on_qm9}

%% file: content/8_1_additional_background.tex
\section{Additional Background}
\label{appendix:sec:additional_background}

%\todo{mention that eSCN convolution and this work mainly deal with $SO(3)$ and do not consider inversion}

%We provide more details on background here, and we note that they are introduced in a way that facilitates discussing the proposed method and that other works~\cite{tfn, 3dsteerable, kondor2018clebsch, cormorant, se3_transformer, segnn} also provide relevant mathematical background, which helps the presentation of this work.

%%%In this section, we provide additional mathematical background on group equivariance helpful for the discussion of the proposed method.
%In this section, we provide additional mathematical background helpful for the discussion of the proposed method.
%Other works~\cite{tfn, 3dsteerable, kondor2018clebsch, cormorant, se3_transformer, segnn} also provide similar background.
%We encourage interested readers to see these works~\cite{zee, Dresselhaus2007} for more in-depth and pedagogical presentations.

We first provide relevant mathematical background on group theory and equivariance.
We note that most of the content is adapted from Equiformer~\citep{equiformer} and that these works~\citep{zee, Dresselhaus2007} have more in-depth and pedagogical discussions.
Then, we provide mathematical details of eSCN convolutions.

\subsection{Group Theory}
\label{appendix:subsec:group_theory}

\paragraph{Definition of Groups.}
%A group is an algebraic structure that consists of a set $G$ and a binary operator $\circ: G \times G \rightarrow G$ and is typically denoted as $G$.
A group is an algebraic structure that consists of a set $G$ and a binary operator $\circ: G \times G \rightarrow G$.
Typically denoted as $G$, groups satisfy the following four axioms:
\begin{enumerate}
    \item Closure: $g \circ h \in G$ for all $g, h \in G$.
    \item Identity: There exists an identity element $e \in G$ such that $g \circ e = e \circ g = g$ for all $g \in G$.
    \item Inverse: For each $g \in G$, there exists an inverse element $g^{-1} \in G$ such that $g \circ g^{-1} = g^{-1} \circ g = e$.
    \item Associativity: $g \circ h \circ i = (g \circ h) \circ i = g \circ (h \circ i)$ for all $g, h, i \in G$.
\end{enumerate}

%In this work, we focus on 3D rotation, translation and inversion.
%\revision{
%Equivariant neural networks focus on 3D rotation, translation and optionally inversion.
%}
In this work, we consider 3D Euclidean symmetry, and relevant groups are:
\begin{enumerate}
    \item The Euclidean group in three dimensions $E(3)$: 3D rotation, translation and inversion.
    \item The special Euclidean group in three dimensions $SE(3)$: 3D rotation and translation.
    \item The orthogonal group in three dimensions  $O(3)$: 3D rotation and inversion.
    \item The special orthogonal group in three dimensions  $SO(3)$: 3D rotation.
    %\item The translational group $T(3)$: 3D translation.
\end{enumerate}

Since eSCN~\citep{escn} and this work only consider equivariance to 3D rotation and invariance to 3D translation but not inversion, we mainly discuss $SE(3)$-equivariance in the main text and in appendix and note that more details of $E(3)$-equivariance can be found in the work of Equiformer~\citep{equiformer}.

\paragraph{Group Representations.}

Given a vector space $X$, the way a group $G$ acts on $X$ is given by the group representation $D_X$. 
$D_X$ is parameterized by $g \in G$, with $D_{X}(g): X \rightarrow X$.
Group representations $D_X$ are invertible matrices, and group transformations, or group actions, take the form of matrix multiplications.
%Groups can be used to define transformations.
%The actions of groups define transformations.
%\revision{Groups can be used to describe transformations on vector spaces.}
%Formally, a transformation acting on vector space $X$ parametrized by group element $g \in G$ is an injective function $T_{g}: X \rightarrow X$.
%A powerful result of group representation theory is that these transformations can be expressed as matrices which act on vector spaces via matrix multiplication.
%\revision{More importantly, these transformations can be defined as matrices which act on vector spaces via matrix multiplication.}
%These matrices are called the group representations.
%\revision{We call these matrices the group representations.}
%Formally, a group representation $D: G \rightarrow GL(N)$ is a mapping between a group $G$ and a set of $N \times N$ invertible matrices.
%Formally, a group representation $D: G \rightarrow GL(N)$ \revision{maps} between a group $G$ and a set of $N \times N$ invertible matrices.
%The group representation $D(g): X \rightarrow X$ maps an $N$-dimensional vector space $X$ onto itself and satisfies  $D(g)D(h) = D(g \circ h)$ for all $g, h \in G$.
%\revision{Given a group element $g \in G$}, the group representation $D(g): X \rightarrow X$ maps \revision{the} $N$-dimensional vector space $X$ onto itself and satisfies \revision{that} $D(g)D(h) = D(g \circ h)$ for all $g, h \in G$.
This definition of group representations satisfies the requirements of groups, including associativity, $D(g)D(h) = D(g \circ h)$ for all $g, h \in G$.
%
%There are many ways to represent a group.
%If there exists a change of basis $P$ in the form of an $N \times N$ matrix for a given representation $D(g)$ such that $P^{-1} D(g) P = D'(g)$
%How a group is represented depends on the vector space it acts on.
%If there exists a change of basis $P$ in the form of an $N \times N$ matrix such that $P^{-1} D(g) P = D'(g)$ for all $g \in G$, then we say the two group representations are equivalent.
We say that the two group representations $D(g)$ and $D'(g)$ are equivalent if there exists a change-of-basis $N \times N$ matrix $P$ such that $P^{-1} D(g) P = D'(g)$ for all $g \in G$.
%\todo{Add a short descript on the change-of-basis matrix.}
%\revision{The group element $g$ acts on independent subspaces of the vector space if $D'(g)$ is block diagonal.}
$D(g)$ is reducible if $D'(g)$ is block diagonal for all $g \in G$, meaning that $D'(g)$ acts on multiple independent subspaces of the vector space. 
Otherwise, the representation $D(g)$ is said to be irreducible.
%Reducible representations can be decomposed into a direct sum (concatenation) of irreducible representations.
%\revision{In that case, we say that the representation $D(g)$ is reducible since it can be reduced to a block diagonal matrix and that $D'(g)$ is irreducible.}
%If $D'(g)$ is block diagonal, which means that $g$ acts on independent subspaces of the vector space, the representation $D(g)$ is reducible.
%that makes $D'(g)$ block diagonal for all $g \in G$, meaning $g$ acts on independent subspaces of the vector space, the representation $D(g)$ is deemed reducible.
%A particular class of representations that are convenient for generalizable and composable functions are irreducible representations or ``irreps'', which cannot be further reduced.
%A particular class of representations that are convenient for composable functions are irreducible representations or ``irreps'', which cannot be further reduced.
Irreducible representations, or irreps, are a class of representations that are convenient for composing different group representations.
%We can express any group representation of $SO(3)$ as a direct sum (concatenation) of irreps \cite{zee,Dresselhaus2007,e3nn}:
%\begin{equation}
%D(g) = P ^{-1} \left ( \bigoplus_{i} D_{l_i}(g) \right)  P =
%P^{-1}
%\begin{pmatrix}
%D_{l_0}(g) & & \\
%& D_{l_1}(g) & & \\
%& & ...... \\
%\end{pmatrix}
%P
%\end{equation}
%where $D_{l_i}(g)$ are Wigner-D matrices with degree $l_i$ as mentioned in Sec.~\ref{subsec:irreducible_representations}.
Specifically, for the case of $SO(3)$, Wigner-D matrices are irreducible representations, and we can express any group representation of $SO(3)$ as a direct sum (concatentation) of Wigner-D matrices~\citep{zee,Dresselhaus2007,e3nn}:
\begin{equation}
D(g) = P ^{-1} \left ( \bigoplus_{i} D^{(L_i)}(g) \right)  P =
P^{-1}
\begin{pmatrix}
D^{(L_0)}(g) & & \\
& D^{(L_1)}(g) & & \\
& & \ddots \\
\end{pmatrix}
P
\end{equation}
where $D^{(L_i)}(g)$ are Wigner-D matrices of degree $L_i$.

% already mention in background
%\paragraph{Irreducible Representations of $SO(3)$.}

\begin{comment}
\subsection{$SO(3)$ Tensor products}
\label{appendix:subsec:tensor_products}
We use the term $SO(3)$ tensor products to refer to two combined operations: the multiplication of two vector spaces $X \times Y \rightarrow Z$ and the decomposition of the resulting vector space into irreducible representations. For example, if the group representation on vector space $X$, $D_X$, and vector space $Y$, $D_Y$ are both irreducible, their tensor product is generally reducible, i.e. $D_Z$ is generally a direct sum (concatenation) of irreducible representations. In the case of $SO(3)$, 
the change of basis to go from the direct product of $D_X \times D_Y \rightarrow D_Z$ which is generally reducible and not block diagonal to $D_{Z'}$ which is reducible and block diagonal are called the Clebsch-Gordan coefficients. The Clebsch-Gordan coefficients $C^{(L_3, m_3)}_{(L_1, m_1) (L_2, m_2)}$.  give the selection rules for $SO(3)$, i.e. they convey which irreps are produced when two irreps are producted. Thus, the coefficients depend on the irrep indices of the two incoming irreps $(L_1, m_1)$ and $(L_2, m_2)$ and the outgoing irrep $(L_3, m_3)$.
\end{comment}

\subsection{Equivariance}
\label{appendix:subsec:equivariance}

A function $f$ mapping between vector spaces $X$ and $Y$ is equivariant to a group of transformations $G$ if for any input $x \in X$, output $y \in Y$ and group element $g \in G$, we have $f(D_X(g) x) = D_Y(g) f(x) = D_Y(g) y$, where $D_X(g)$ and $D_Y(g)$ are transformation matrices or group representations parametrized by $g$ in $X$ and $Y$.
Additionally, $f$ is invariant when $D_Y(g)$ is an identity matrix for any $g \in G$.

%\paragraph{Definition of Equivariance and Invariance.}
%Equivariance is a property of a function $f: X \rightarrow Y$ mapping between vector spaces $X$ and $Y$.
%Given a group $G$ and group representations $D_{X}(g)$ and $D_{Y}(g)$ in input and output spaces $X$ and $Y$, $f$ is equivariant to G if $D_Y(g) f(x) = f(D_X(g)x)$ for all $x \in X$ and $g \in G$.
%Invariance corresponds to the case that $f(x) = D_Y(g) f(x) = f(D_X(g) x)$.
%Invariance corresponds to the case where $D_Y(g)$ is the identity $I$ for all $g \in G$.

%The outputs and intermediate features of group equivariant neural networks are designed to be equivariant to data transformation defined by a group.
As neural networks comprise many composable operations, equivariant neural networks comprise many equivariant operations to maintain the equivariance of input, intermediate, and output features.
Incorporating equivariance as a strong prior knowledge can help improve data efficiency and generalization of neural networks~\citep{nequip, quantum_scaling, neural_scale_of_chemical_models}.
In this work, we achieve equivariance to 3D rotation by operating on vector spaces of $SO(3)$ irreps, incorporate invariance to 3D translation by acting on relative positions, but do not consider inversion.

\subsection{eSCN Convolution}
\label{appendix:subsec:escn_convolution}

Message passing is used to update equivariant irreps features and is typically implemented as $SO(3)$ convolutions. 
A traditional $SO(3)$ convolution interacts input irrep features $x_{m_i}^{(L_i)}$ and spherical harmonic projections of relative positions $Y_{m_f}^{(L_f)}(\vec{r}_{ts})$ with an $SO(3)$ tensor product with Clebsch-Gordan coefficients $C^{(L_o, m_o)}_{(L_i, m_i), (L_f, m_f)}$.
Since tensor products are compute-intensive, eSCN convolutions~\citep{escn} are proposed to reduce the complexity of tensor products when they are used in $SO(3)$ convolutions.
Rotating the input irreps features $x_{m_i}^{(L_i)}$ based on the relative position vectors $\vec{r}_{ts}$ simplifies the tensor products and enables reducing $SO(3)$ convolutions to $SO(2)$ linear operations.
Below we provide the mathematical details of $SO(3)$ convolutions built from tensor products and how rotation can reduce their computational complexity.

Tensor products interact type-$L_i$ vector $x^{(L_i)}$ and type-$L_f$ vector $f^{(L_f)}$ to produce type-$L_o$ vector $y^{(L_o)}$ with Clebsch-Gordan coefficients $C^{(L_o, m_o)}_{(L_i, m_i), (L_f, m_f)}$.
Clebsch-Gordan coefficients $C^{(L_o, m_o)}_{(L_i, m_i), (L_f, m_f)}$ are non-zero only when $| L_i - L_o | \leq L_f \leq | L_i + L_o |$.
Each non-trivial combination of $L_i \otimes L_f \rightarrow L_o$ is called a path, and each path is independently equivariant and can be assigned a learnable weight $w_{L_i, L_f, L_o}$.

We consider the message $m_{ts}$ sent from source node $s$ to target node $t$ in an $SO(3)$ convolution.
The $L_o$-th degree of $m_{ts}$ can be expressed as:
\begin{equation}
m^{(L_o)}_{ts} = \sum_{L_{i}, L_{f}} w_{L_i, L_f, L_o} \left(x_s^{(L_i)} \otimes^{L_o}_{L_i, L_f} Y^{(L_f)}(\hat{r}_{ts}) \right)
%h^{(L_3)}_{m_3} = (f^{(L_1)} \otimes g^{(L_2)})_{m_3} = \sum_{m_1 = -L_1}^{L_1} \sum_{m_2 = -L_2}^{L_2} C^{(L_3, m_3)}_{(L_1, m_1)(L_2, m_2)} f^{(L_1)}_{m_1} g^{(L_2)}_{m_2}
\label{appendix:eq:so3_convolution}
\end{equation}
where $x_s$ is the irreps feature at source node $s$, $x_s^{(L_i)}$ denotes the $L_i$-th degree of $x_s$, and $\hat{r}_{ts} = \frac{\vec{r}_{ts}}{| \vec{r}_{ts} |}$.
The spherical harmonic projection of relative positions $Y^{(L_f)}(\hat{r}_{ts})$ becomes sparse if we rotate $\hat{r}_{ts}$ with a rotation matrix $R_{ts}$ to align with the direction of $L = 0$ and $m = 0$, which corresponds to the z axis traditionally but the y axis in the conventions of $\texttt{e3nn}$~\citep{e3nn}.
Concretely, given $R_{ts} \hat{r}_{ts}$ aligned with the y axis, $Y^{(L_f)}_{m_f}( R_{ts} \hat{r}_{ts}) \neq 0$ only for $m_f = 0$.
Without loss of equivariance, we re-scale $Y^{(L_f)}_{0}( R_{ts} \vec{r}_{ts})$ to be one.
Besides, we denote $D^{(L_i)}(R_{ts}) = D^{(L_i)}$ and $D^{(L_o)}(R_{ts}) = D^{(L_o)}$ as Wigner-D matrices of degrees $L_i$ and $L_o$ based on rotation matrix $R_{ts}$, respectively, and we define $D^{(L_i)}x_s^{(L_i)} = \tilde{x}_s^{(L_i)}$.
Therefore, by rotating $x_s^{(L_i)}$ and $Y^{(L_f)}$ based on $\hat{r}_{ts}$, we can simplify Eq.~\ref{appendix:eq:so3_convolution} as follows:
\begin{equation}
\begin{aligned}
%m^{(L_o)}_{ts} &= \sum_{L_{i}, L_{f}} w_{L_i, L_f, L_o} \left(x_s^{(L_i)} \otimes^{L_o}_{L_i, L_f} Y^{(L_f)}(\hat{r}_{ts}) \right) \\
m^{(L_o)}_{ts} &= \left( D^{(L_o)}(R_{ts}) \right)^{-1} \sum_{L_{i}, L_{f}} w_{L_i, L_f, L_o} \left(D^{(L_i)}(R_{ts}) x_s^{(L_i)} \otimes^{L_o}_{L_i, L_f} Y^{(L_f)}(R_{ts} \hat{r}_{ts}) \right) \\
&= \left( D^{(L_o)} \right)^{-1} \sum_{L_{i}, L_{f}} w_{L_i, L_f, L_o} \bigoplus_{m_o} \left( \sum_{m_i, m_f} \left(D^{(L_i)} x_s^{(L_i)} \right)_{m_i} C^{(L_o, m_o)}_{(L_i, m_i), (L_f, m_f)} \left(Y^{(L_f)}(R_{ts} \hat{r_{ts}})\right)_{m_f} \right) \\
&= \left( D^{(L_o)} \right)^{-1} \sum_{L_{i}, L_{f}} w_{L_i, L_f, L_o} \bigoplus_{m_o} \left( \sum_{m_i} \left(D^{(L_i)} x_s^{(L_i)} \right)_{m_i} C^{(L_o, m_o)}_{(L_i, m_i), (L_f, 0)} \right) \\
&= \left( D^{(L_o)} \right)^{-1} \sum_{L_{i}, L_{f}} w_{L_i, L_f, L_o} \bigoplus_{m_o} \left( \sum_{m_i} \left( \tilde{x}_s^{(L_i)} \right)_{m_i} C^{(L_o, m_o)}_{(L_i, m_i), (L_f, 0)} \right) \\
\label{appendix:eq:rotated_so3_convolution}
\end{aligned}
\end{equation}
%where $D^{(L_i)}(R_{ts}) = D^{(L_i)}$ and $D^{(L_o)}(R_{ts}) = D^{(L_o)}$ denote Wigner-D matrices of degrees $L_i$ and $L_o$ based on rotation matrix $R_{ts}$, respectively, $\bigoplus$ denotes concatenation, and $D^{(L_i)}x_s^{(L_i)} = \tilde{x}_s^{(L_i)}$.
where $\bigoplus$ denotes concatenation.
Additionally, given $m_f = 0$, Clebsch-Gordan coefficients $C^{(L_o, m_o)}_{(L_i, m_i), (L_f, 0)}$ are sparse and are non-zero only when $m_i = \pm m_o$, which further simplifies Eq.~\ref{appendix:eq:rotated_so3_convolution}:
\begin{equation}
\begin{aligned}
m^{(L_o)}_{ts} &= \left( D^{(L_o)} \right)^{-1} \sum_{L_{i}, L_{f}} w_{L_i, L_f, L_o} \bigoplus_{m_o} \left( \left( \tilde{x}_s^{(L_i)} \right)_{m_o} C^{(L_o, m_o)}_{(L_i, m_o), (L_f, 0)} + \left(\tilde{x}_s^{(L_i)} \right)_{-m_o} C^{(L_o, m_o)}_{(L_i, -m_o), (L_f, 0)} \right) \\
%h^{(L_3)}_{m_3} = (f^{(L_1)} \otimes g^{(L_2)})_{m_3} = \sum_{m_1 = -L_1}^{L_1} \sum_{m_2 = -L_2}^{L_2} C^{(L_3, m_3)}_{(L_1, m_1)(L_2, m_2)} f^{(L_1)}_{m_1} g^{(L_2)}_{m_2}
\end{aligned}
\label{appendix:eq:pm_m_so3_convolution}
\end{equation}
By re-ordering the summations and concatenation in Eq.~\ref{appendix:eq:pm_m_so3_convolution}, we have:
\begin{equation}
\left( D^{(L_o)} \right)^{-1} \sum_{L_{i}} \bigoplus_{m_o} \left( \left( \tilde{x}_s^{(L_i)} \right)_{m_o} \sum_{L_f} \left(w_{L_i, L_f, L_o} C^{(L_o, m_o)}_{(L_i, m_o), (L_f, 0)} \right) + \left( \tilde{x}_s^{(L_i)} \right)_{-m_o} \sum_{L_f} \left( w_{L_i, L_f, L_o} C^{(L_o, m_o)}_{(L_i, -m_o), (L_f, 0)} \right) \right)\\
%h^{(L_3)}_{m_3} = (f^{(L_1)} \otimes g^{(L_2)})_{m_3} = \sum_{m_1 = -L_1}^{L_1} \sum_{m_2 = -L_2}^{L_2} C^{(L_3, m_3)}_{(L_1, m_1)(L_2, m_2)} f^{(L_1)}_{m_1} g^{(L_2)}_{m_2}
\label{appendix:eq:reorder_pm_m_so3_convolution}
\end{equation}
Instead of using learnable parameters for $w_{L_i, L_f, L_o}$, eSCN proposes to parametrize $\tilde{w}^{(L_i, L_o)}_{m_o}$ and $\tilde{w}^{(L_i, L_o)}_{-m_o}$ as below:
\begin{equation}
\begin{aligned}
\tilde{w}^{(L_i, L_o)}_{m_o} &= \sum_{L_f} w_{L_i, L_f, L_o} C^{(L_o, m_o)}_{(L_i, m_o), (L_f, 0)} = \sum_{L_f} w_{L_i, L_f, L_o} C^{(L_o, -m_o)}_{(L_i, -m_o), (L_f, 0)} \quad \text{for } m >= 0 \\
\tilde{w}^{(L_i, L_o)}_{-m_o} &= \sum_{L_f} w_{L_i, L_f, L_o} C^{(L_o, -m_o)}_{(L_i, m_o), (L_f, 0)} = - \sum_{L_f} w_{L_i, L_f, L_o} C^{(L_o, m_o)}_{(L_i, -m_o), (L_f, 0)} \quad \text{for } m > 0
\end{aligned}
\label{appendix:eq:reparametrization}
\end{equation}
The parametrization of $\tilde{w}^{(L_i, L_o)}_{m_o}$ and $\tilde{w}^{(L_i, L_o)}_{-m_o}$ enables removing the summation over $L_f$ and further simplifies the computation.
By combining Eq.~\ref{appendix:eq:reorder_pm_m_so3_convolution} and Eq.~\ref{appendix:eq:reparametrization}, we have:
\begin{equation}
\begin{aligned}
m^{(L_o)}_{ts} &= \left( D^{(L_o)} \right)^{-1} \sum_{L_i} \bigoplus_{m_o} \left( y_{ts}^{(L_i, L_o)} \right) _{m_o} \\
\left( y_{ts}^{(L_i, L_o)} \right) _{m_o} &= \tilde{w}^{(L_i, L_o)}_{m_o} \left( \tilde{x}_s^{(L_i)} \right)_{m_o} - \tilde{w}^{(L_i, L_o)}_{-m_o} \left( \tilde{x}_s^{(L_i)} \right)_{-m_o} \quad \text{for } m_o > 0\\
\left( y_{ts}^{(L_i, L_o)} \right) _{-m_o} &= \tilde{w}^{(L_i, L_o)}_{-m_o} \left( \tilde{x}_s^{(L_i)} \right)_{m_o} + \tilde{w}^{(L_i, L_o)}_{m_o} \left( \tilde{x}_s^{(L_i)} \right)_{-m_o} \quad \text{for } m_o > 0 \\
\left( y_{ts}^{(L_i, L_o)} \right) _{m_o} &= \tilde{w}^{(L_i, L_o)}_{m_o} \left(\tilde{x}_s^{(L_i)} \right)_{m_o} \quad \text{for } m_o = 0 \\
\end{aligned}
\label{appendix:eq:so2_linear}
\end{equation}
The formulation of $y_{ts}^{(L_i, L_o)}$ coincides with performing $SO(2)$ linear operations~\citep{harmonis_network, escn}.
Additionally, eSCN convolutions can further simplify the computation by considering only a subset of $m_o$ components in Eq.~\ref{appendix:eq:so2_linear}, i.e., $| m_o | \leq M_{max}$.

In summary, efficient $SO(3)$ convolutions can be achieved by first rotating irreps features $x_s^{(L_i)}$ based on relative position vectors $\vec{r}_{ts}$ and then performing $SO(2)$ linear operations on rotated features.
The key idea is that rotation simplifies the computation as in Eq.~\ref{appendix:eq:rotated_so3_convolution},~\ref{appendix:eq:pm_m_so3_convolution},~\ref{appendix:eq:reparametrization}, and~\ref{appendix:eq:so2_linear}.
Please refer to their work~\citep{escn} for more details.
We note that eSCN convolutions consider only simplifying the case of taking tensor products between input irreps features and spherical harmonic projections of relative position vectors.
eSCN convolutions do not simplify general cases such as taking tensor products between input irreps features and themselves~\citep{mace} since the relative position vectors used to rotate irreps features are not clearly defined.

\subsection{$S^2$ Activation}
\label{appendix:subsec:s2_activation}

$S^2$ activation was first proposed in Spherical CNNs~\citep{spherical_cnn}.
Our implementation of $S^2$ activation is the same as that in \texttt{e3nn}~\citep{e3nn}, SCN~\citep{scn} and eSCN~\citep{escn}. 
Basically, we uniformly sample a fixed set of points on a unit sphere along the dimensions of longitude, parametrized by $\alpha \in [0, 2 \pi)$, and latitude, parametrized by $\beta \in [0, \pi)$. 
We set the resolutions $R$ of $\alpha$ and $\beta$ to be $18$ when $L_{max} = 6$, meaning that we will have $324$ $(=18 \times 18)$ points. 
Once the points are sampled, they are kept the same during training and inference, and therefore there is no randomness. 
For each point on the unit sphere, we compute the spherical harmonics projection of degrees up to $L_{max}$. 
We consider an equivariant feature of $C$ channels and each channel contains vectors of all degrees from $0$ to $L_{max}$.
When performing $S^2$ activation, for each channel and for each sampled point, we first compute the inner product between the vectors of all degrees contained in one channel of the equivariant feature and the spherical harmonics projections of a sampled point. 
This results in $R \times R \times C$ values, where the first two dimensions, $R \times R$, correspond to grid resolutions and the last dimension corresponds to channels. 
They can be viewed as 2D grid feature maps and treated as scalars, and we can apply any standard or typical activation functions like SiLU or use standard linear layers performing feature aggregation along the channel dimension. 
After applying these functions, we project back to vectors of all degrees by multiplying those values with their corresponding spherical harmonics projections of sampled points. 
The process is the same as performing a Fourier transform, applying some functions and then performing an inverse Fourier transform.

Moreover, since the inner products between one channel of vectors of all degrees and the spherical harmonics projections of sampled points sum over all degrees, the conversion to 2D grid feature maps implicitly considers the information of all degrees. 
Therefore, $S^2$ activation, which converts equivariant features into 2D grid feature maps, uses the information of all degrees to determine the non-linearity. 
In contrast, gate activation only uses vectors of degree $0$ to determine the non-linearity of vectors of higher degrees. 
For tasks such as force predictions, where the information of degrees is critical, $S^2$ activation can be better than gate activation since $S^2$ activation uses all degrees to determine non-linearity.    

Although there is sampling on a sphere, the works~\citep{spherical_cnn, escn} mention that as long as the number of samples, or resolution $R$, is high enough, the equivariance error can be close to zero. 
Furthermore, eSCN~\citep{escn} empirically computes such errors in Figure 9 in their latest manuscript and shows that the errors of using $L_{max} = 6$ and $R = 18$ are close to $0.2\%$, which is similar to the equivariance errors of tensor products in \texttt{e3nn}~\citep{e3nn}. 
We note that the equivariance errors in \texttt{e3nn} are due to numerical precision.

%% file: content/4_related_works.tex
%\vspace{-3mm}
\section{Related Works}
\label{sec:related_work}
%\vspace{-3mm}

%\paragraph{\textit{SE(3)/E(3)}-Equivariant GNNs.}
\subsection{\textit{SE(3)/E(3)}-Equivariant GNNs}
\label{subsec:equivariant_networks}

Equivariant neural networks~\citep{tfn, kondor2018clebsch, 3dsteerable, se3_transformer, l1net, geometric_prediction, nequip, gvp, painn, egnn, se3_wavefunction, segnn, torchmd_net, eqgat, allergo, mace, equiformer, escn} use equivariant irreps features built from vector spaces of irreducible representations (irreps) to achieve equivariance to 3D rotation~\citep{tfn, 3dsteerable, kondor2018clebsch}.
They operate on irreps features with equivariant operations like tensor products.
Previous works differ in equivariant operations used in their networks and how they combine those operations.
TFN~\citep{tfn} and NequIP~\citep{nequip} use equivariant graph convolution with linear messages built from tensor products, with the latter utilizing extra equivariant gate activation~\citep{3dsteerable}.
SEGNN~\citep{segnn} introduces non-linearity to  messages passing~\citep{neural_message_passing_quantum_chemistry, gns} with equivariant gate activation, and the non-linear messages improve upon linear messages.
SE(3)-Transformer~\citep{se3_transformer} adopts equivariant dot product attention~\citep{transformer} with linear messages.
Equiformer~\citep{equiformer} improves upon previously mentioned equivariant GNNs by combining MLP attention and non-linear messages. Equiformer additionally introduces equivariant layer normalization and regularizations like dropout~\citep{dropout} and stochastic depth~\citep{drop_path}.
However, the networks mentioned above rely on compute-intensive $SO(3)$ tensor products to mix the information of vectors of different degrees during message passing, and therefore they are limited to small values for maximum degrees $L_{max}$ of equivariant representations.
SCN~\citep{scn} proposes rotating irreps features based on relative position vectors and identifies a subset of spherical harmonics coefficients, on which they can apply unconstrained functions.
They further propose relaxing the requirement for strict equivariance and apply typical functions to rotated features during message passing, which trades strict equivariance for computational efficiency and enables using higher values of $L_{max}$.
eSCN~\citep{escn} further improves upon SCN by replacing typical functions with $SO(2)$ linear layers for rotated features and imposing strict equivariance during message passing.
%While SCN and eSCN enable higher $L_{max}$, their network design still follows that of SEGNN and therefore have room for improvement.
%\revision{While SCN and eSCN enable higher $L_{max}$, their network design still follows that of SEGNN, which is less performant than Equiformer.}
However, except using more efficient operations for higher $L_{max}$, SCN and eSCN mainly adopt the same network design as SEGNN, which is less performant than Equiformer.
In this work, we propose EquiformerV2, which includes all the benefits of the above networks by incorporating eSCN convolutions into Equiformer and adopts three additional architectural improvements.

%\vspace{-3mm}
%\paragraph{Invariant GNNs.}
\subsection{Invariant GNNs}
\label{subsec:invariant_networks}

Prior works~\citep{schnet, cgcnn, physnet, dimenet, dimenet_pp, orbnet, spherenet, spinconv, gemnet, gemnet_xl, gasteiger_gemnet_oc_2022} extract invariant information from 3D atomistic graphs and operate on the resulting graphs augmented with invariant features.
Their differences lie in leveraging different geometric features such as distances, bond angles (3 atom features) or dihedral angles (4 atom features).
SchNet~\citep{schnet} models interaction between atoms with only relative distances.
DimeNet series~\citep{dimenet, dimenet_pp} use triplet representations of atoms to incorporate bond angles.
%SphereNet~\cite{spherenet} and GemNet~\cite{gemnet} further extend to consider dihedral angles, which has been shown by GemNet~\cite{gemnet} to be universal approximators.
SphereNet~\citep{spherenet} and GemNet~\citep{gemnet, gasteiger_gemnet_oc_2022} further include dihedral angles by considering quadruplet representations.
However, the memory complexity of triplet and quadruplet representations of atoms do not scale well with the number of atoms, and this requires additional modifications like interaction hierarchy used by GemNet-OC~\citep{gasteiger_gemnet_oc_2022} for large datasets like OC20~\citep{oc20}.
Additionally, for the task of predicting DFT calculations of energies and forces on the large-scale OC20 dataset, invariant GNNs have been surpassed by equivariant GNNs recently.

%% file: content/8_2_details_of_architecture.tex
\section{Details of Architecture}
\label{appendix:sec:details_of_architecture}

In this section, we define architectural hyper-parameters like maximum degrees and numbers of channels  in certain layers in EquiformerV2, which are used to specify the detailed architectures in Sec.~\ref{appendix:subsec:training_details}, Sec.~\ref{appendix:subsec:additional_comparison_on_oc20_is2re}, Sec.~\ref{appendix:subsec:oc22_training_details} and Sec.~\ref{appendix:subsec:qm9_training_details}.
Besides, we note that eSCN~\citep{escn} and this work mainly consider $SE(3)$-equivariance.

We denote embedding dimensions as $d_{embed}$, which defines the dimensions of most irreps features.
Specifically, the output irreps features of all modules except the output head in Figure~\ref{fig:equiformer_v2}a have dimension $d_{embed}$.
%Specifically, the output irreps features of all modules except the output head in Figure {\color{crimson}1}a have dimension $d_{embed}$.
For separable $S^2$ activation as illustrated in Figure~\ref{fig:activation}c, we denote the resolution of point samples on a sphere as $R$, which can depend on maximum degree $L_{max}$, and denote the unconstrained functions after projecting to point samples as $F$.
%For separable $S^2$ activation as illustrated in Figure {\color{crimson}2}c, we denote the resolution of point samples on a sphere as $R$, which can depend on maximum degree $L_{max}$, and denote the unconstrained functions after projecting to point samples as $F$.

For equivariant graph attention in Figure~\ref{fig:equiformer_v2}b, the input irreps features $x_i$ and $x_j$ have dimension $d_{embed}$.
%For equivariant graph attention in Figure {\color{crimson}1}b, the input irreps features $x_i$ and $x_j$ have dimension $d_{embed}$.
The dimension of the irreps feature $f_{ij}^{(L)}$ is denoted as $d_{attn\_hidden}$.
Equivariant graph attention can have $h$ parallel attention functions.
For each attention function, we denote the dimension of the scalar feature $f_{ij}^{(0)}$ as $d_{attn\_alpha}$ and denote the dimension of the value vector, which is in the form of irreps features, as $d_{attn\_value}$.
For the separable $S^2$ activation used in equivariant graph attention, the resolution of point samples is $R$, and we use a single SiLU activation for $F$.
We share the layer normalization in attention re-normalization across all $h$ attention functions but have different $h$ linear layers after that.
The last linear layer projects the dimension back to $d_{embed}$.
The two intermediate $SO(2)$ linear layers operate with maximum degree $L_{max}$ and maximum order $M_{max}$.

For feed forward networks (FFNs) in Figure~\ref{fig:equiformer_v2}d, we denote the dimension of the output irreps features of the first linear layer as $d_{ffn}$.
%For feed forward networks (FFNs) in Figure {\color{crimson}1}d, we denote the dimension of the output irreps features of the first linear layer as $d_{ffn}$.
%For the separable $S^2$ activation in Figure~\ref{fig:equiformer_v2}d, the resolution of point samples is $R$, and $F$ consists of a two-layer MLP, with each linear layer followed by SiLU, and a final linear layer.
For the separable $S^2$ activation used in FFNs, the resolution of point samples is $R$, and $F$ consists of a two-layer MLP, with each linear layer followed by SiLU, and a final linear layer.
The linear layers have the same number of channels as $d_{ffn}$.
%The FFN in the last Transformer block has output dimension $d_{feature}$, and we set $d_{ffn}$ of the last FFN, which is followed by output head, to be $d_{feature}$ as well.
%Thus, two hyper-parameters $d_{ffn}$ and $d_{feature}$ are used to specify architectures of FFNs and the output dimension after Transformer blocks.

For radial functions, we denote the dimension of hidden scalar features as $d_{edge}$.
For experiments on OC20, same as eSCN~\citep{escn}, we use Gaussian radial basis to represent relative distances and additionally embed the atomic numbers at source nodes and target nodes with two scalar features of dimension $d_{edge}$.
The radial basis and the two embeddings of atomic numbers are fed to the radial function to generate edge distance embeddings.

%Irreps features contain channels of vectors with degrees up to $L_{max}$.
%We denote $C_L$ type-$L$ vectors as $(C_L, L)$ and $C_{(L, p)}$ type-$(L, p)$ vectors as $(C_{(L, p)}, L, p)$ and use brackets to represent concatenations of vectors.
%For example, the dimension of irreps features containing $256$ type-$0$ vectors and $128$ type-$1$ vectors can be represented as $[(256, 0), (128, 1)]$.
%We can extend this to include inversion by introducing parity $p$ and using $(C_{(L, p)}, L, p)$.

The maximum degree of irreps features is denoted as $L_{max}$.
All irreps features have degrees from $0$ to $L_{max}$ and have $C$ channels for each degree.
We denote the dimension as $(L_{max}, C)$.
For example, irreps feature $x_{irreps}$ of dimension $(6, 128)$ has maximum degree $6$ and $128$ channels for each degree.
The dimension of scalar feature $x_{scalar}$ can be expressed as $(0, C_{scalar})$.

Following Equiformer~\citep{equiformer}, we apply dropout~\citep{dropout} to attention weights and stochastic depth~\citep{drop_path} to outputs of equivariant graph attention and feed forward networks.
However, we do not apply dropout or stochastic depth to the output head.

%% file: content/8_3_details_of_experiments_on_oc20.tex
\section{Details of Experiments on OC20}
\label{appendix:sec:details_of_experiments_on_oc20}

\subsection{Detailed Description of OC20 Dataset}
\label{appendix:subsec:detailed_description_of_oc20_dataset}

%Our experiments focus on the large and diverse OC20 dataset~\citep{oc20} (Creative Commons Attribution 4.0 License),
%which consists of ${\sim}1.2M$ DFT relaxations for training and evaluation,
%which consists of $1.2$M DFT relaxations for training and evaluation,
%computed with the revised Perdew-Burke-Ernzerhof (RPBE) functional~\citep{rpbe}.
%Each structure in OC20 has an adsorbate molecule placed on a catalyst surface,
%and the core task is Structure-to-Energy-Forces (or S2EF),~\ie
%to predict the energy of the structure and per-atom forces.
The large and diverse OC20 dataset~\citep{oc20} (Creative Commons Attribution 4.0 License) consists of $1.2$M DFT relaxations for training and evaluation, computed with the revised Perdew-Burke-Ernzerhof (RPBE) functional~\citep{rpbe}.
%which consists of ${\sim}1.2M$ DFT relaxations for training and evaluation,
%which consists of $1.2$M DFT relaxations for training and evaluation,
%computed with the revised Perdew-Burke-Ernzerhof (RPBE) functional~\citep{rpbe}.
Each structure in OC20 has an adsorbate molecule placed on a catalyst surface, and the core task is Structure to Energy Forces (S2EF), which is to predict the energy of the structure and per-atom forces.
Models trained for the S2EF task are evaluated on energy and force mean
absolute error (MAE).
These models can in turn be used for performing structure relaxations by using
the model's force predictions to iteratively update the atomic positions until a
relaxed structure corresponding to a local energy minimum is found.
These relaxed structure and energy predictions are evaluated on the
Initial Structure to Relaxed Structure (IS2RS) and Initial Structure to Relaxed
Energy (IS2RE) tasks.
%
%While the largest S2EF split of OC20 consists of ${\sim}$134M training
%structures spanning 56 elements, we first report ablations of EquiformerV2
%variants trained on the S2EF-2M subset in \tableref{tab:ablations_all}
%(similar to Gasteiger~\etal~\cite{gasteiger_gemnet_oc_2022})
%before scaling up to the larger S2EF-All and S2EF-All + MD splits
%(\tableref{tab:oc20_all_results}).
%
The ``All'' split of OC20 contains $134$M training structures spanning $56$ elements, the ``MD'' split consists of $38$M structures, and the ``2M'' split has $2$M structures.
For validation and test splits, there are four sub-splits containing in-distribution adsorbates and catalysts (ID), out-of-distribution adsorbates (OOD Ads), out-of-distribution catalysts (OOD Cat), and out-of-distribution adsorbates and catalysts (OOD Both).

\begin{comment}
\begin{enumerate}
    \item \todo{Training and architectural hyper-parameters.}
    \begin{enumerate}
        \item \todo{S2EF-2M.}
        \item \todo{S2EF-All.}
    \end{enumerate}

    \item \todo{Training time/inference time and number of parameters.}
    \begin{enumerate}
        \item \todo{S2EF-2M.}
        \begin{enumerate}
            \item \todo{Architectural improvements.}
            \item \todo{Scaling of Parameters.}
            \item \todo{Speed-Accuracy Trade-offs.}
        \end{enumerate}
        \item \todo{S2EF-All.}
    \end{enumerate}

    \item \todo{Details of running relaxations.}

    \item \todo{Details of running AdsorbML.}
        \begin{enumerate}
            \item \todo{Runtime (in terms of GPU-hours and CPU-hours) for each $k$ and each model if possible.}
        \end{enumerate}
\end{enumerate}
\end{comment}

\begin{table}[t]
\centering
\scalebox{0.65}{
\begin{tabular}{llll}
%\cline{7-11}
\toprule[1.2pt]
& \multicolumn{1}{c}{Base model setting on} & \multicolumn{1}{c}{EquiformerV2 ($31$M) on} & \multicolumn{1}{c}{EquiformerV2 ($153$M) on} \\
Hyper-parameters & \multicolumn{1}{c}{S2EF-2M} & \multicolumn{1}{c}{S2EF-All+MD} & \multicolumn{1}{c}{S2EF-All/S2EF-All+MD} \\
\midrule[1.2pt]
Optimizer & AdamW & AdamW & AdamW \\
%Learning rate scheduling & Cosine learning rate with linear warmup \\
Learning rate scheduling & Cosine learning rate with & Cosine learning rate with & Cosine learning rate with \\
& linear warmup & linear warmup & linear warmup \\
Warmup epochs & $0.1$ & $0.01$ & $0.01$ \\
Maximum learning rate & $2 \times 10 ^{-4}$ & $4 \times 10 ^{-4}$ & $4 \times 10 ^{-4}$ \\
Batch size & $64$ & $512$ & $256$ for S2EF-All, \\
& & & $512$ for S2EF-All+MD \\
Number of epochs & $12$ & $3$ & $1$ \\
Weight decay & $1 \times 10 ^{-3}$ & $1 \times 10 ^{-3}$ & $1 \times 10 ^{-3}$\\
Dropout rate & $0.1$ & $0.1$ & $0.1$ \\
Stochastic depth & $0.05$ & $0.1$ & $0.1$ \\
Energy coefficient $\lambda_{E}$ & $2$ & $4$ & $2$ for S2EF-All, \\
& & & $2, 4$ for S2EF-All+MD \\
Force coefficient $\lambda_{F}$ & $100$ & $100$ & $100$ \\
Gradient clipping norm threshold & $100$ & $100$ & $100$ \\
Model EMA decay & $0.999$ & $0.999$ & $0.999$ \\
%\midrule[0.6pt]
Cutoff radius ($\angstrom$) & $12$ & $12$ & $12$ \\
Maximum number of neighbors & $20$ & $20$ & $20$ \\
Number of radial bases & $600$ & $600$ & $600$ \\
Dimension of hidden scalar features in radial functions $d_{edge}$ & $(0, 128)$ & $(0, 128)$ & $(0, 128)$ \\
%Hidden size of radial function & $64$ \\
%Number of hidden layers in radial function & $2$ \\

%\midrule[0.6pt]

%\multicolumn{2}{c}{Equiformer} \\
%\\
Maximum degree $L_{max}$ & $6$ & $4$ & $6$ \\
Maximum order $M_{max}$ & $2$ & $2$ & $3$\\
Number of Transformer blocks & $12$ & $8$ & $20$ \\
Embedding dimension $d_{embed}$ & $(6, 128)$ & $(4, 128)$ & $(6, 128)$ \\
%Spherical harmonics embedding dimension $d_{sh}$ & $[(1, 0), (1, 1)]$ \\
$f_{ij}^{(L)}$ dimension $d_{attn\_hidden}$ & $(6, 64)$ & $(4, 64)$ & $(6, 64)$ \\
Number of attention heads $h$ & $8$ & $8$ & $8$ \\
$f_{ij}^{(0)}$ dimension $d_{attn\_alpha}$ & $(0, 64)$ & $(0, 64)$ & $(0, 64)$ \\
Value dimension $d_{attn\_value}$ & $(6, 16)$ & $(4, 16)$ & $(6, 16)$\\
%Attention head dimension $d_{head}$ & $[(32, 0), (16, 1)]$\\
Hidden dimension in feed forward networks $d_{ffn}$ & $(6, 128)$ & $(4, 128)$ & $(6, 128)$ \\
Resolution of point samples $R$ & $18$ & $18$ & $18$ \\

%\midrule[0.6pt]

%\multicolumn{2}{c}{$E(3)$-Equiformer} \\
%\\
%Number of Transformer blocks & $6$ \\
%Embedding dimension $d_{embed}$ & $[(256, 0, e), (64, 0, o), (64, 1, e), (64, 1, o)]$ \\
%Spherical harmonics embedding dimension $d_{sh}$ & $[(1, 0, e), (1, 1, o)]$ \\
%Number of attention heads $h$ & $8$ \\
%Attention head dimension $d_{head}$ & $[(32, 0, e), (8, 0, o), (8, 1, e), (8, 1, o)]$ \\
%Hidden dimension in feed forward networks $d_{ffn}$ & $[(768, 0, e), (192, 0, o), (192, 1, e), (192, 1, o)]$ \\
%Output feature dimension $d_{feature}$ & $[(512, 0, e)]$\\

\bottomrule[1.2pt]
\end{tabular}
}
\vspace{2mm}
\caption{Hyper-parameters for the base model setting on OC20 S2EF-2M dataset and the main results on OC20 S2EF-All and S2EF-All+MD datasets.
}
\label{appendix:tab:oc20_s2ef_hyperparameters}
\end{table}

\begin{table}[t]
\begin{adjustwidth}{-5cm}{-5cm}
\centering
\scalebox{0.55}{
\begin{tabular}{lccccccrrr}
\toprule[1.2pt]
& Attention & & & & & Number of & \multicolumn{1}{c}{Training time} & \multicolumn{1}{c}{Inference speed} & \multicolumn{1}{c}{Number of} \\
Training set & re-normalization & Activation & Normalization & $L_{max}$ & $M_{max}$ & Transformer blocks & \multicolumn{1}{c}{(GPU-hours)} & \multicolumn{1}{c}{(Samples / GPU sec.)} & \multicolumn{1}{c}{parameters} \\
%\shline
\midrule[1.2pt]
\multirow{10}{*}{S2EF-2M} & \xmark & Gate & LN & $6$ & $2$ & $12$ & $965$ & $19.06$ & $91.06$M \\
& \cmark & Gate & LN & 6 & 2 & 12 & $998$ & $19.07$ & $91.06$M \\
& \cmark & $S^{2}$ & LN & $6$ & $2$ & $12$ & $1476$ & $12.80$ & $81.46$M \\
& \cmark & Sep. $S^{2}$ & LN & $6$ & $2$ & $12$ & $1505$ & $12.51$ & $83.16$M \\
& \cmark & Sep. $S^{2}$ & SLN & $6$ & $2$ & $12$ & $1412$ & $13.22$ & $83.16$M \\

& \cmark & Sep. $S^{2}$ & SLN & $4$ & $2$ & $12$ & $965$ & $19.86$ & $44.83$M \\
& \cmark & Sep. $S^{2}$ & SLN & $8$ & $2$ & $12$ & $2709$ & $7.86$ & $134.28$M \\
& \cmark & Sep. $S^{2}$ & SLN & $6$ & $3$ & $12$ & $1623$ & $11.92$ & $95.11$M \\
& \cmark & Sep. $S^{2}$ & SLN & $6$ & $4$ & $12$ & $2706$ & $7.98$ & $102.14$M \\
& \cmark & Sep. $S^{2}$ & SLN & $6$ & $6$ & $12$ & $3052$ & $7.13$ & $106.63$M \\

\midrule
S2EF-All & \cmark & Sep. $S^{2}$ & SLN & $6$ & $3$ & $20$ & $20499$ & $6.08$ & $153.60$M \\

\midrule
S2EF-All+MD ($\lambda_{E} = 2$)& \cmark & Sep. $S^{2}$ & SLN & $6$ & $3$ & $20$ & $32834$ & $6.08$ & $153.60$M \\

\midrule
S2EF-All+MD ($\lambda_{E} = 4$)& \cmark & Sep. $S^{2}$ & SLN & $4$ & $2$ & $8$ & $16931$ & $29.21$ & $31.06$M \\
S2EF-All+MD ($\lambda_{E} = 4$)& \cmark & Sep. $S^{2}$ & SLN & $6$ & $3$ & $20$ & $37692$ & $6.08$ & $153.60$M \\
\bottomrule[1.2pt]
\end{tabular}
}
\vspace{2mm}
\end{adjustwidth}
\caption{Training time, inference speed and numbers of parameters of different models trained on OC20 S2EF-2M, S2EF-All and S2EF-All+MD datasets.
All numbers are measured on V100 GPUs with 32GB.
}
\label{appendix:tab:training_time_inference_speed_and_numbers_of_parameters}
\end{table}

\subsection{Training Details}
\label{appendix:subsec:training_details}

\paragraph{Hyper-Parameters.}
We summarize the hyper-parameters for the base model setting on OC20 S2EF-2M dataset and the main results on OC20 S2EF-All and S2EF-All+MD datasets in Table~\ref{appendix:tab:oc20_s2ef_hyperparameters}.
%For the ablation studies on OC20 S2EF-2M dataset, when trained for $20$ or $30$ epochs as in Table {\color{crimson}1b}, we increase the learning rate from $2 \times 10^{-4}$ to $4 \times 10^{-4}$.
For the ablation studies on OC20 S2EF-2M dataset, when trained for $20$ or $30$ epochs as in~\tableref{tab:training_epochs}, we increase the learning rate from $2 \times 10^{-4}$ to $4 \times 10^{-4}$.
%When using $L_{max} = 8$ as in Table {\color{crimson}1c}, we increase the resolution of point samples $R$ from $18$ to $20$.
When using $L_{max} = 8$ as in~\tableref{tab:l_degrees}, we increase the resolution of point samples $R$ from $18$ to $20$.
%We vary $L_{max}$ and the widths for speed-accuracy trade-offs in Figure {\color{crimson}4}.
We vary $L_{max}$ and the widths for speed-accuracy trade-offs in Figure~\ref{fig:combined_train_test_throughput}.
Specifically, we first decrease $L_{max}$ from $6$ to $4$.
Then, we multiply $h$ and the number of channels of $(d_{embed}, d_{attn\_hidden}, d_{ffn})$ by $0.75$ and $0.5$.
We train all models for $30$ epochs.
The same strategy to scale down eSCN models is adopted for fair comparisons.

\paragraph{Training Time, Inference Speed and Numbers of Parameters.}
Table~\ref{appendix:tab:training_time_inference_speed_and_numbers_of_parameters} summarizes the training time, inference speed and numbers of parameters of models in Tables~\ref{tab:ablations} (Index 1, 2, 3, 4, 5),~\ref{tab:l_degrees},~\ref{tab:m_orders} and~\ref{tab:oc20_all_results}.
%Table~\ref{appendix:tab:training_time_inference_speed_and_numbers_of_parameters} summarizes the training time, inference speed and numbers of parameters of models in Tables {\color{crimson}1a} (row 1, 2, 3, 4, 5), {\color{crimson}1c}, {\color{crimson}1d} and {\color{crimson}2}.
%For S2EF-2M dataset, models are trained for $12$ epochs.
V100 GPUs with 32GB are used to train all models.
We use $16$ GPUs to train each individual model on S2EF-2M dataset, $64$ GPUs for S2EF-All, $64$ GPUs for EquiformerV2 ($31$M) on S2EF-All+MD, and $128$ GPUs for EquiformerV2 ($153$M) on S2EF-All+MD.

\subsection{Details of Running Relaxations}
\label{appendix:subsec:details_of_running_relaxations}
%A structural relaxation is a local optimization where atom positions are iteratively updated based on the forces to minimize the energy.
A structural relaxation is a local optimization where atom positions are iteratively updated based on forces to minimize the energy of the structure.
%We performed all of our ML relaxations using the LBFGS optimizer (quasi-Newton) implemented in the Open Catalyst Github repository, full optimizer settings are listed below.
We perform ML relaxations using the LBFGS optimizer (quasi-Newton) implemented in the Open Catalyst Github repository~\citep{oc20}.
%All OC20 optimizations were allowed to run for up to 200 steps or until $F_{max} \leq 0.02$ eV/\angstrom, whereas AdsorbML relaxations were allowed to run for 300 steps or until $F_{max} \leq 0.02$ eV/\angstrom. These settings were chosen to be consistent with prior work.
The structural relaxations for OC20 IS2RE and IS2RS tasks are allowed to run for $200$ steps or until the maximum predicted force per atom $F_{max} \leq 0.02$ eV/\angstrom, and the relaxations for AdsorbML are allowed to run for $300$ steps or until $F_{max} \leq 0.02$ eV/\angstrom.
These settings are chosen to be consistent with prior works.
We run relaxations on V100 GPUs with 32GB.
The computational cost of running relaxations with EquiformerV2 ($153$M) for OC20 IS2RE and IS2RS tasks is $1011$ GPU-hours, and that of running ML relaxations for AdsorbML is $1075$ GPU-hours.
%\ad{The time for running relaxations with EquiformerV2 ($31$M) is 240 GPU-hours for IS2RE and IS2RS, and 298 GPU-hours for AdsorbML.}
The time for running relaxations with EquiformerV2 ($31$M) is 240 GPU-hours for OC20 IS2RE and IS2RS and 298 GPU-hours for AdsorbML.

\subsection{Details of AdsorbML}
\label{appendix:subsec:details_of_adsorbml}

%We ran the AdsorbML algorithm on the OC20-Dense dataset in accordance with the procedure laid out in the paper, which is summarized here:
We run the AdsorbML algorithm on the OC20-Dense dataset in accordance with the procedure laid out in the paper~\citep{adsorbml}, which is summarized here:

\begin{enumerate}
  %\item Run ML relaxations on all initial structures in OC20-Dense. There are around 1000 different adsorbate-surface combinations with $\sim$90 adsorbate placements per combination, so $\sim$90k in total.
  \item Run ML relaxations on all initial structures in the OC20-Dense dataset. There are around $1000$ different adsorbate-surface combinations with about $90$ adsorbate placements per combination, and therefore we have roughly $90$k structures in total.
  %\item Remove invalid relaxed structures based on the physical constraints and sort based on the lowest ML energy.
  \item Remove invalid ML relaxed structures based on physical constraints and rank the other ML relaxed structures in order of lowest to highest ML predicted energy.
  %\item Take the top 5 (lowest energy) ML relaxed structures for each system and run DFT single-points ($\sim$5000). DFT single-points calculations were run with VASP using the same functional RPBE and settings as the original AdsorbML experiments.
  \item Take the top $k$ ML relaxed structures with the lowest ML predicted energies for each adsorbate-surface combination and run DFT single-point calculations.
  The single-point calculations are performed on the ML relaxed structures to improve the energy predictions without running a full DFT relaxation and are run with VASP using the same setting as the original AdsorbML experiments.
  As shown in Table~\ref{tab:adsorbml}, we vary $k$ from $1$ to $5$.
  %\revision{As shown in Table {\color{crimson}3}, we vary $k$ from $1$ to $5$.}
  %\todo{AdsorbML CPU-hours for DFT single-point calculations.}
  %The calculations are run with VASP using the same functional RPBE and settings as the original AdsorbML experiments.
  %\revision{The DFT single-point calculations are performed once for each relaxed structure and aim at providing a better energy measurement of the ML relaxed structures.}
  %\item Compute success and speedup metrics based on our lowest DFT single-point energy per system and the DFT labels provided with the dataset.
  \item Compute success and speedup metrics based on our lowest DFT single-point energy per adsorbate-surface combination and the DFT labels provided in the OC20-Dense dataset.
\end{enumerate}

\begin{figure}[t]
  \includegraphics[width=0.4\textwidth]{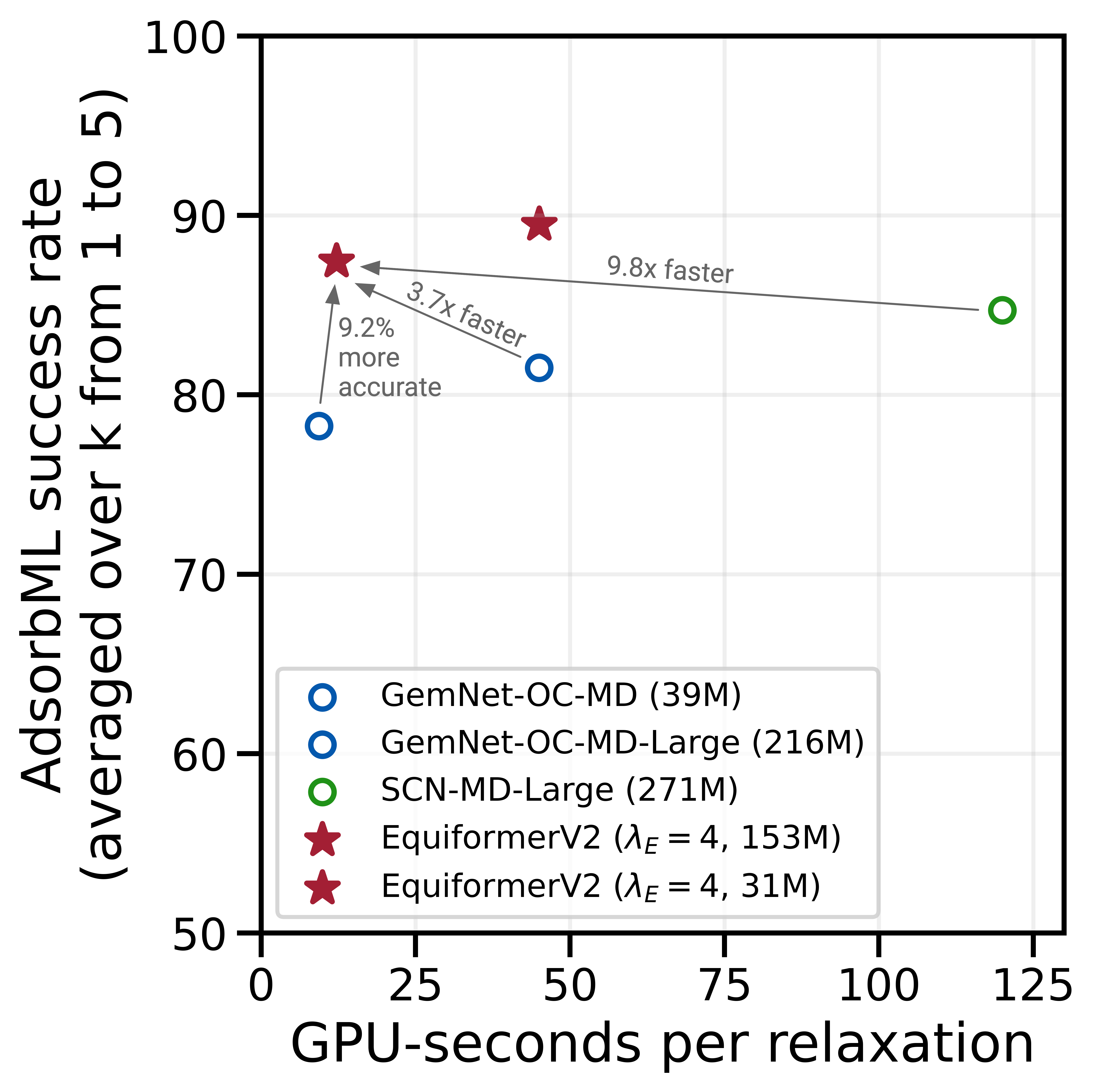}
  \centering
  \vspace{1mm}
  \caption{Speed-accuracy trade-offs of different models when used in the AdsorbML algorithm.
  }
  \vspace{-4mm}
  \label{appendix:fig:adsorbml_speed_sucess}
\end{figure}

\begin{figure}[t]
    \includegraphics[width=0.85\linewidth]{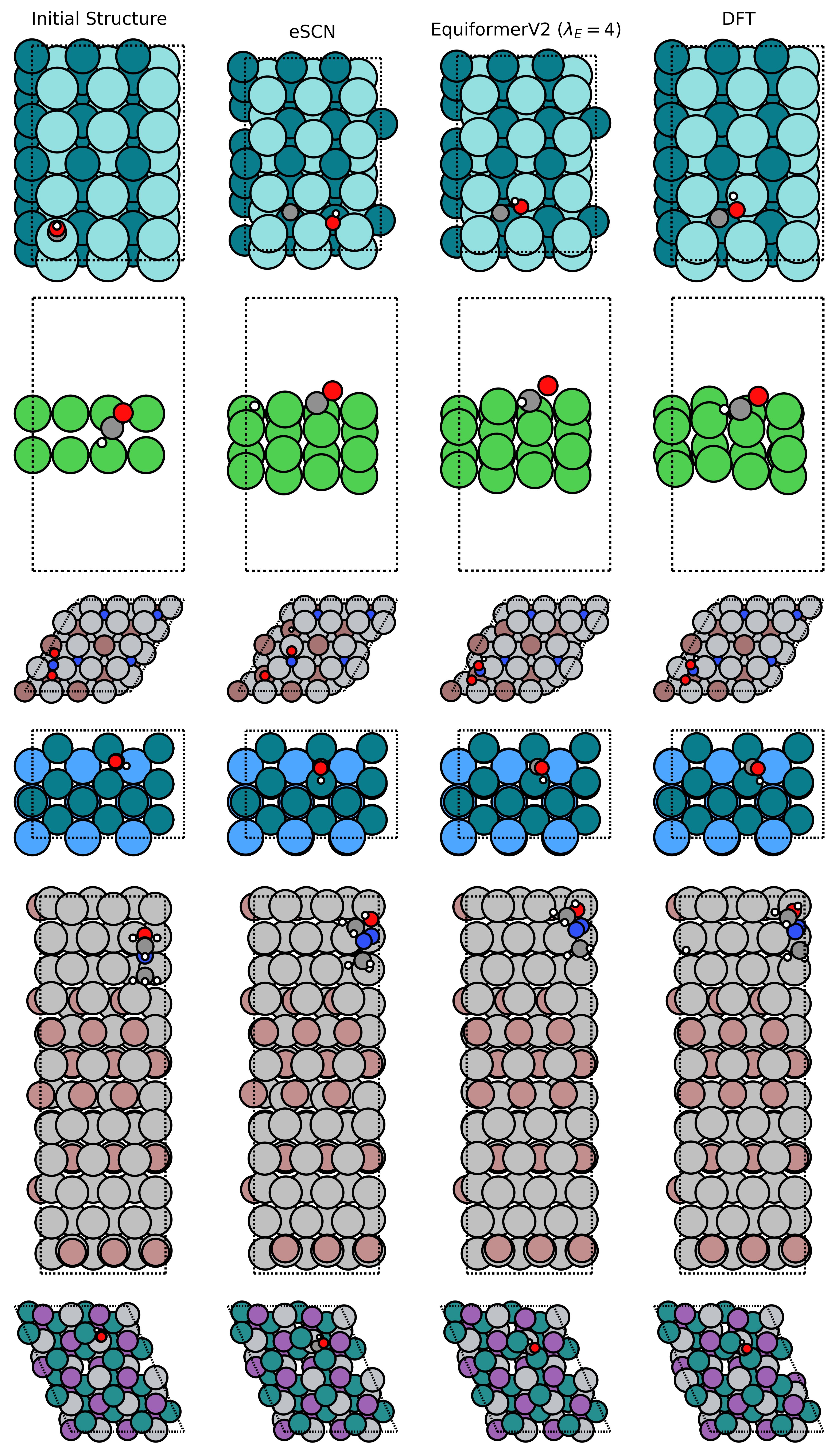}
    \centering
    \vspace{1mm}
    \caption{Qualitative examples of the initial configuration of an adsorbate
    on a catalyst surface (column 1), and corresponding relaxed configurations
    obtained from eSCN~\citep{escn} (column 2), EquiformerV2 (column 3), and DFT
    (column 4). All examples are selected from the OC20-Dense dataset~\citep{adsorbml}.
    We show top-down views of each structure, with dashed lines showing the boundary of the unit cell repeating in the $x$ and $y$ directions.}
    \vspace{-4mm}
    \label{appendix:fig:relaxations_qualitative}
\end{figure}

To better understand the speed-accuracy trade-offs of different models, we compare the AdsorbML success rate averaged over $k$ from $1$ to $5$ and average GPU-seconds of running one structure relaxation in Figure~\ref{appendix:fig:adsorbml_speed_sucess}.
We visualize some examples of relaxed structures from eSCN~\citep{escn}, EquiformerV2 and DFT in Figure~\ref{appendix:fig:relaxations_qualitative}.

\begin{table}[t]
\centering
\scalebox{0.65}{
\begin{tabular}{lll}
%\cline{7-11}
\toprule[1.2pt]
& \multicolumn{1}{c}{EquiformerV2 on OC20 IS2RE} & \multicolumn{1}{c}{EquiformerV2 on OC20 IS2RE} \\
& \multicolumn{1}{c}{} & \multicolumn{1}{c}{with IS2RS auxiliary task and Noisy Nodes} \\
\midrule[1.2pt]
Optimizer & AdamW & AdamW \\
%Learning rate scheduling & Cosine learning rate with linear warmup \\
Learning rate scheduling & Cosine learning rate with & Cosine learning rate with \\
& linear warmup & linear warmup \\
Warmup epochs & $2$ & $2$ \\
Maximum learning rate & $4 \times 10 ^{-4}$ & $5 \times 10 ^{-4}$ \\
Batch size & $32$ & $64$ \\
Number of epochs & $20$ & $40$ \\
Weight decay & $1 \times 10 ^{-3}$ & $1 \times 10 ^{-3}$ \\
Dropout rate & $0.2$ & $0.2$ \\
Stochastic depth & $0.05$ & $0.1$ \\
Gradient clipping norm threshold & $100$ & $100$ \\
Model EMA decay & $0.999$ & $0.999$ \\
Cutoff radius ($\angstrom$) & $5.0$ & $5.0$ \\
Maximum number of neighbors & $50$ & $50$ \\
Number of radial basis & $128$ & $128$ \\
Dimension of hidden scalar features in radial functions $d_{edge}$ & $(0, 64)$ & $(0, 64)$ \\
Maximum degree $L_{max}$ & $6$ & $6$ \\
Maximum order $M_{max}$ & $2$ & $2$ \\
Number of Transformer blocks & $6$ & $15$ \\
Embedding dimension $d_{embed}$ & $(6, 128)$ & $(6, 128)$ \\
$f_{ij}^{(L)}$ dimension $d_{attn\_hidden}$ & $(6, 64)$ & $(6, 64)$ \\
Number of attention heads $h$ & $8$ & $8$ \\
$f_{ij}^{(0)}$ dimension $d_{attn\_alpha}$ & $(0, 64)$ & $(0, 64)$ \\
Value dimension $d_{attn\_value}$ & $(6, 16)$ & $(6, 16)$ \\
Hidden dimension in feed forward networks $d_{ffn}$ & $(6, 128)$ & $(6, 128)$ \\
Resolution of point samples $R$ & $18$ & $18$ \\
\bottomrule[1.2pt]
\end{tabular}
}
\vspace{2mm}
\caption{Hyper-parameters for OC20 IS2RE dataset.
%\todo{Check detailed config.}
}
\label{appendix:tab:oc20_is2re_hyperparameters}
\end{table}

\subsection{Additional Comparison with Equiformer on OC20 IS2RE}
\label{appendix:subsec:additional_comparison_on_oc20_is2re}

\paragraph{Training Details.}
We follow the same setting as Equiformer~\citep{equiformer} and train two EquiformerV2 models on OC20 IS2RE dataset without and with IS2RS auxiliary task.
We use the same radial basis function as Equiformer.
When IS2RS auxiliary task is adopted, we use a linearly decayed weight for loss associated with IS2RS, which starts at $15$ and decays to $1$ and adopt Noisy Nodes data augmentation~\citep{noisy_nodes}.
The hyper-parameters are summarized in Table~\ref{appendix:tab:oc20_is2re_hyperparameters}.
We train EquiformerV2 on OC20 IS2RE with 16 V100 GPUs with 32GB and train on OC20 IS2RE with IS2RS auxiliary task and Noisy Nodes data augmentation with 32 V100 GPUs.
The training costs are $574$ and $2075$ GPU-hours, and the numbers of parameters are $36.03$M and $95.24$M.

\paragraph{Results.}
\revision{
The comparison is shown in Table~\ref{appendix:tab:oc20_is2re_results}.
Without IS2RS auxiliary task, EquiformerV2 overfits the training set due to higher degrees and achieves worse results than Equiformer.
However, with IS2RS auxiliary task and Noisy Nodes data augmentation, EquiformerV2 achieves better energy MAE. 
The different rankings of models under different settings is also found in Noisy Nodes~\citep{noisy_nodes}, where they mention a node-level auxiliary task can prevent overfitting and enable more expressive models to perform better.
}

\begin{table}[t!]
\resizebox{\textwidth}{!}{%
\begin{tabular}{lcccccccccc}
\toprule[1.2pt]
%& \multicolumn{5}{c}{\begin{tabular}[c]{@{}c@{}}IS2RE direct \\ Energy MAE (meV)$\downarrow$\end{tabular}} & \multicolumn{5}{c}{\begin{tabular}[c]{@{}c@{}}IS2RE direct with IS2RS auxiliary task and Noisy Nodes \\ Energy MAE (meV)$\downarrow$\end{tabular}} \\ %\hline
& \multicolumn{5}{c}{IS2RE direct} & \multicolumn{5}{c}{IS2RE direct with IS2RS auxiliary task and Noisy Nodes} \\
\cmidrule(lr){2-6} \cmidrule(lr){7-11} 
& \multicolumn{5}{c}{Energy MAE (meV)$\downarrow$} & \multicolumn{5}{c}{Energy MAE (meV)$\downarrow$} \\
Model        & ID              & OOD Ads          & OOD Cat          & OOD Both          & Average          & ID                   & OOD Ads                & OOD Cat               & OOD Both               & Average               \\
\midrule[1.2pt] 

Equiformer~\citep{equiformer} & \textbf{508.8}           & \textbf{627.1}            & \textbf{505.1}            & \textbf{554.5}             & \textbf{548.9}            & 415.6                & 497.6                  & 416.5                 & 434.4                  & 441.0                 \\
EquiformerV2 & 516.1           & 704.1            & 524.5            & 636.5             & 595.3            & \textbf{400.4}                & \textbf{459.0}                  & \textbf{406.2}                 & \textbf{401.8}                  & \textbf{416.9} \\
\bottomrule[1.2pt]
\end{tabular}}
\vspace{2mm}
\caption{OC20 IS2RE results on the validation split.
With IS2RS auxiliary task and Noisy Nodes~\citep{noisy_nodes} data augmentation, EquiformerV2 achieves better energy MAE.
}
\label{appendix:tab:oc20_is2re_results}
\end{table}

%% file: content/8_4_details_of_experiments_on_oc22.tex
\section{Details of Experiments on OC22}
\label{appendix:sec:details_of_experiments_on_oc22}

\subsection{Training Details}
\label{appendix:subsec:oc22_training_details}

The hyper-parameters for OC22 dataset is summarized in Table~\ref{appendix:tab:oc22_hyperparameters}. 
We use 32 V100 GPUs with 32GB and train two EquiformerV2 models with different energy coefficients $\lambda_E$ and force coefficeints $\lambda_F$.
The number of parameters is $121.53$M, and the training cost of each model is $4552$ GPU-hours. 
The time for running relaxations for OC22 IS2RE is $38$ GPU-hours.

\begin{table}[t]
\centering
\scalebox{0.65}{
\begin{tabular}{ll}
%\cline{7-11}
\toprule[1.2pt]
Hyper-parameters & Value or description \\
\midrule[1.2pt]
Optimizer & AdamW \\
Learning rate scheduling & Cosine learning rate with linear warmup \\
Warmup epochs & $0.1$ \\
Maximum learning rate & $2 \times 10 ^{-4}$ \\
Batch size & $128$ \\
Number of epochs & $6$ \\
Weight decay & $1 \times 10 ^{-3}$ \\
Dropout rate & $0.1$ \\
Stochastic depth & $0.1$ \\
Energy coefficient $\lambda_{E}$ & $1, 4$ \\
Force coefficient $\lambda_{F}$ & $1, 100$ \\
Gradient clipping norm threshold & $50$ \\
Model EMA decay & $0.999$ \\
Cutoff radius ($\angstrom$) & $12$ \\
Maximum number of neighbors & $20$ \\
Number of radial bases & $600$ \\
Dimension of hidden scalar features in radial functions $d_{edge}$ & $(0, 128)$ \\
Maximum degree $L_{max}$ & $6$ \\
Maximum order $M_{max}$ & $2$ \\
Number of Transformer blocks & $18$ \\
Embedding dimension $d_{embed}$ & $(6, 128)$ \\
$f_{ij}^{(L)}$ dimension $d_{attn\_hidden}$ & $(6, 64)$ \\
Number of attention heads $h$ & $8$ \\
$f_{ij}^{(0)}$ dimension $d_{attn\_alpha}$ & $(0, 64)$ \\
Value dimension $d_{attn\_value}$ & $(6, 16)$ \\
Hidden dimension in feed forward networks $d_{ffn}$ & $(6, 128)$ \\
Resolution of point samples $R$ & $18$ \\
\bottomrule[1.2pt]
\end{tabular}
}
\vspace{2mm}
\caption{Hyper-parameters for OC22 dataset.
}
\label{appendix:tab:oc22_hyperparameters}
\end{table}

%% file: content/8_5_details_of_experiments_on_qm9.tex
\section{Details of Experiments on QM9}
\label{appendix:sec:details_of_experiments_on_qm9}

\subsection{Additional Results of Training with Noisy Nodes}
\label{appendix:subsec:qm9_additional_results}

Similar to Sec.~\ref{appendix:subsec:additional_comparison_on_oc20_is2re}, we train EquiformV2 on QM9 with Noisy Nodes~\citep{noisy_nodes} to show that the performance gain brought by using higher degrees can be larger when trained with a node-level auxiliary task and data augmentation.
In Table~\ref{appendix:tab:qm9_noisy_nodes_results}, we summarize the results and compare with previous works using Noisy Nodes and pre-training via denoising~\citep{pretraining_via_denoising}.
When trained with Noisy Nodes, EquiformerV2 performs better than Equiformer on more tasks.
Specifically, without Noisy Nodes, EquiformerV2 is better than Equiformer on 9 out of the 12 tasks, similar on 1 task, and worse on the other 2 tasks. 
With Noisy Nodes, EquiformerV2 achieves better MAE on 10 tasks, similar on 1 task, and worse on 1 task. 
Additionally, we note that EquiformerV2 with Noisy Nodes is overall better than GNS-TAT with both pre-training and Noisy Nodes~\citep{pretraining_via_denoising} on 9 out of the 12 tasks even though EquiformerV2 is not pre-trained on PCQM4Mv2 dataset, which is more than $30\times$ larger than QM9. 
This shows that a more expressive model can match the performance with significantly less data.

%\todo{Showing that with Noisy Nodes, EquiformerV2 performs better than EquiformerV1 on more tasks.}

\begin{table}[t]
\centering
\resizebox{1.0\textwidth}{!}{
\begin{tabular}{llcccccccccccc}
\toprule[1.2pt]
& Task & $\alpha$ & $\Delta \varepsilon$ & $\varepsilon_{\text{HOMO}}$ & $\varepsilon_{\text{LUMO}}$ & $\mu$ & $C_{\nu}$ & $G$ & $H$ & $R^2$ & $U$ & $U_0$ & ZPVE \\ 
Model & Units & $a_0^3$ & meV & meV & meV & D & cal/mol K & meV & meV & $a_0^2$ & meV & meV & meV\\
\midrule[1.2pt]

Equiformer~\citep{equiformer} & & .046 & 30 & 15 & 14 & .011 & .023 & 7.63 & 6.63 & .251 & 6.74 & 6.59 & 1.26 \\
Equiformer~\citep{equiformer} + NN$^\dagger$ & & .040 & 26.4 & 13.7 & 13.0 & .011 & \textbf{.020} & 5.49 & 4.61 & .235 & 4.81 & 4.61 & 1.18 \\

GNS-TAT + NN~\citep{pretraining_via_denoising}$^\ddagger$ & & .047 & 25.7 & 17.3 & 17.1 & .021 & .022 & 7.41 & 6.42 & 0.65 & 6.39 & 6.39 & 1.080 \\ 
GNS-TAT + NN + pretraining ~\citep{pretraining_via_denoising}$^\ddagger$ & & .040 & \textbf{22.0} & 14.9 & 14.7 & .016 & \textbf{.020} & 6.90 & 5.79 & 0.44 & 5.76 & 5.76 & \textbf{1.018} \\ 

\midrule

EquiformerV2 & & .050          & 29       & 14        & 13        & .010 & .023 & 7.57 & 6.22 & .186 &  6.49 & 6.17 & 1.47 \\
EquiformerV2 + NN & & \textbf{.039} & 24.2 & \textbf{12.2} & \textbf{11.4} & \textbf{.009} & \textbf{.020} & \textbf{5.34} & \textbf{4.24} &  \textbf{.182} & \textbf{4.28} & \textbf{4.34} & 1.21 \\
%\midrule[0.6pt]
%Equiformer + NAT & 0.4737 & 0.7245 & 0.4862 & 0.6493 & 0.5834 & 6.01 & 2.45 & 5.41 & 2.71 & \\
\bottomrule[1.2pt]

\end{tabular}
}
\vspace{1mm}
%\end{adjustwidth}
\caption{
Mean absolute error results on QM9 test set when trained with Noisy Nodes~\citep{noisy_nodes}.  
%\note{Dropout rate = 0.1.}
%%%$\dagger$ denotes using different training, validation, testing data partitions. % as mentioned in SEGNN~\cite{segnn}.
$\dagger$ denotes that the results are produced by this work.
$\ddagger$ denotes using different data partitions.
``NN'' denotes Noisy Nodes, and ``pretraining'' denotes pretraining on PCQM4Mv2 dataset~\citep{pcqm4mv2}.
The performance gain from Equiformer to EquiformerV2 becomes larger when trained with Noisy Nodes.  
%$\ddagger$ denotes results from SE(3)-Transformer~\cite{se3_transformer}.
}
%\vspace{-10pt}
\label{appendix:tab:qm9_noisy_nodes_results}
\end{table}

\subsection{Training Details}
\label{appendix:subsec:qm9_training_details}

We follow the data partition of Equiformer.
For the tasks of $\mu$, $\alpha$, $\varepsilon_{\text{HOMO}}$, $\varepsilon_{\text{LUMO}}$, $\Delta\varepsilon$, and $C_{\nu}$, we use batch size $= 64$, the number of epochs $= 300$, learning rate $= 5 \times 10^{-4}$, Gaussian radial basis functions with the number of bases $= 128$, the number of Transformer blocks $=6$, weight decay $= 5 \times 10^{-3}$, and dropout rate $= 0.2$ and use mixed precision for training.
For the task of $R^2$, we use batch size $= 48$, the number of epochs $= 300$, learning rate $= 1.5 \times 10^{-4}$, Gaussian radial basis functions with the number of bases $= 128$, the number of Transformer blocks $=5$, weight decay $= 5 \times 10^{-3}$, and dropout rate $= 0.1$ and use single precision for training.
For the task of ZPVE, we use batch size $= 48$, the number of epochs $= 300$, learning rate $= 1.5 \times 10^{-4}$, Gaussian radial basis functions with the number of bases $= 128$, the number of Transformer blocks $=5$, weight decay $= 5 \times 10^{-3}$, and dropout rate $= 0.2$ and use single precision for training.
For the task of $G$, $H$, $U$, and $U_0$, we use batch size $= 48$, the number of epochs $= 300$, learning rate $= 1.5 \times 10^{-4}$, Gaussian radial basis functions with the number of bases $= 128$, the number of Transformer blocks $=5$, weight decay $= 0.0$, and dropout rate $= 0.0$ and use single precision for training.
Other hyper-parameters are the same across all the tasks, and we summarize them in Table~\ref{appendix:tab:qm9_hyperparameters}.
We use a single A6000 GPU and train different models for different tasks.
The training costs are $72$ GPU-hours for mixed precision training and $137$ GPU-hours for single precision training.
The number of parameters are $11.20$M for $6$ blocks and $9.35$M for $5$ blocks.

%\todo{Add Noisy Nodes detail -- partially corruptes structures, denoising probability.}

%\todo{Add Noisy Nodes detail -- hyper-parameters and training time.}

As for training with Noisy Nodes as mentioned in Sec.~\ref{appendix:subsec:qm9_additional_results}, we add noise to atomic coordinates and incorporate a node-level auxiliary task of denoising atomic coordinates.
We thus introduce four additional hyper-parameters, which are noise standard deviation $\sigma_{\text{denoise}}$, denoising coefficient $\lambda_{\text{denoise}}$, denoising probability $p_{\text{denoise}}$ and corrupt ratio $r_{\text{denoise}}$. 
The noise standard deviation $\sigma_{\text{denoise}}$ denotes the standard deviation of Gaussian noise added to each xyz component of atomic coordinates.
The denoising coefficient $\lambda_{\text{denoise}}$ controls the relative importance of the auxiliary task compared to the original task.
The denoising probability $p_{\text{denoise}}$ denotes the probability of adding noise to atomic coordinates and optimizing for both the auxiliary task and the original task. 
Using $p_{\text{denoise}} < 1$ enables taking original atomistic structures without any noise as inputs and optimizing for only the original task for some training iterations.
The corrupt ratio $r_{\text{denoise}}$ denotes the ratio of the number of atoms, which we add noise to and denoise, to the total number of atoms. 
Using $r_{\text{denoise}} < 1$ allows only adding noise to and denoising a subset of atoms within a structure.
For the task of $R^2$, we use $\sigma_{\text{denoise}} = 0.02$, $\lambda_{\text{denoise}} = 0.1$, $p_{\text{denoise}} = 0.5$ and $r_{\text{denoise}} = 0.125$.
For other tasks, we use $\sigma_{\text{denoise}} = 0.02$, $\lambda_{\text{denoise}} = 0.1$, $p_{\text{denoise}} = 0.5$ and $r_{\text{denoise}} = 0.25$.
We share the above hyper-parameters for training EquiformerV2 and Equiformer, and we add one additional block of equivariant graph attention for the auxiliary task.
We slightly tune other hyper-parameters when trained with Noisy Nodes.
For Equiformer, we additionally use stochastic depth $=0.05$ for the tasks of $\alpha, \Delta \varepsilon, \varepsilon_{\text{HOMO}}, \varepsilon_{\text{LUMO}}, \text{ and } C_{\nu}$.
As for EquiformerV2, we additionally use stochastic depth $=0.05$ for the tasks of $\mu, \Delta \varepsilon, \varepsilon_{\text{HOMO}}, \varepsilon_{\text{LUMO}}, \text{ and } C_{\nu}$.
We increase the number of blocks from $5$ to $6$ and increase the batch size from $48$ to $64$ for the tasks of $G, H, U, \text{and } U_0$.
We increase the learning rate from $1.5 \times 10^{-4}$ to $5 \times 10^{-4}$ and increase the number of blocks from $5$ to $6$ for the task of $R^2$.

\begin{table}[t]
\centering
\scalebox{0.65}{
\begin{tabular}{ll}
%\cline{7-11}
\toprule[1.2pt]
Hyper-parameters & Value or description \\
\midrule[1.2pt]
Optimizer & AdamW \\
Learning rate scheduling & Cosine learning rate with linear warmup \\
Warmup epochs & $5$ \\
Maximum learning rate & $1.5 \times 10^{-4}, 5 \times 10 ^{-4}$ \\
Batch size & $48, 64$ \\
Number of epochs & $300$ \\
Weight decay & $0.0, 5 \times 10^{-3}$ \\
Dropout rate & $0.0, 0.1, 0.2$ \\
Stochastic depth & $0.0, 0.05$ \\
Cutoff radius ($\angstrom$) & $5.0$ \\
Maximum number of neighbors & $500$ \\
Number of radial bases & $128$ \\
Dimension of hidden scalar features in radial functions $d_{edge}$ & $(0, 64)$ \\
Maximum degree $L_{max}$ & $4$ \\
Maximum order $M_{max}$ & $4$ \\
Number of Transformer blocks & $5, 6$ \\
Embedding dimension $d_{embed}$ & $(4, 96)$ \\
$f_{ij}^{(L)}$ dimension $d_{attn\_hidden}$ & $(4, 48)$ \\
Number of attention heads $h$ & $4$ \\
$f_{ij}^{(0)}$ dimension $d_{attn\_alpha}$ & $(0, 64)$ \\
Value dimension $d_{attn\_value}$ & $(4, 24)$ \\
Hidden dimension in feed forward networks $d_{ffn}$ & $(4, 96)$ \\
Resolution of point samples $R$ & $18$ \\

\midrule

Noise standard deviation $\sigma_{\text{denoise}}$ & $0.02$ \\
Denoising coefficient $\lambda_{\text{denoise}}$ & $0.1$ \\ 
Denoising probability $p_{\text{denoise}}$ & $0.5$ \\
Corrupt ratio $r_{\text{denoise}}$ & $0.125, 0.25$ \\

\bottomrule[1.2pt]
\end{tabular}
}
\vspace{2mm}
\caption{Hyper-parameters for QM9 dataset.
}
\label{appendix:tab:qm9_hyperparameters}
\end{table}

\subsection{Ablation Study on Architectural Improvements}
\label{appendix:subsec:qm9_ablation_study}

We conduct ablation studies on the proposed architectural improvements using the task of $\Delta \varepsilon$ on QM9 and compare with Equiformer baseline~\citep{equiformer}.
The reults are summarized in Table~\ref{appendix:tab:qm9_ablation_study}.
The comparison between Index 0 and Index 1 shows that directly increasing $L_{max}$ from $2$ to $4$ and using eSCN convolutions degrade the performance.
This is due to overfitting since the QM9 dataset is smaller, and each structure in QM9 has fewer atoms, less diverse atom types and much less angular variations than OC20 and OC22. 
%This is not surprising since the QM9 dataset has much fewer examples in the training set than the OC20 and OC22 datasets, and each example has less atoms, less diverse atom types and much less angular variations.
%All of these lead to overfitting when simply using higher degrees.
Comparing Index 1 and Index 2, attention re-normalization clearly improves the MAE result.
Although using $S^2$ activation is stable here (Index 3) unlike OC20, it results in higher error than using gate activation (Index 2) and the Equiformer baseline (Index 0). 
When using the proposed separable $S^2$ activation (Index 4), we achieve lower error than using gate activation (Index 2). 
We can further reduce the error by using the proposed separable layer normalization (Index 5).
Comparing Index 0 and Index 5, we note that the proposed architectural improvements are necessary to achieve better results than the baseline when using higher degrees on QM9.
%Overall, the ranking of relative performance of different models is the same as that on the 2M split of the OC20 S2EF dataset, and the proposed three architectural improvements are generalizable. 
Overall, these ablation results follow the same trends as OC20.

\begin{table}[t!]
\centering
\scalebox{0.65}{
\begin{tabular}{lccccc}
\toprule[1.2pt]
& Attention & & & & \\
Index & re-normalization & Activation & Normalization & $L_{max}$ & $\Delta \varepsilon$ MAE (meV) \\
%\shline
\midrule[1.2pt]
0 & \multicolumn{3}{l}{Equiformer baseline} & $2$ & 29.98 \\
\midrule 
1 & \xmark & Gate & LN & $4$ & 30.46 \\
2 & \cmark & Gate & LN & $4$ & 29.51 \\
3 & \cmark & $S^2$ & LN & $4$ & 30.23 \\
4 & \cmark & Sep. $S^2$ & LN & $4$ & 29.31 \\
5 & \cmark & Sep. $S^2$ & SLN & $4$ & 29.03 \\
\bottomrule[1.2pt]
\end{tabular}
}
\vspace{2mm}
\caption{
Ablation studies on the proposed architectural improvements using the task of $\Delta \varepsilon$ of the QM9 dataset.
}
\label{appendix:tab:qm9_ablation_study}
\end{table}